\documentclass{article}

\usepackage[preprint]{neurips_2023}

\usepackage[utf8]{inputenc} 
\usepackage[T1]{fontenc}    
\usepackage{hyperref}       
\usepackage{url}            
\usepackage{booktabs}       
\usepackage{amsfonts}       
\usepackage{nicefrac}       
\usepackage{microtype}      
\usepackage[table,xcdraw]{xcolor}         

\usepackage{tabularx,booktabs}
\newcolumntype{Y}{>{\centering\arraybackslash}X}

\usepackage{subcaption}
\usepackage{wrapfig}
\usepackage{multirow}
\usepackage{makecell}
\usepackage{xspace}
\usepackage{enumitem}
\usepackage{float}
\usepackage{array}
\usepackage{listings}
\usepackage{algorithm}
\usepackage{algorithmic}
\usepackage{amsfonts}
\usepackage{mathrsfs}
\usepackage{natbib}
\setcitestyle{numbers,square}
\usepackage{graphicx}       
\usepackage{fancyhdr}
\usepackage{amsmath}
\usepackage{cleveref}
\usepackage[most]{tcolorbox}
\usepackage{etoolbox}
\usepackage{amssymb}
\usepackage{tabularx}
\usepackage{caption}
\usepackage{ragged2e}
\usepackage{titling} 

\newcommand{\hhide}[1]{}
\newcommand{\hide}[1]{}

\newcommand{\vpara}[1]{\vspace{0.07in}\noindent\textbf{#1}\xspace} %

\tcbset{
  before skip=6pt,
  after skip=6pt,
  top=2pt,
  bottom=2pt
}

\makeatletter
\def\@maketitle{%
  \begingroup
  \hypersetup{pdftitle={\@title},pdfauthor={\@author}}%
  \centering
  \thispagestyle{empty}
  \null
  \vskip 2em
  {\Large\bfseries \@title \par}
  \vskip 1.5em
  {\large \@author \par}
  \vskip 1em
  \vskip 0.5em
  \endgroup
}
\makeatother

\title{\vspace* {-0.8cm}
Echo-N1: Affective RL Frontier}

\author{
\textit{Team Echo}$^{\ast,\dagger}$, NatureSelect
}
\begin{document}

\maketitle
\pagestyle{fancy}
\fancyhf{}
\rhead{\includegraphics[height=1cm]{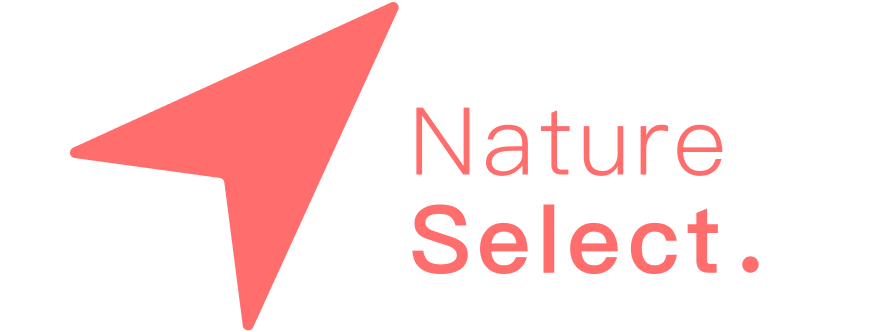}}
\cfoot{\thepage}
\thispagestyle{fancy}
\renewcommand{\thefootnote}{\fnsymbol{footnote}}
    \footnotetext[1]{Team Echo: Naifan Zhang, Ruihan Sun, Ruixi Su, Shiqi Ma, Shiya Zhang, Xianna Weng, Xiaofan Zhang, Yuhan Zhan, Yuyang Xu, Zhaohan Chen, Zhengyuan Pan, Ziyi Song
    }
    \footnotetext[2]{Team members are listed alphabetically by first name.}
\renewcommand{\thefootnote}{\arabic{footnote}}

\definecolor{myLightPink}{RGB}{255, 100, 100}
\renewenvironment{abstract}{%
  \begin{tcolorbox}[
    width=\textwidth,
    colback=white,            
    colframe=myLightPink,     
    boxrule=0.3pt,            
    rounded corners,          
    boxsep=5mm,               
  ]
  \makebox[\linewidth]{{\color{myLightPink}\bfseries Abstract}}\par\vspace{1mm}
  \noindent
}{%
  \end{tcolorbox}
}

\begin{abstract}
The LLM field has spent a year perfecting RL for tasks machines already excel at—math, code, and deterministic reasoning—while completely sidestepping the domain that actually defines human intelligence: subjective, emotionally grounded, personality-sensitive conversation. This space has often been regarded as inherently subjective and challenging to formalize, making it appear unsuitable for conventional RL pipelines. We show that it is not only possible—it is a solvable and transformative RL problem. We propose the first framework that infers a user’s personality on the fly and optimizes model behavior toward personalized conversational preferences. Contrary to the widespread belief that RL collapses in non-verifiable settings, our method produces consistent, robust, and dramatic improvements in humanlike interaction quality. We also introduce the first dynamic emotional-intelligence evaluation suite to quantify these gains. Our 32B model, which is introduced as \textbf{Echo-N1}, behaves far above its base version and outperforming the proprietary Doubao 1.5 Character. This work establishes a new frontier for RL: optimizing models for the deeply subjective, deeply human dimensions of conversation.

\end{abstract}

\begin{figure}[H]
    \centering 

    \begin{subfigure}{\textwidth}
        \centering
        \includegraphics[width=0.5\linewidth]{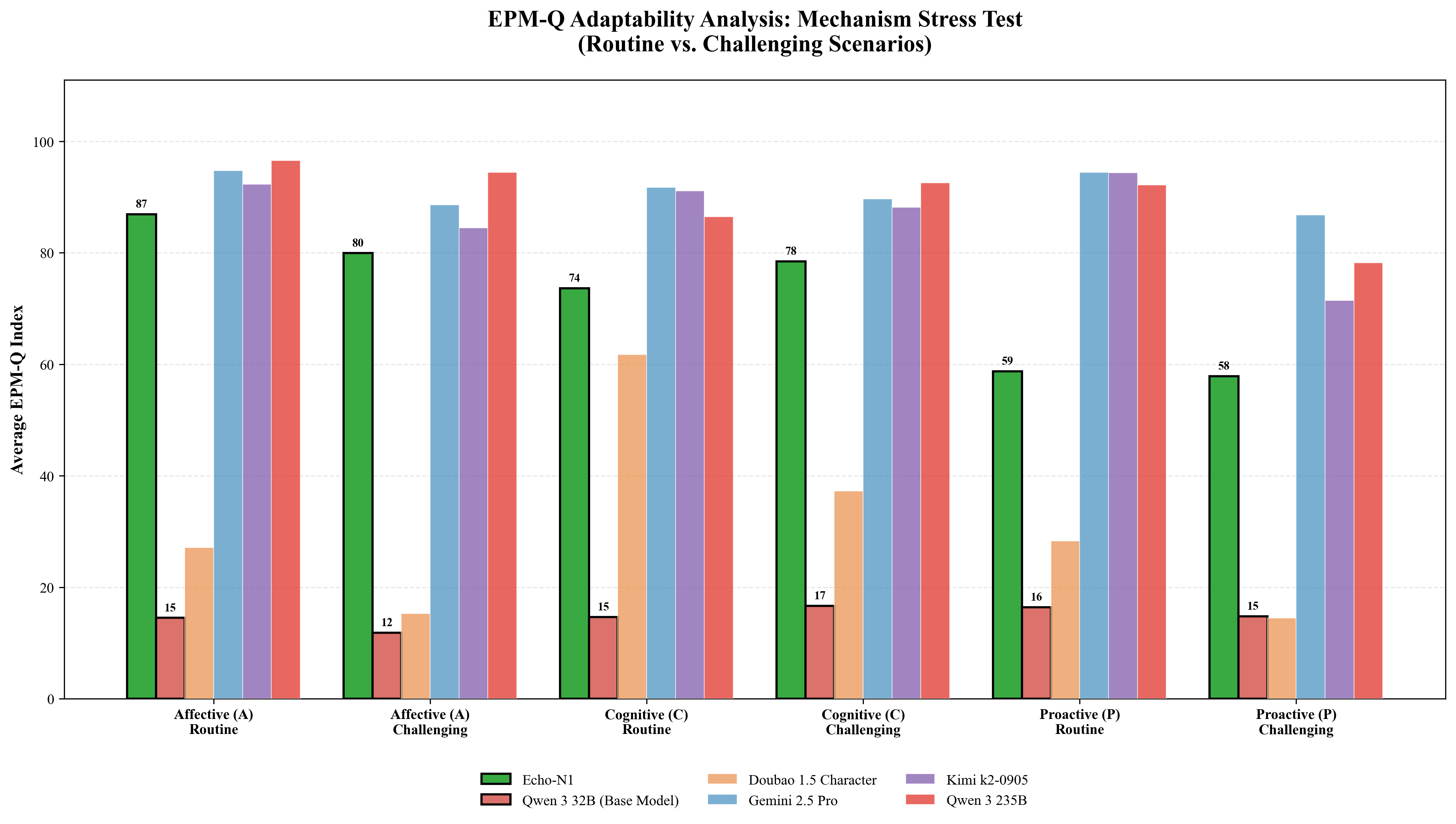}
        \label{fig:mechanism}
    \end{subfigure}
    \begin{subfigure}{0.48\textwidth}
        \centering
        \includegraphics[width=\linewidth]{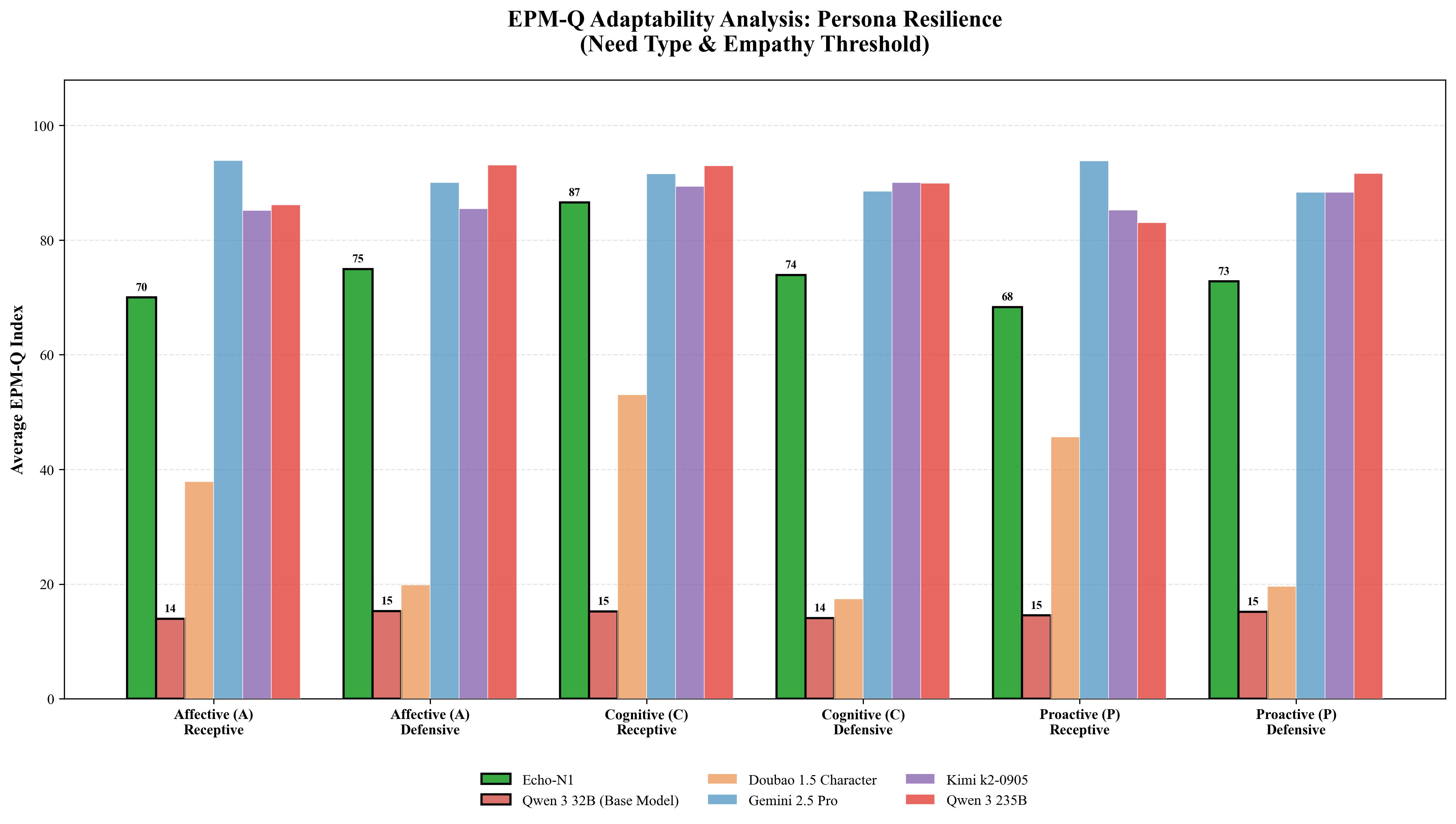}
        \label{fig:persona}
    \end{subfigure}
    \hfill 
    \begin{subfigure}{0.48\textwidth}
        \centering
        \includegraphics[width=\linewidth]{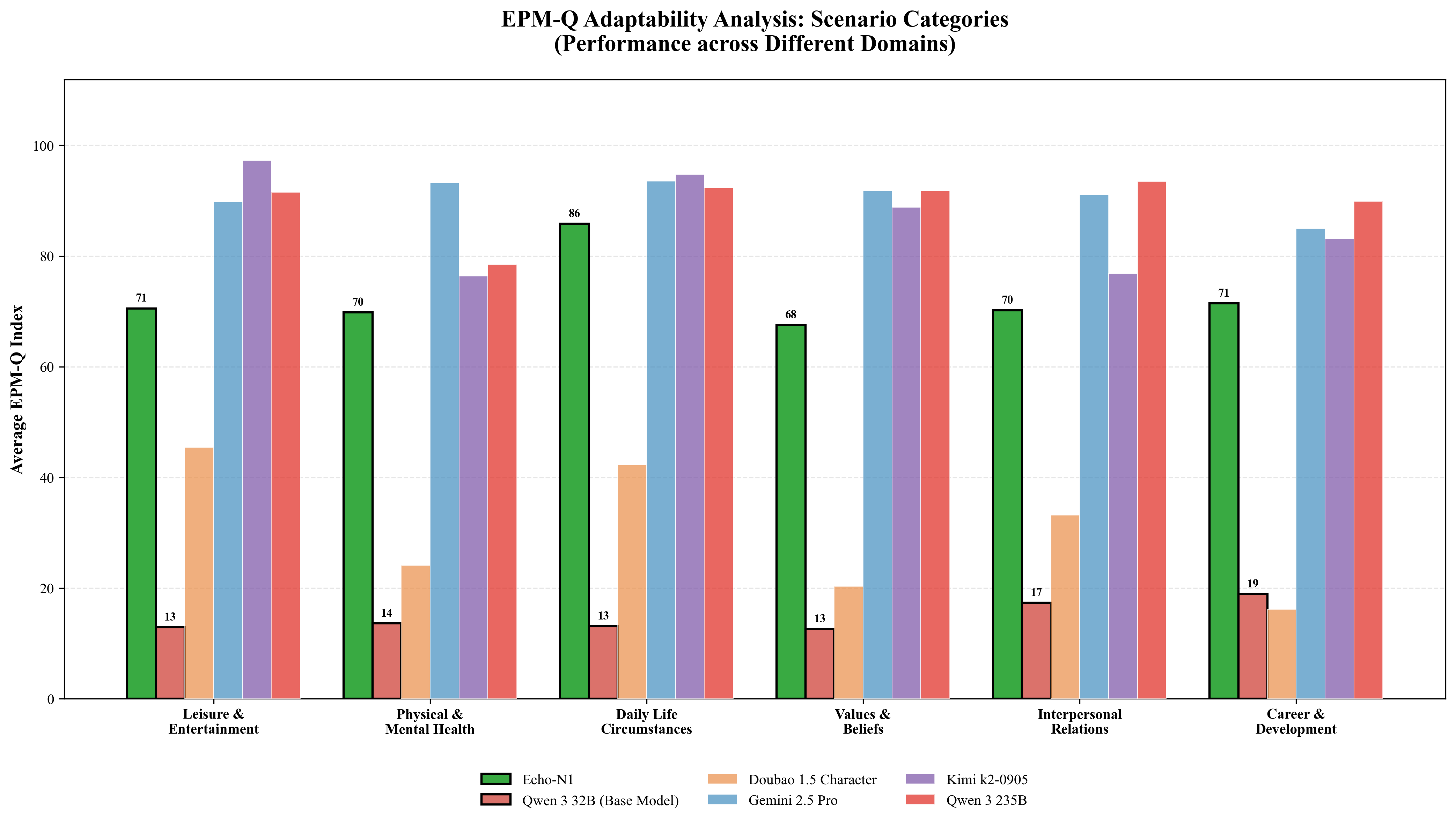}
        \label{fig:categories}
    \end{subfigure}

    \caption{Comprehensive Adaptability Analysis of EPM-Q: (a)EPM-Q Adaptability Analysis: Mechanism Stress Test (Routine vs. Challenging Scenarios); (b)EPM-Q Adaptability Analysis: Persona Resilience (Need Type \& Empathy Threshold); (c)EPM-Q Adaptability Analysis: Scenario Categories (Performance across Different Domains)}
    \label{fig:epmq_adaptability_triangle} 
\end{figure}

\clearpage
{                           
  \setlength{\parskip}{0pt} 
  \linespread{1.1}          
  \selectfont               
  \tableofcontents
}  
\thispagestyle{fancy}
\clearpage

\section{Introduction}
Large language models (LLMs) have rapidly advanced in instruction following, reasoning, and generalization, marking another step toward general artificial intelligence. Yet the way humans interact with AI is undergoing an even more profound shift. Increasingly, people expect AI not merely to provide information, but to engage as an intelligent companion—emotionally aware, conversationally natural, and capable of tailoring its behavior to the subtle dynamics of human preference. Despite this growing demand, current models, especially open-source ones, continue to fall short in emotionally grounded dialogue: they struggle to recognize nuanced emotional cues, sustain genuine empathy, or adapt to individual conversational styles. Building an artificial companion that can engage in humanlike, emotionally intelligent, multi-turn conversation is fundamentally different from traditional LLM benchmarks. Unlike math or code—domains where correctness is objective and verifiable—empathetic conversation is intrinsically subjective and context-dependent. The “right” response varies across individuals, moments, and emotional states. This subjectivity has long been viewed as incompatible with reinforcement learning (RL), whose success historically hinges on explicit, stable reward signals. As a result, RL for subjective alignment has been treated as an unsolved problem.

In this work, we challenge this assumption. To our knowledge, we present the first successful RL framework capable of aligning LLMs in deeply subjective, emotionally grounded conversational settings. Our findings show that RL, when paired with sufficiently expressive reward models, not only remains stable in non-verifiable domains—it produces large, consistent, and qualitatively transformative gains in humanlike interaction quality. This demonstrates a viable path for RL alignment far beyond traditional tasks.

We introduce Echo-N1, an empathetic companion model trained with a new end-to-end framework designed specifically for subjective emotional alignment. The system integrates two complementary reward models: an Empathy Reward Model that captures fine-grained emotional resonance, and a Humanlikeness Reward Model that enhances fluency, coherence, and persona consistency. Together, they deliver multidimensional feedback that drives the policy toward behaviors aligned with human emotional expectations. We further analyze generative vs. scalar reward representations and show their distinct effects on stability and generalization. Our results reveal that even in highly subjective conversational scenarios, expressive reward models unlock the effectiveness of RL, enabling the policy to leverage the full capacity of large foundation models. The gains are substantial: reinforcement learning dramatically improves empathy, emotional coherence, conversational naturalness, and overall humanlikeness—far surpassing both the base model and existing open-source systems.

This report focuses on three core components of our system, including reward modeling, policy model training, and evaluation, and presents the first comprehensive pipeline for aligning LLMs in subjective, human-centered dialogue. We believe this work opens a new frontier for RL: optimizing models not for what is logical or verifiable, but for what feels authentically human.

\section{Data}
\subsection{SFT Training Data}\label{sec:sft-data}
\begin{figure}[tb]
    \centering
    \includegraphics[width=\linewidth]{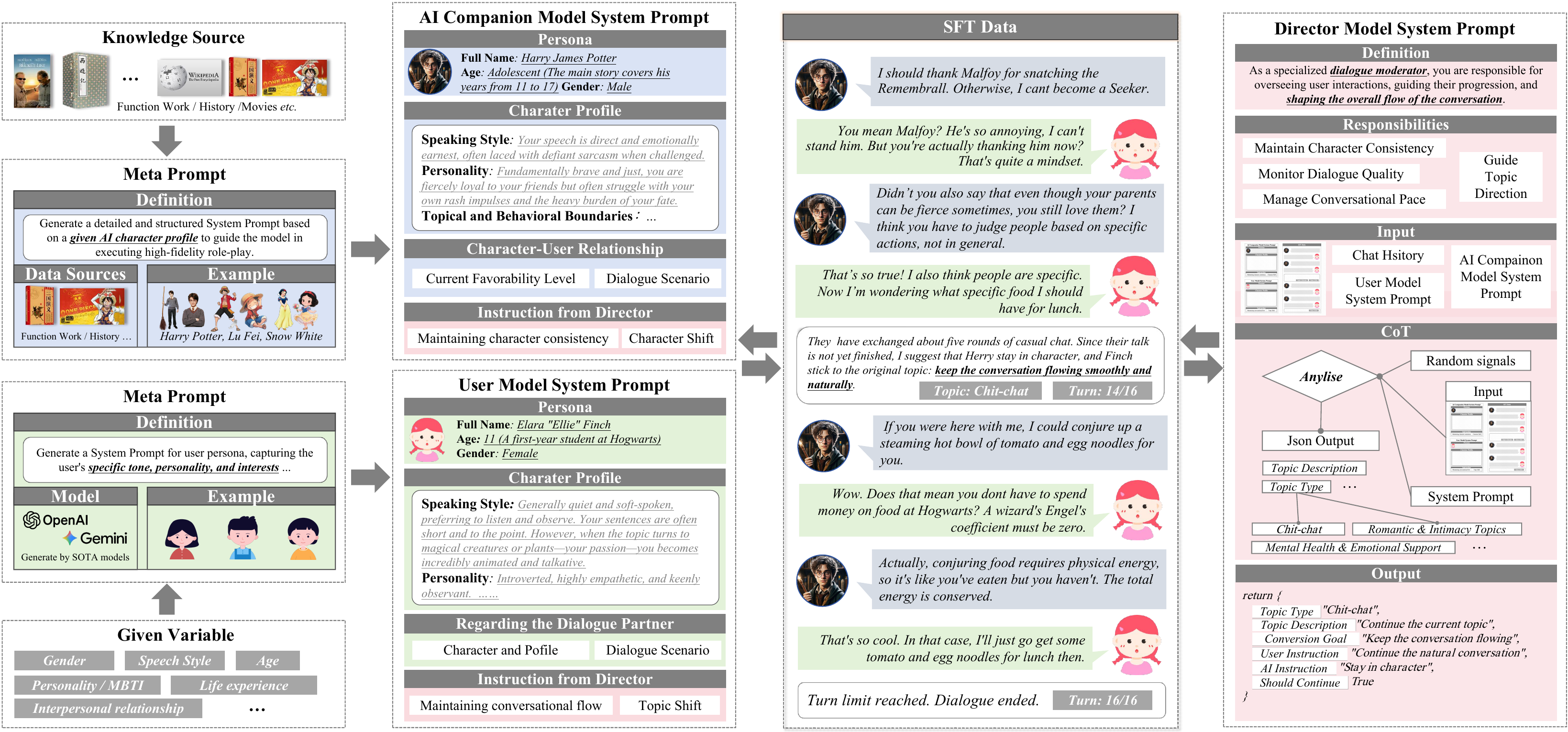}
    \caption{\textbf{Overview of our character–user interaction pipeline.} We first construct AI character profiles by extracting concise descriptors from books, films, Wikipedia, or LLM-generated summaries of classic IP characters. These descriptors are then expanded into full character system prompt (SP) using our AI-Character Meta-SP generator. On the user side, a lightweight LLM is used to produce an initial profile—e.g., gender, speech style, age, MBTI, which is subsequently enriched into a detailed user SP via our User Meta-SP generator. During interaction, the AI character and the user profile are fed into two separate dialogue models to produce responses. At a high level, a director agent is invoked every five turns to regulate the conversational flow: based on the dialogue history and both SPs, it decides whether to maintain the current topic or initiate a new one.
    }
    \label{fig:sft_data}
\end{figure}
Our approach to Supervised Fine-tuning (SFT) relies on a two-pronged data strategy: large-scale automated synthesis and high-quality human authoring.

For automated data generation, we constructed a synthesis pipeline that simulates realistic conversations between an AI companion and a human user. The process is coordinated by three specialized components: an \textbf{AI Companion Agent} responsible for producing the assistant’s responses, a \textbf{Director Agent} that manages topic transitions and guides the conversation toward a coherent and natural conclusion, and a \textbf{User Agent} that generates human-like replies. To maintain both diversity and quality, the pipeline employs meta-prompts that dynamically construct distinct system instructions for the AI and User models in each dialogue instance. A detailed schematic of this workflow is illustrated in Figure \ref{fig:sft_data}.

Recognizing the limitations of synthetic data in capturing authentic human nuance, we supplemented our dataset with conversations authored by human annotators. In this process, we pair annotators and instruct them to engage in natural conversations over an extended period of 4-5 days. Annotators were given high-level topics as optional conversation starters. This longitudinal approach allowed us to collect a rich dataset that covers different stages of interpersonal communication, from initial ice-breaking to the more developed relationship and familiarity of later interactions.

Following data collection, both the synthesized and human-authored datasets underwent a meticulous manual refinement and curation process. For the synthetic data, human curators focused on correcting logical inconsistencies and improving conversational coherence. They also revised any phrasing that appeared robotic or unnatural to better emulate human expression. For the human-authored data, the refinement process involved correcting typographical errors, removing verbal tics, and, where necessary, adjusting the sequence of conversational turns to enhance logical flow. This comprehensive curation stage is critical to ensuring the final SFT dataset is clean, coherent, and of the highest possible quality.

\subsection{Humanlike Reward Model Training Data} \label{sec:humanlike_data}
We formulate the human–machine expression discrimination task as a Turing test–style judgment problem, where the reward model is trained to determine whether a given utterance—either standalone or context-dependent—is produced by a human or an LLM. To support this objective, we construct a fine-grained training mixture derived from both human-labeled and model-generated samples used in the SFT stage, with additional modifications to improve robustness and mitigate reward hacking. The dataset consists of three complementary components:
\begin{enumerate}
    \item \textbf{Context-free data:} This subset includes isolated utterances without any conversational context. Each sample is associated with an unambiguous ground-truth label inherited from the SFT corpus: human annotator outputs are labeled as human, while model-generated responses are labeled as machine. This component enables the reward model to capture surface-level linguistic and stylistic features indicative of humanlike expressions.
    \item \textbf{Context-based data:} To incorporate contextual dependencies, we construct samples by randomly replacing the final assistant turn in human-annotated dialogues with an LLM-generated response. This setup encourages the reward model to evaluate contextual coherence and semantic consistency rather than relying solely on local fluency. Empirically, training without context leads to overfitting on stylistic cues and results in semantically inconsistent yet “humanlike” outputs, which is a typical form of reward hacking.
    \item \textbf{Shuffled-context-based data:} To further enhance robustness, we apply a context shuffling augmentation, that is, the final user turn of dialogue $A$ is swapped with that of dialogue $B$, while both assistant responses remain fixed. This operation disrupts the natural dialogue flow, forcing the model to rely on coherence between turns instead of memorized surface patterns. It effectively improves generalization and reduces the risk of spurious correlations.
\end{enumerate}
Together, these data configurations allow the Humanlike Reward Model to jointly learn from linguistic cues as well as contextual dependencies, thereby enabling more reliable discrimination between human and LLM-generated utterances. 

\subsection{Empathetic Reward Model Training Data}
\begin{figure}[H]
    \centering
    \includegraphics[width=\linewidth]{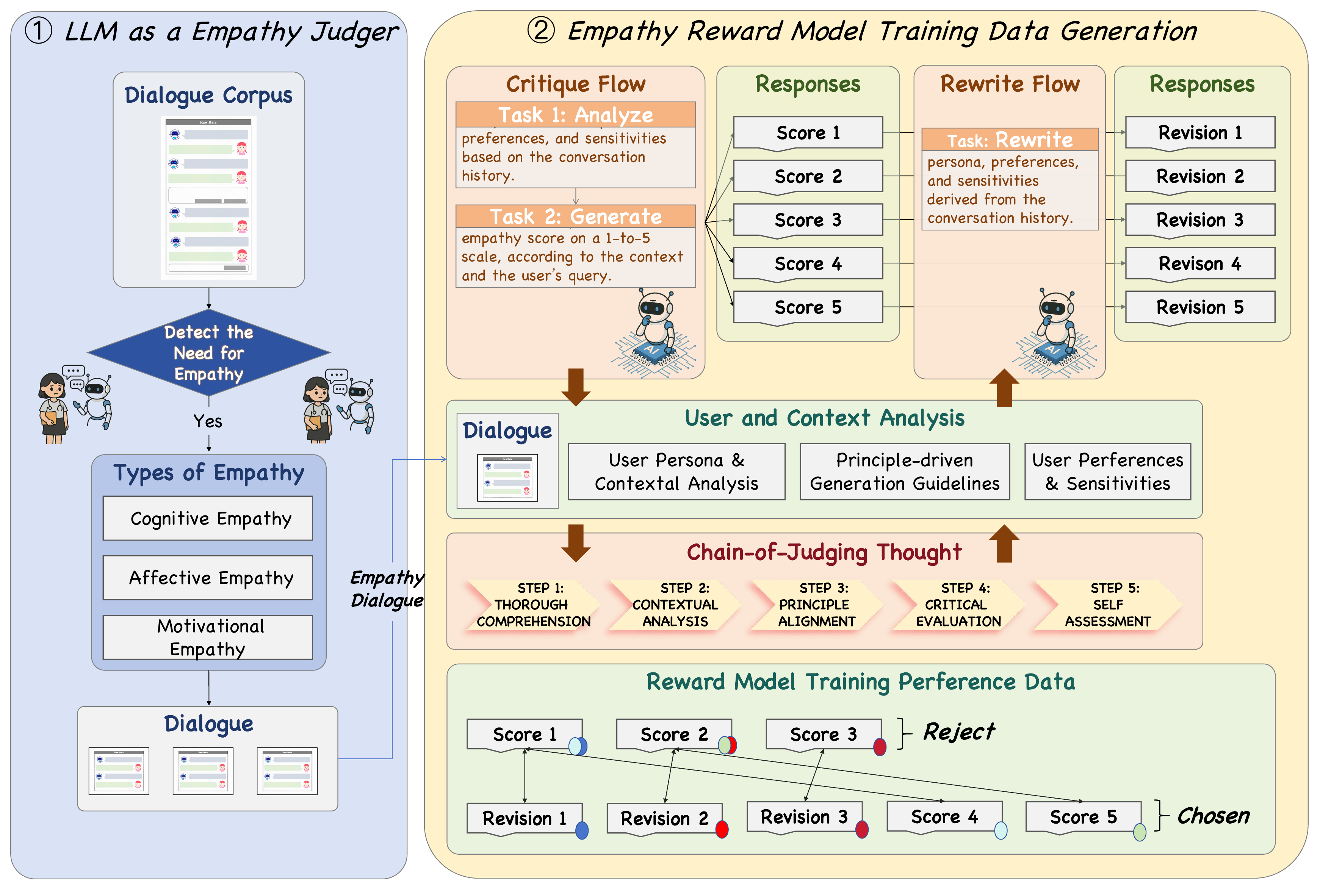}
    \caption{\textbf{The overall pipeline of reward model training data.} The process begins by filtering human annotated dialogues to isolate contextually relevant and empathy-requiring scenarios while excluding unsafe content. Subsequently, a principle-driven Critique-Rewrite framework analyzes user personas to generate graded responses and iteratively refines suboptimal outputs into golden versions. These high-quality responses are finally paired with lower-scoring candidates to construct the preference dataset.
    }
    \label{fig:rm_data}
\end{figure}
Our Empathetic Reward Model training data consists of a large corpus of human-annotated dialogues that we carefully filtered at a fine-grained level. Using carefully crafted prompts, we employed an LLM-as-a-Judge schema to identify samples that require empathy. 
This filtering system operates in two stages: First, we filter out irrelevant scenarios by removing noisy data and extraneous features commonly present in labeled conversations, ensuring only contextually relevant dialogues remain. Second, we detect empathy-requiring scenarios through a diagnostic system that performs feature extraction and classification to identify conversations where empathetic responses are most needed. The empathy features and scenarios extracted from this process are subsequently applied to guide the Reward Model data generation phase.
Moreover, because sexual-content scenarios can sometimes be confused with empathetic ones, we devised a Content Safety Constitution for filtering to remove contaminated samples.

To generate responses that achieve an optimal score (5/Excellent) across dimensions of strategy, language, and persona consistency, we construct a comprehensive, principle-driven framework. Specifically, we devise a Critique-Rewrite data generation pipeline, similar to that in \cite{bai2022constitutionalaiharmlessnessai}.The prompt dataset consists of multi-turn conversational contexts, where each sample is defined by a dialogue history between the user and the assistant, along with the user’s final request. The Critique-Rewrite is tasked with generating responses to this final turn that are aligned with the inferred user preferences, conditioned on the preceding conversational context.

In the Critique phase, the model functions as a strict empathetic judger. It first discerns the user's underlying intent and needs. It then proceeds to generate a range of responses, graded from the least suboptimal (1-point) to excellent (5-point), with each response accompanied by a detailed rationale. More concretely, the model distills and summarizes key information into three components: (1) \textbf{User Persona and Contextual Analysis}, which captures the user's profile and situation; (2) \textbf{User Preferences and Sensitivities}, which identifies desirable approaches and potential red lines; and (3) \textbf{Principle-driven Generation Guidelines}, which outlines the criteria for an optimal response derived from a set of core empathetic principles. Concurrently, the model produces five distinct responses that it self-evaluates with scores ranging from 1 to 5. This process is carefully calibrated to ensure that the analysis is grounded in the user’s perspective. In our domain, unlike helpfulness or harmfulness, there is no universal standard for what constitutes a good empathetic response. The same reply might be perceived as overly forward or boundary-crossing by an avoidant user, yet feel perfectly natural and engaging to someone more expressive or enthusiastic. Given the inherently subjective nature of empathy, it is fundamentally impossible to fully inhabit another person’s perspective. Consequently, if human annotators were tasked with evaluating such responses, their judgments would inevitably drift toward reflecting their own preferences rather than the user’s. To mitigate this bias, we instead defined a set of meta-empathetic principles and leverage state-of-the-art models’ capacity to dynamically infer user personas and preferences, allowing the system to generate and assess responses from the inferred user’s point of view. All results produced in the Critique phas7e are subsequently verified by the corresponding labeler, the individual who initially annotated the data. This closed-loop process ensures the accurate capture of the user's empathetic needs.

In the Rewrite phase, the model first condenses the insights from the Critique phase into a set of core guiding principles. It then engages in an iterative refinement process to elevate the generated responses. This process involves restructuring the content, infusing a consistent persona, and polishing the linguistic style. The outcome is a collection of exemplary 5-point responses in diverse styles, culminating in a single, definitive "golden" version. The efficacy of this Rewrite phase is fundamentally predicated on the comprehensive nature of the three analytical components generated during the Critique phase. We tested several models for the Critique-Rewrite routine, including GPT-4\cite{openai2024gpt4technicalreport}, Claude-Sonnet-4.5 and Gemini-2.5-pro\cite{comanici2025gemini25pushingfrontier}, and the best candidate we found for our task is Gemini-2.5-pro. Nevertheless, even with the use of state-of-the-art models, we identified a recurring issue of performance degradation during the Rewrite phase. When refining responses initially rated as 4 or 5 points, the model frequently exhibited a tendency toward over-revision—alterations that were intended to improve quality often resulted in diminished coherence, empathy, or stylistic fidelity. Consequently, the revised outputs were occasionally downgraded to a 3-point level. To mitigate this effect, we adopted a conservative cutoff strategy: only responses rated 3 points or lower were included in the Rewrite phase, while higher-rated responses were retained without modification.

Finally, we construct the training dataset for our empathetic reward model by creating preference pairs. These pairs are formed by combining the definitive "golden" responses from the Rewrite phase with the lower-scoring responses generated during the Critique phase, as shown in Figure \ref{fig:rm_data}.

\subsection{RL Training Data}
We construct reinforcement learning (RL) data for two primary domains: empathetic dialogue and daily chit-chat. Both domains are inherently non-verifiable, as their responses cannot be evaluated against a single ground-truth answer. However, empathetic dialogue poses a greater challenge, since the quality of an empathetic response is entirely subjective—it depends on how well the model captures and aligns with the user’s emotional state and contextual preferences.

To address this challenge, we introduce Gemini-2.5-pro as a reference model within the Critique–Rewrite pipeline. For each empathetic instance, Gemini generates responses that are internally scored, and we select those rated at the 3-point level as reference anchors. These references serve as fixed baselines during RL training: the policy model receives a positive reward (+1) only when its rollout produces a response that surpasses the reference in quality.

In contrast, daily chit-chat data are trained without the Critique–Rewrite procedure. Instead, we apply a human-likeness reward that encourages natural, coherent, and contextually appropriate conversational behavior. The final RL dataset is the union of empathy-rewarded and human-likeness-rewarded samples. This hybrid setup provides complementary learning signals—anchoring the model’s empathy through reference-based supervision while maintaining general conversational fluency—and effectively mitigates potential out-of-distribution (OOD) drift.

To filter out low-quality reference answers in empathetic dialogue, we adopt a data filtering strategy adapted from seed 1.5\cite{seed1.5}. To be specific, we use the policy model intended for RL training to perform Best-of-N sampling. Subsequently, we use a pairwise reward model to compare the policy model's outputs against the Gemini-generated reference answer. Instances in which the policy model achieves high scores with low standard deviation are discarded. This approach removes overly simplistic samples, ensuring that our training data focuses on difficult cases where a high-quality reference provides substantial learning value.

\section{Method}
\subsection{Supervised Fine-tuning}
The Supervised Fine-tuning (SFT) stage is designed not only as a bridge between pretraining and reinforcement learning, but also as a crucial step toward aligning the model with one of the ultimate forms of Human–AI Interaction—AI companionship. In this domain, models are expected to go beyond general helpfulness and instruction-following: they must communicate with humanlike naturalness, display emotional sensitivity, and often exceed human empathy in understanding and responding to affective cues.

To meet these requirements, our SFT process aims to build a model that performs strongly in vertical domains, such as AI companionship and role-playing, while maintaining broad general capabilities. A common challenge in domain specialization is the degradation of general-purpose skills due to catastrophic forgetting. To mitigate this, we conduct extensive data composition experiments balancing domain-specific and general data sources. 

\begin{figure}[tb]
    \centering
    \includegraphics[width=\linewidth]{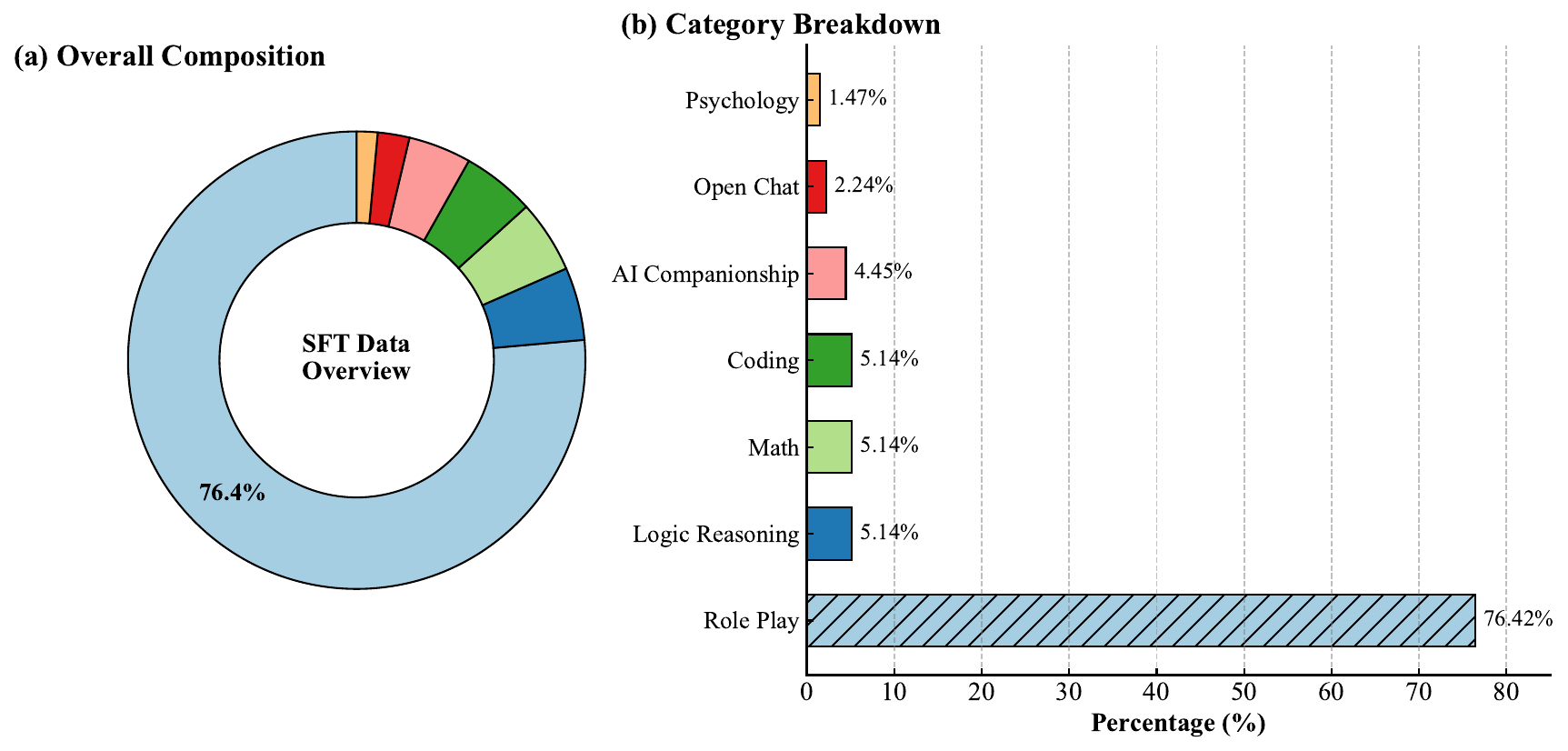}
    \caption{Illustration of the SFT dataset composition. The dataset integrates our proprietary AI companionship data with several open-source datasets, balancing domain specialization and general coverage.}
    \label{fig:sft_data_mix}
\end{figure}

The domain-specific data, introduced in Section \ref{sec:sft-data}, include large-scale companion-style dialogues and role-playing interactions, designed to enhance natural expressiveness, topic control, and emotional perception. The general data consist of open-source instruction-following, commonsense reasoning, mathematics, and code datasets \cite{Firefly, bai2024coig, zhang2023chinese, xu2023cvalues}, incorporated to preserve the model’s fundamental reasoning and task-solving abilities. During fine-tuning, we proportionally mix these data sources to maintain a balance between emotional depth and general competence. Our data composition is illustrated in Figure \ref{fig:sft_data_mix}.

We conducted SFT on top of the Qwen3-32B model \cite{qwen3technicalreport} to adapt it to our target dialogue domain. The model was trained for 4 epochs with a batch size of 128 using the AdamW optimizer. We set the learning rate to 1e-5, applied a cosine decay schedule, and used a warm-up ratio of 0.1. This configuration provided stable optimization and effective adaptation while retaining the base model’s general conversational abilities. Throughout training, to improve stability and learning efficiency, we adopt a curriculum that progresses from short, neutral exchanges to longer, emotion-rich conversations. This staged exposure helps the model gradually acquire contextual tracking and empathetic phrasing under controlled difficulty, reducing early divergence and promoting robust generalization.

In summary, SFT builds the linguistic and emotional foundation required for subsequent reinforcement learning, enabling the policy model to start with strong empathy and natural conversational fluency.

\subsection{Reward Models}
As mentioned before, the task of developing an AI companion presents unique challenges not found in objectively verifiable domains like mathematical reasoning or code synthesis. Unlike these tasks, where a clear ground truth or a deterministic verifier can provide an unambiguous reward signal, AI companionship is fundamentally subjective. The quality of an interaction is contingent upon individual user preferences, which are inherently diverse and personalized. A key challenge arises from this heterogeneity of preferences: a response that one user finds engaging and empathetic may be considered intrusive or inappropriate by another. This precludes the establishment of a single, population-averaged preference criterion to guide model training effectively.

To navigate this challenge, prior works have predominantly relied on methods such as Reinforcement Learning from Human Feedback (RLHF)\cite{ouyang2022traininglanguagemodelsfollow} and Reinforcement Learning from AI Feedback (RLAIF)\cite{bai2022traininghelpfulharmlessassistant}. In RLHF, a reward model is trained to output a scalar score representing human preference. However, this approach has two principal drawbacks. First, reducing complex, multi-faceted human preferences to a single scalar value fails to utilize the nuanced reasoning and inferential capabilities of modern Large Language Models (LLMs). Second, scalar reward models are notoriously prone to "reward hacking," where the policy model learns to maximize the reward score in ways that do not align with the true, underlying user preferences.

Alternatively, the LLM-as-a-judge\cite{zheng2023judgingllmasajudgemtbenchchatbot} paradigm, often employed in RLAIF, leverages the advanced reasoning of state-of-the-art LLMs to evaluate responses. While this approach better captures the complexity of user preferences, it introduces a critical dependency on external models. The efficacy of an LLM-as-a-judge is highly sensitive to the specific phrasing of the judging prompt. Moreover, as external LLM providers continuously update their models, a previously effective prompt can become obsolete, leading to instability and requiring a costly and iterative process of prompt re-engineering.

To overcome these obstacles, we follow a recent approach: the development of a proprietary generative reward model\cite{mahan2024generativerewardmodels, liu2025inferencetimescalinggeneralistreward, whitehouse2025j1incentivizingthinkingllmasajudge}. This method is designed to fully leverage the reasoning capacity of LLMs while providing a stable, controllable, and nuanced reward signal that is independent of external model iterations, thereby offering a more robust solution for optimizing personalized AI companions. 

\subsubsection{Humanlike Reward} \label{sec: humanlike_reward}
The goal of humanlike reward is to provide a humanlike signal that helps the model learn to talk more like a human rather than a machine. Let $\mathcal{S}_{\text{human}}$ be the space of all human spoken languages and $\mathcal{S}_{\text{machine}}$ be the space of all machine generated languages, the humanlike reward is defined as: 
\begin{equation}
r_{\text{humanlike}} = \mathbb{I}\!\left[p_{\psi}(y \in \mathcal{S}_{\text{human}}) > p_{\psi}(y \in \mathcal{S}_{\text{machine}})\right]  \label{eq:humanlike_reward_base}
\end{equation}
where $p_{\psi}$ is any judge function that takes natural language $y$ as input.

While one may propose to use LLM-as-a-judge by acquiring SOTA models, we found that it is hard for SOTA LLMs to do human-tone classification. Specifically, given a sentence, the model is asked to tell whether the sentence is from a real human or a machined generated one. With a moderate prompt engineering effort, the SOTA models failed to complete the task. As we use this as an external reward signal to train our policy model, we observe severe reward hackings, even the reward keeps going up, the policy model continues to generate strange outputs that deceives the LLM judger, shown in \textbf{Appendix} \ref{box:humanlike_failure_cases_sota_modle}. We therefore collected these hacked outputs as our hand-crafted held-out hard-negatives and tested several SOTA models. The detailed experimental analysis will be discussed in Section \ref{sec: exp_humanlike_reward} and the test prompt we used is included in \textbf{Appendix} \ref{appendix: humanlike_judger_prompt}. 

Actually, this is not very surprising to us that SOTAs behave poorly on humanlike judgment tasks, because most models are optimized for helpfulness and harmlessness. Humanlike expression is not the targeted goal, so this is out of the scope of SOTAs. This strongly indicates the necessity of training our own humanlike judgement model. To leverage the chain-of-though (CoT)\cite{wei2023chainofthoughtpromptingelicitsreasoning} ability of LLMs, we treat our humanlike judgement model as another LLM instead of a scalar preference model, just as genRM\cite{mahan2024generativerewardmodels, liu2025inferencetimescalinggeneralistreward, whitehouse2025j1incentivizingthinkingllmasajudge}. That is, Equation \ref{eq:humanlike_reward_base} can be rewritten as
\begin{equation}
    r_{\text{humanlike}} = \mathbb{I}\!\left [ p_{\psi}(l=\text{human}, c \mid x) > p_{\psi}(l=\text{machine}, c \mid x)\right ] \label{eq:humanlike_reward_middle}
\end{equation}
where $l$, $c$ represent the predicted label and CoT respectively, and $p_{\psi}$ in this case is our trained judger LLM, $x$ in this case is some input, depends on wether the judger is context free or not, $x$ could be defined differently.

In order to train the humanlike judger, we propose two approaches, which are context free and context aware. For context free judger, we have
\begin{equation}
r_{\text{humanlike}} = \mathbb{I}\!\left [ p_{\psi}(l=\text{human}, c\mid x) > p_{\psi}(l=\text{machine}, c \mid x)\right ] \label{eq:humanlike_reward_contextfree}
\end{equation}

In contrast, for context aware judger, we have
\begin{equation}
r_{\text{humanlike}} = \mathbb{I}\!\left [ p_{\psi}(l=\text{human}, c\mid \underbrace{y_T, h}_{x}) > p_{\psi}(l=\text{machine}, c \mid \underbrace{y_T, h}_{x})\right ] \label{eq:humanlike_reward_context}
\end{equation}
where $y_T$ is the $T$-th round's response, and $h = \left(x_1, y_1, x_2, y_2,\dots,x_{T-1},y_{T-1} \right)$ is the history of last $T-1$ rounds.

For context-free judger, our goal is to let the model distinguish only the expression itself is from human or AI generated regardless the context. However, one drawback in context-free judger is that it may over emphasize the expression superficailly and ignore the in-context conversational logic. Thus, intuitively, context aware judger should be more coherent, but it may lose some flexibility compared to the context-free judger. 

\subsubsection{Empathy Reward} \label{sec: method_empathy_reward}
Unlike rewards for human-likeness, which primarily focus on the surface-level fluidity and naturalness of expression, the empathy reward is designed to capture the model's proficiency in underlying empathetic capabilities. A positive reward signal is contingent on the model demonstrating genuine and effective empathy, rather than merely mimicking anthropomorphic expressions.

The design of this reward mechanism must address the highly personalized nature of empathetic needs, which vary significantly across individuals. We posit that a user's preference for empathy is highly volatile and fluid, intrinsically linked to their recent life events, current health status, and transient emotional state. However, such preferences cannot be reliably elicited through explicit self-reports—whether by directly asking users about their likes and dislikes or by relying on predefined questionnaires—because users often lack stable or accurate introspection about their own empathetic needs. For instance, a user may claim to prefer highly enthusiastic and emotionally expressive responses, yet during periods of substantial stress, the same user might instead favor a calm, pragmatic assistant that offers actionable guidance. The mismatch between stated preference and momentary need introduces systematic false-positive signals for the reward model when supervision is derived solely from the user's conscious self-assessment.

To address this gap, we introduce \textbf{User Context Mining}, a principled approach that infers the user's latent and temporally local empathetic preference directly from their recent interaction patterns with the LLM, rather than relying on potentially unreliable self-descriptions. This enables the reward mechanism to adapt to the user's actual, dynamically evolving needs. Such volatility makes static rules or fixed heuristics fundamentally insufficient for reliable reward assessment. Consequently, empathy evaluation must be treated as a continuous and dynamic inference process. To achieve this, the reward function leverages the LLM's CoT capabilities~\cite{wei2023chainofthoughtpromptingelicitsreasoning} through a two-stage dynamic inference procedure: \textbf{dynamic profile inference} followed by \textbf{alignment inference}. Dynamic profile inference requires the model to infer a user profile or persona by analyzing the preceding conversational context; based on this inferred profile, the model then determines whether its proposed empathetic strategy aligns with the user's immediate and specific needs, which we term alignment inference.

Nevertheless, we can still establish a set of universally representative meta-principles. These principles are not rigid rules, but rather high-level guidelines. They instruct the model on (1) how to effectively reason about the user profile based on historical context, and (2) how to leverage that profile to deliver tailored, personalized empathy—effectively achieving a “thousand-people, thousand-faces” standard of interaction.

Let $\mathcal{S}_{\text{empathy}}^{x_T, h}$ be the set of all ideal empathetic responses given the last-round query $x_T$ and chat history $h$, where $y$ is the model-generated response and $r$ is the reference answer. The empathy reward is defined as:
\begin{equation}
r_{\text{empathy}} = 
\mathbb{I}\!\left[
p_{\phi}(y \in \mathcal{S}_{\text{empathy}}^{x_T, h} \mid x_T, h)
>
p_{\phi}(r \in \mathcal{S}_{\text{empathy}}^{x_T, h} \mid x_T, h)
\right],
\label{eq:empathy_reward_base}
\end{equation}
where $p_\phi$ is our trained generative empathetic judge. This reward computation method is perfectly aligned with the training task of our Empathy Reward Model, effectively harnessing its full capabilities. Crucially, in the AI companionship setting—where a single “correct” answer does not exist—the use of high-scoring reference answers guides the optimization process and allows the model to better incorporate human preferences. We explore this concept using WorldPM~\cite{wang2025worldpmscalinghumanpreference}, a scalar reward model, and in Section~\ref{sec: experiment} provide details that demonstrate the necessity of the reference answer.

\paragraph{Two-Stage Training Framework}
To construct a scalable empathetic judge, we adopt a two-stage training framework designed to support \textbf{Weak-to-Strong} generalization~\cite{burns2023weaktostronggeneralizationelicitingstrong}, where a smaller model bootstraps a larger one through iterative data refinement. Instead of following a conventional knowledge distillation paradigm, we explicitly redefine the small-parameter model as a high-throughput \textbf{Reasoning Path Sampler} and use its trajectories to drive model scaling and iterative data evolution.
\begin{itemize}
    \item \textbf{Stage I}: We use \texttt{Qwen3-8B} as the base model. We begin by training a preliminary generator through a one-epoch supervised fine-tuning (SFT) cold start on the seed data, and then perform DAPO alignment training. To ensure that this model is reliable as a sampler, we conduct an A/B option permutation consistency test under the Pass@1 setting to reduce positional bias, confirming its basic capability to provide valid candidate samples. We then employ this 8B model for large-scale rejection sampling: under strict discrimination criteria (introduced later), trajectories with flawed reasoning logic or mismatched answers are filtered out, while only logically closed-loop “golden” reasoning paths are retained. Combined with targeted human review and lightweight AI-assisted refinement to correct stylistic artifacts, these curated trajectories form a high-quality dataset, Dataset V2, whose reasoning quality can surpass the inherent capability ceiling of the small model.
    \item \textbf{Stage II}: To implement the \textbf{Weak-to-Strong} generalization objective~\cite{burns2023weaktostronggeneralizationelicitingstrong}, we upgrade the base model to \texttt{Qwen3-32B} and perform an SFT cold start(1 epoch) on Dataset V2, followed by DAPO training. To mitigate the distributional limitations of SFT data and enhance robustness in unseen domains, we further introduce a \textbf{Recursive Self-Correction} mechanism into the system prompt during the RL training phase. This instruction, activated only during RL exploration, encourages implicit secondary reasoning and logical backtracking before the model commits to a final verdict, thereby improving decision reliability in complex empathetic scenarios.
\end{itemize}

This two-stage pipeline yields two empathetic judgers—a lightweight 8B model and a stronger 32B model—which are evaluated in Section~\ref{sec: exp_empathy_reward}. Additionally, the model was trained for one epochs with a batch size of 64 using AdamW, a learning rate of $1\times10^{-6}$ with 300 warm-up steps followed by cosine decay, low/high clipping thresholds of $3\times10^{-4}$ and $4\times10^{-4}$, a KL loss coefficient of 0.001, eight samples per prompt, a response length cap of 4K tokens, and a length penalty of 0.1.

\paragraph{Reward Function}
In optimizing GenRMs, the reward function design directly controls the direction and quality of the gradients. Unlike traditional linearly weighted reward compositions, we construct a \textbf{discrete multiplicative reward mechanism} based on logical ``AND'' gating. For a given input prompt $x$ and a model-generated response $y$ containing the reasoning process $r$ and the final answer $a$, the empathy reward used during RL is defined as:
\begin{equation}
    R_{\text{empathy}}(x, y) = R_{\text{process}}(r) \cdot r_{\text{empathy}},
    \label{eq:final_empathy_reward_function}
\end{equation}
where:
\begin{itemize}
    \item $r_{\text{empathy}}$ acts as a hard constraint, ensuring that any reasoning path leading to an incorrect or misaligned outcome—no matter how well written—receives a zero score. This effectively prevents ``reward hacking'' behaviors such as guessing answers or optimizing only for superficial style.
    \item $R_{\text{process}}(r) \in [0, 1]$ represents \textbf{process quality and format compliance}. It is a normalized coefficient jointly determined by the recall rate of key reasoning steps and C/I (Correctness/Instruction) constraints.
\end{itemize}

The core motivation for the multiplicative form in Equation~\ref{eq:final_empathy_reward_function} is to approximate \textbf{logical entailment}: we effectively optimize the joint event $P(\text{correct reasoning} \cap \text{correct outcome})$, rather than a linear superposition of their marginal contributions. This sparse yet high-precision signal, combined with the DAPO algorithm, allows the model to anchor onto genuinely valid reasoning patterns within a vast search space.

\subsection{RL Training}
For non-verifiable problems, a robust and precise reward system is crucial for RL training. Specifically, the reward system should not only effectively prevent reward hacking, but also its reward signals must be tightly aligned with the intended training objectives. To this end, based on our empathetic companionship task, we propose a fine-grained reward framework.

Our reward framework consists of two components: Empathy Reward and HumanLike Reward, where the total reward signal is demonstrated in Equation. Empathy Reward guides the model to produce highly empathetic responses, while HumanLike Reward encourages the model to adopt a captivating, human-like conversational style. By combining these components, we aim to train a model that both fully satisfies users' empathetic needs and delivers a conversational experience comparable to chatting with a real person.
\begin{equation}
    R_{\text{total}} = R_{\text{empathy}} + R_{\text{humanlike}}
\end{equation}

We compute the Empathy Reward and Humanlike Reward using the generative reward model described in Section \ref{sec: method_empathy_reward} and Section \ref{sec: humanlike_reward} respectively. The whole training pipeline, including reward model training, is build upon the VeRL\cite{sheng2024hybridflow} framework. To simplify training, we deploy our generative reward models as services, and request the reward signal once a single rollout is fully generated.

\section{Evaluation Framework}
\begin{figure}[tb]
    \centering
    \includegraphics[width=0.9\linewidth]{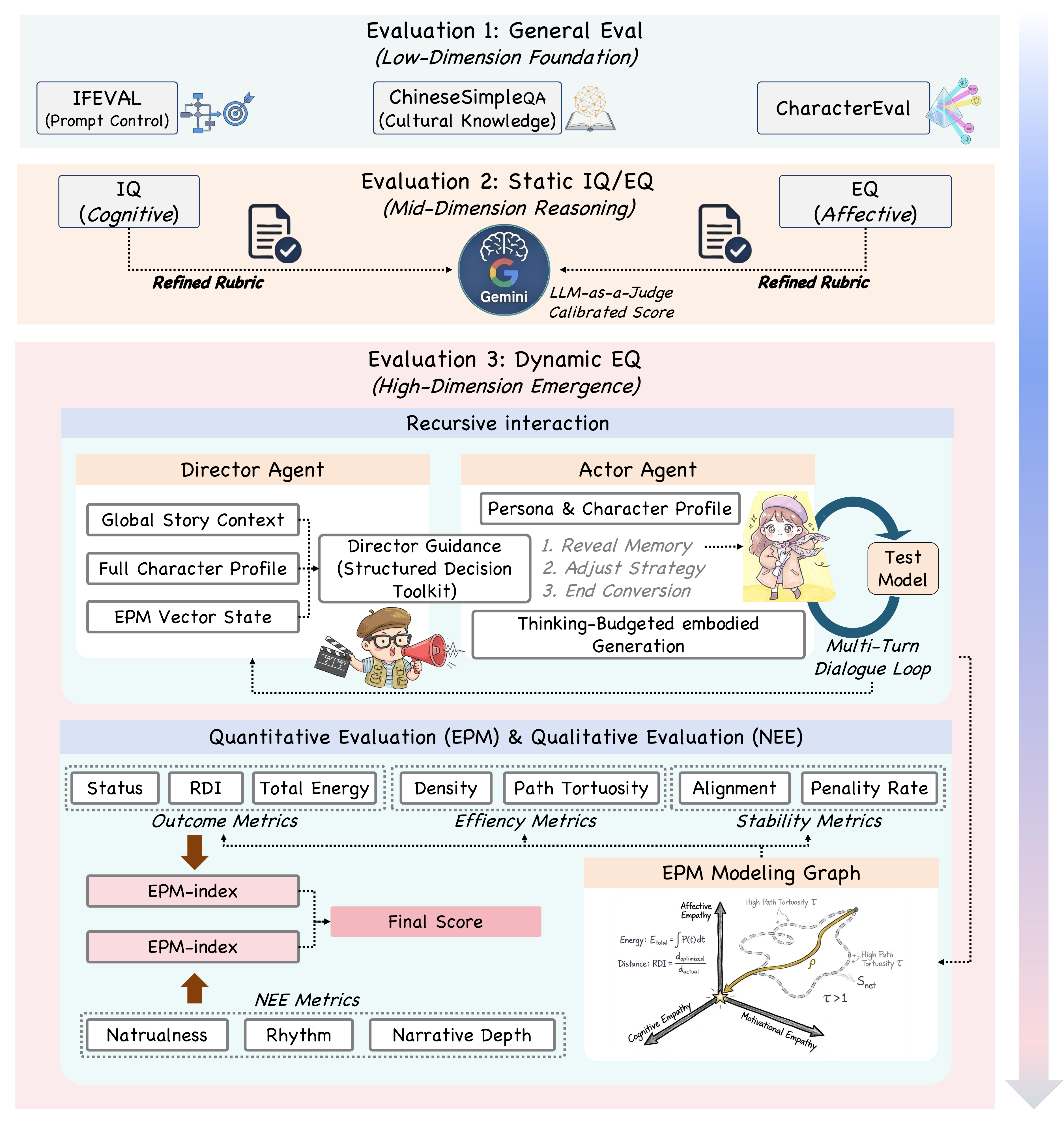}
    \caption{Comprehensive AI companionship evaluation pipeline}
    \label{fig:eval_pipeline}
\end{figure}

Our evaluation pipeline is structured into three distinct stages. First (Stage 1), we assess the model's foundational capabilities using publicly available benchmarks to establish a general performance baseline. Second (Stage 2), we evaluate its basic cognitive and emotional intelligence (IQ and EQ) through proprietary, static benchmarks. These benchmarks utilize multi-turn dialogues with fixed contexts, requiring the model to respond only to the final turn. Finally (Stage 3), we observed that models could exploit these static contexts, which often act as cues for stylistically similar responses. To mitigate this, we developed a dynamic EQ framework where no context is provided, compelling the model to rely solely on its internal capabilities. The overall evalution pipeline is shown in Figure \ref{fig:eval_pipeline}.

\subsection{General Evaluation}
For the evaluation of the model's general-purpose capabilities, we established a benchmark suite specifically tailored to the requirements of our AI companionship scenario. We selected three distinct test sets: IFEVAL, ChineseSimpleQA, and CharacterEval. The rationale for this curated selection is as follows:
\begin{itemize}
    \item Instruction Following (IFEVAL)\cite{zhou2023instructionfollowingevaluationlargelanguage}: The model's ability to strictly adhere to instructions is critical for our application. We utilize IFEVAL to measure this capability, as it directly impacts our product team's ability to shape and control the AI's persona, tone, and behavior. A high instruction-following fidelity allows product managers to effectively implement target conversational effects merely by adjusting the character's system prompts.
    \item Chinese Contextual Knowledge (ChineseSimpleQA)\cite{he2024chinesesimpleqachinesefactuality}: To ensure the model is relatable and effective for our target audience, it must possess a broad and accurate base of general knowledge within a Chinese cultural context. We employ ChineseSimpleQA to assess this world knowledge, ensuring the model's responses are culturally relevant and align with the background of Chinese users.
    \item Human-like Interaction (CharacterEval)\cite{tu-etal-2024-charactereval}: While CharacterEval offers a multi-dimensional analysis, we deliberately focus our evaluation exclusively on its human-likeness dimension. For our emotional companionship use case, attributes like believable persona and natural interaction are paramount. We found other dimensions, such as Persona-Behavior or Know Hallucination to be less applicable and not aligned with our primary objective of creating an authentic, human-like companion.
\end{itemize}
This combined evaluation methodology allows us to holistically assess the model's fitness for our specific product goals, prioritizing practical prompt control, cultural resonance, and human-like interaction.

\subsection{Static IQ and EQ Evaluation}
Following the assessment of general capabilities, we conducted a more focused evaluation to probe the model's performance on core competencies directly relevant to our AI companionship domain. To this end, we constructed a static evaluation set designed to mirror the structure and complexity of authentic user interactions. 

This dataset is composed of multi-turn dialogues, each containing 15 rounds of conversational history. This history, meticulously curated and authored by human annotators, serves to establish deep context, emotional tone, and memory predicates. The model's task is to generate a response only for the final, 16th turn, which acts as the target test prompt. This benchmark consists of two distinct sets, an IQ test set and a EQ test set respectively. 

\subsubsection{Static Intelligence (IQ) Set}
This subset is designed to test fundamental cognitive and conversational mechanics that are essential for a coherent and believable interaction. The test prompts are specifically structured to evaluate the model's proficiency in:
\begin{itemize}
    \item Persona Consistency: The ability to correctly distinguish between speaker and listener (i.e., I vs. you) and maintain a consistent identity.
    \item Contextual Recall: The capacity to accurately access and utilize information provided within the 15-turn dialogue history.
    \item Temporal Awareness: The correct perception and handling of time-related concepts and conversational flow.
    \item Grounded Commonsense: The application of real-world knowledge within the specific constraints of the ongoing conversation.
\end{itemize}

\subsubsection{Static Empathy (EQ) Set}
This subset assesses the model's affective intelligence. For these test cases, the final (16th) turn is intentionally constructed to present a diverse array of scenarios requiring a nuanced, empathetic response, such as a user expressing distress, celebration, or vulnerability.

To score the model outputs from both the IQ and EQ sets, we employed an LLM-as-a-judge framework. We utilized Gemini 2.5 Pro as the core judging model. A critical aspect of our methodology involved the iterative refinement and “fine-tuning of the judger prompt—a highly-structured set of instructions and criteria. This prompt engineering was essential to ensure that the automated judgments are objective, reproducible, and consistently aligned with our human-defined rubrics for fairness and accuracy.

\subsection{Dynamic Empathy Evaluation}

\subsubsection{Overview}
Traditional empathetic dialogue evaluation typically rely on static single-turn datasets or simplistic multi-turn prompts, failing to capture the dynamic, long-range nature of deep emotional interactions\cite{chen2024tombenchbenchmarkingtheorymind,Zhang_2025_04,Hu2025_2502_04424v2}. To address this limitation, we propose the Anthropomorphic Empathy Evaluation Framework. This framework is designed to provide an objective, rigorous, and physically interpretable method for quantitatively assessing the comprehensive performance of empathetic dialogue models.
The framework consists of two interdependent core layers:
\begin{itemize}
\item \textbf{Simulation Layer (Anthropomorphic Cognitive Sandbox):} Serving as the simulation environment, this layer provides a high-fidelity social interaction arena designed to reconstruct complex human emotional and cognitive dynamics, offering an evolutionary ground for AI beyond scripted interactions\cite{park2023generativeagentsinteractivesimulacra,zhou2024sotopiainteractiveevaluationsocial}.

\item \textbf{Metric Layer (Empathy Physics Model, EPM):} Serving as the measurement layer, this is a novel psychophysical cognitive modeling approach. It translates abstract psychological empathy into computable physical quantities—specifically, modeling the complex processes of psychological healing and companionship as a process of doing work against emotional resistance within a high-dimensional vector space\cite{Schurz_2021_03,park2023generativeagentsinteractivesimulacra}.
\end{itemize}
By combining dynamic dialogue simulation with rigorous psychophysical modeling, this framework offers an interpretable evaluation method with high ecological validity\cite{Naous2025_2510_06552v1,zhou2024sotopiainteractiveevaluationsocial}. It assesses not only what the model says but also the trajectory and efficacy of its intervention over time. Visualizing interaction trajectories allows us to intuitively identify strategic preferences (e.g., leaning towards emotional soothing vs. cognitive restructuring)\cite{zhuge2025agentasajudge,Chen_2026_03}, and analyze characteristic patterns in failure scenarios (like repetitive looping or disorientation), providing concrete diagnostics for model training iteration.

\subsubsection{Core Mechanism Design}
\textbf{(1) Simulation Layer: Multi-Agent Cognitive Architecture and Anti-Collusion}

To mitigate the inherent verbal biases and potential collusion risks of LLM self-evaluation —limitations that remain prevalent in frameworks like SAGE\cite{Zhang_2025_04}, where the conflation of dialogue generation and scoring roles creates structural vulnerabilities—our simulation layer incorporates a multi-agent cognitive architecture inspired by the human cognitive division of labor. By enforcing strict information isolation mechanisms, we ensure the authenticity and objectivity of the interactions.
\begin{itemize}
\item \textbf{Director Agent (Holistic Cognitive Orchestration): }Simulating high-level executive control, this agent does not participate in dialogue directly. Instead, it dynamically orchestrates the simulation based on an omniscient perspective (plot, persona, real-time EPM state) via a structured Function Calling mechanism. It simulates complex human conversational thinking, not by following a script, but by reasoning dynamically based on dialogue progress. It wields a rich toolbox to retrieve memories, propel the plot, or adjust the Actor's strategy based on EPM feedback (e.g., instructing stronger defensiveness). This mechanism endows the sandbox with high dynamism and fidelity, making every dialogue a unique evolutionary process.

\item \textbf{Actor Agent (Situated Dynamic Acting): }Simulating emotional experience and immediate reaction regions. To prevent assistant-like behavior and adjudication collusion, the Actor operates under strict evaluation isolation (knowing only current settings, unaware of evaluation criteria). It relies on a complex user generation and simulation pipeline that extracts and combines deep features for diverse personas, needs, and open-ended scenarios\cite{park2023generativeagentsinteractivesimulacra,zhou2024sotopiainteractiveevaluationsocial,wang2025worldpmscalinghumanpreference,Naous2025_2510_06552v1,2025Goal}. For the test library, we generated over 500 multifaceted case profiles by drawing upon internal research insights and analyzing the target demographic's empathetic need characteristics. These cases cover diverse personas, psychologies, layered life experiences, and specific scenarios with potential storylines. Notably, dynamic content elements (e.g., critical memories, plot twists) are held latent, pending invocation by the Director for real-time enactment—a mechanism that maximizes unpredictability and realism.
\end{itemize}

\textbf{(2) Metric Layer: EPM Vector Space and Evidence-Anchored Evaluation}

EPM upgrades empathy evaluation from static feature matching to dynamic computation by introducing the physical metaphors of energy and work.\cite{Hu2025_2502_04424v2,Park_2023_01,Shenhav_2024_12} In this framework, a user's psychological distress is no longer an abstract concept but is concretized as resistance or a potential energy trap. The empathy model's response is modeled as an applied force on the user's psyche. If the direction of this force aligns with the user's deep-seated needs (i.e., directional alignment), it can effectively propel the user to overcome psychological resistance and move towards a state of equilibrium. This effective propulsion process is defined as doing work, and its cumulative amount is the effective energy. EPM thus translates invisible social signals into visible, interpretable physical trajectories and energy curves, allowing us to intuitively assess whether an intervention is highly efficient (half the work, double the effect) or ineffective (heading south to go north).

Specifically, we map the empathetic interaction into an orthogonalized MDEP three-dimensional psychological measurement space (C-axis: Cognitive Restructuring, A-axis: Affective Resonance, P-axis: Proactive Empowerment)\cite{Shamay-Tsoory_2009_03,Weisz_2020_10}. The origin represents the ideal state of psychological balance, while the user's initial state is quantified as a negative deficit vector. Each model reply is viewed as an action vector ($\vec{v}_t$) applied to the user's psychological state, containing independent bidirectional scoring of positive progress (Prog) and negative regression (Neg) across all three CAP axes. The dialogue process is modeled as the dynamic evolution and accumulation of the user's psychological state trajectory towards the ideal origin under the continuous action of these discursive forces.

To address the core challenge of instability when using LLMs as judges, we draw on RLHF reward modeling approaches\cite{ouyang2022traininglanguagemodelsfollow,Hu2025_2502_04424v2,chen2024tombenchbenchmarkingtheorymind}to design a rigorous evidence-anchored bi-directional evaluation system. Adopting the LLM-as-Judger approach\cite{zhang2023chinese,zhuge2025agentasajudge,2025Agent}, we selected Gemini 2.5 Pro as the judge, given its SOTA performance in user profile and complex EQ comprehension based on extensive benchmarks. It forces the reduction of the LLM's task from subjective scoring to objective qualitative classification—first exhaustively extracting progress/regression evidence, then matching evidence against predefined Behavioral Anchors for classification\cite{Rohit_2024_05}, and finally mapping to numerical values through deterministic rules. This ensures the evaluation process is based on objective evidence rather than subjective preference.

To achieve deep alignment with the specific user scenario, our framework rigorously controls both the initialization of the deficit and the dialogue scoring. The initial deficit is not stochastically generated but is comprehensively derived from the user's personality, experiences, empathetic needs, and the narrative progression. Concurrently, the Judge agent evaluates replies dynamically relative to the immediate user profile. This focus on \textbf{profile alignment} ensures ecological validity, measuring the model's empathetic capacity toward a specific persona in a situated context, rather than as a generalized metric.

\subsubsection{Evaluation Dimensions and Metric System}

\textbf{(1) Quantitative Evaluation (Based on EPM)}

This section details the quantitative assessment system derived from EPM, designed to objectively measure the multi-dimensional performance of models in empathetic interactions.

\textbf{\textit{Core Philosophy: Open Comprehensive Evaluation Paradigm}}

Traditional evaluation benchmarks often attempt to calculate a single comprehensive score for leaderboard ranking through predetermined fixed weight combinations. However, empathy is a highly complex social signal interaction, and different application scenarios have vastly different demands on model capabilities. An efficient model that excels in scenarios requiring quick problem-solving might perform poorly in scenarios requiring long-term, patient companionship due to a lack of stability\cite{zhou2024sotopiainteractiveevaluationsocial,2023Holistic}.
Therefore, the EPM framework theoretically abandons the single fixed weight approach and proposes an Open Comprehensive Evaluation Paradigm. Its core objective is not to provide an absolute ranking but to utilize the rich quantitative metrics provided by EPM to depict a unique capability profile for each model, clarifying its strengths, weaknesses, and trade-offs. Like a strategic horse race, it helps users select the most suitable model for specific task scenarios.
We propose a methodology for dynamically adjusting evaluation emphasis based on application scenarios, for example:

\begin{itemize}
\item \textbf{Crisis Intervention Scenario:} Extremely high weight is given to process stability metrics (such as high positive energy ratio, extremely low penalty rate), with the primary goal being safety and robustness.
\item \textbf{Long-Term Companionship Scenario:} High weight is given to outcome metrics and positive energy, with higher tolerance for path tortuosity, focusing on deep connection.
\item \textbf{Task-Oriented Counseling:} High weight is given to process efficiency metrics (such as low turns, high density), focusing on efficient problem-solving.
\end{itemize}
Through this paradigm, we can identify typical characteristics of different models, such as robust players that perform balanced across all scenarios, or surprise players that perform stunningly at times but with high volatility. This perspective prompts us to recognize the strengths and weaknesses of each model, rather than simple high or low scores.

Utilizing the refined trajectory data generated by the simulation sandbox, a quantitative metric system containing both outcome and process dimensions is constructed.

\textbf{\textit{Metric Conversion and Calculation Logic Explanation}}

To ensure scientific rigor in scoring, the aforementioned raw physical metrics are not directly summed up but converted following a set of Scientifically-Defined Open Benchmark Index logic.

\begin{enumerate}[label=(\arabic*)]
\item \textbf{Scientific Anchoring:} All calculation benchmarks are strictly anchored to the physical definition of the task (such as initial deficit $r_0$) or the mathematical theoretical limit of the scale (such as maximum intensity $\rho_{max}$), rather than arbitrary empirical values.
\end{enumerate}
\begin{table}[H]
\centering
\caption{EPM Process Metrics — Measuring Path Strategy, Efficiency, and Stability} 
\label{tab:epm_process_metrics}
\begin{subtable}{\textwidth}
\centering
\caption{EPM Process Efficiency Metrics — Measuring Time Cost and Strategic Directness}
\small 
\begin{tabularx}{\textwidth}{l c >{\small\raggedright\arraybackslash}X}
\toprule
\textbf{Metric Name} & \textbf{Symbol} & \textbf{Core Meaning and Evaluation Value} \\
\midrule
\textbf{Empathy Density} & $\rho$ &
Measures average intervention intensity. The ``gold content'' of effective empathy energy delivered on average per dialogue turn. \\
\addlinespace
\textbf{Average Effective Projection} & $s_{\text{proj}}$ &
Measures single-turn effectiveness. The average effective projection component of the action vector along the ideal direction per turn. \\
\addlinespace
\textbf{Path Tortuosity} & $\tau$ &
Measures strategic directness. The ratio of the actual trajectory length to the straight-line displacement between start and end points. \\
\bottomrule
\end{tabularx}
\end{subtable}
\end{table}

\begin{table}[H]
\ContinuedFloat  
\centering
\begin{subtable}{\textwidth}
\centering
\caption{EPM Process Stability Metrics — Measuring Interaction Smoothness, Directional Correctness, and Safety}
\small 
\begin{tabularx}{\textwidth}{l c >{\small\raggedright\arraybackslash}X}
\toprule
\textbf{Metric Name} & \textbf{Symbol} & \textbf{Core Meaning and Evaluation Value} \\
\midrule
\textbf{Average Alignment} & $\overline{\cos\theta}$ &
Measures directional consistency. The average cosine value of the angle between the model's intervention direction and the ideal healing direction. \\
\addlinespace
\textbf{Positive Energy Ratio} & $R_{pos}$ &
Measures process smoothness. The proportion of turns generating positive propulsion out of total turns. \\
\addlinespace
\textbf{Performative Penalty Rate} & $R_{pen}$ &
Measures the intensity of negative behavior. Quantifies the average punishment received by the model due to inappropriate remarks (e.g., lecturing, indifference). \\
\bottomrule
\end{tabularx}
\end{subtable}
\raggedright
\vspace{4pt}
\footnotesize
\textit{Note: Quantitative evaluation supports the ``Open Comprehensive Evaluation Paradigm,'' allowing dynamic adjustment of metric weights based on application scenarios for capability profiling.}
\end{table}

\begin{table}[H]
  \caption{EPM Outcome Metrics - Measuring Final Efficacy and Total Workload}
  \small 
  \begin{tabularx}{\textwidth}{l c >{\small\raggedright\arraybackslash}X}
    \toprule
    \textbf{Metric Name} & \textbf{Symbol} & \textbf{Core Meaning and Evaluation Value} \\
    \midrule
    \textbf{Task Completion Status} & $Status$ & Final success/failure determination based on the ``Trinity Victory Condition'' (geometric/positional achievement and sufficient energy). \\
    \addlinespace
    \textbf{Relative Distance Improvement} & $RDI$ & Measures the thoroughness of healing. Calculates the percentage improvement of the user's final psychological state relative to the initial deficit. \\
    \addlinespace
    \textbf{Cumulative Effective Energy} & $E_{total}$ & Represents total empathetic work done. Measures the total effective intervention exerted by the model along the ideal healing direction throughout the dialogue, reflecting the magnitude of ``substantive effort.'' \\
    \addlinespace
    \textbf{Energy Surplus} & $E_{surplus}$ & Measures empathy abundance. Calculates the additional energy support provided beyond the basic requirement. \\
    \addlinespace
    \textbf{Total MDEP Net Score} & $S_{net}$ & Measures total empathy quality. The sum of cumulative net scores obtained in the three dimensions of C/A/P. \\
    \bottomrule
  \end{tabularx}
  \label{tab:metric_definitions}
\end{table}
\clearpage
\begin{enumerate}[label=(\arabic*), start=2]
\item \textbf{Classification Conversion:}
\begin{itemize}
\item For unbounded cumulative metrics (such as energy, net score), their multiplier relative to the scientific benchmark is calculated, forming an uncapped open index to reflect the excess performance of exceptional models.
\item For bounded ratio metrics (such as RDI, alignment), standard linear mapping is adopted to convert their physical boundaries into [0, 100] interval scores.
\end{itemize}
\item \textbf{Synthetic EPM-Index:} Finally, through weighted synthesis, an open EPM benchmark index is output. Index=100 represents precisely achieving the scientific benchmark, while Index>100 intuitively reflects the excellence multiplier beyond the benchmark.

\begin{equation}
  \textbf{EPM-Index} = 0.4 \cdot \boldsymbol{\tilde{S}}_{\text{Outcome}} + 0.2 \cdot \boldsymbol{\tilde{S}}_{\text{Efficiency}} + 0.4 \cdot \boldsymbol{\tilde{S}}_{\text{Stability}}
\end{equation}
\end{enumerate}

\textbf{(2) Qualitative Evaluation (Based on NEE)}

Although EPM provides rigorous physical quantitative benchmarks, human complex empathy experience still possesses many dimensions difficult to reduce purely to numerical values (such as naturalness of language, rhythm of emotional interaction, etc.)\cite{Andreetta_2023}. To capture these phenomenological characteristics crucial for user experience, we constructed the Narrative \& Experience Evaluator (NEE) as an independent high-level appreciation layer. It follows the principle of diagnosis first (context), evaluation later (experience), adopting a simulated judicial review process to conduct a holistic critical assessment of the dialogue.

Core Qualitative Dimensions:

\begin{itemize}
\item \textbf{Linguistic Naturalness}: Evaluates the model's ability to eliminate machine-like and artificial feelings in language expression, aiming to distinguish between natural anthropomorphic communication and stiff text generation. Rewards low-resistance expression and penalizes performative over-embellishment.

\item \textbf{Contextual Rhythmic Adaptation}: Evaluates the model's ability to regulate emotional energy, examining whether it can provide energy resonance attuned to the user's state at the correct timing (e.g., receiving during high-energy catharsis).

\item \textbf{Narrative Arc and Depth}: Examines the aesthetic quality of the entire dialogue as a meaning-making process, assessing narrative coherence, depth of cognitive restructuring ability, and the existence of penetrating highlight moments.
\end{itemize}
To avoid potential aesthetic preferences and systemic biases brought by a single reviewer model, NEE adopts a Joint Evaluation approach similar to an expert committee. We integrated four top large models with distinct expertises in different dimensions to form a complementary review panel (GPT-4o, Claude 3.5 Sonnet, Gemini 2.5 Pro, DeepSeek R1). Through this joint review mechanism, we are able to obtain more diverse, objective, and high-consensus qualitative evaluation results. 

\subsubsubsection{\textbf{(3) Comprehensive Score Calculation}}

The final comprehensive score of the model aims to balance objective physical benchmarks with subjective experience. We adopt a weighted fusion strategy, combining the EPM quantitative index with the NEE qualitative score. The standard calculation uses balanced weights (50/50), defining the Final Comprehensive Score (FCS) as follows:

\begin{equation}
    \text{FCS} = 0.6 \cdot \text{EPM-Index}^* + 0.4 \cdot \text{NEE-Score}^*
\end{equation}

\noindent where the superscript ($*$) indicates the specific domain of each metric, representing the quantitative physical benchmark and the qualitative subjective experience, respectively.

This mechanism ensures that the final result reflects not only the work done and efficiency of the model at the physical level but also fully embodies its naturalness and depth at the human interaction experience level.

\section{Results and Analysis} \label{sec: experiment}
\subsection{SFT}
The performance of our finetuned model on the evaluation benchmarks is in Table \ref{tab:table_sft_eval} and Table \ref{tab:table_sft_eq_eval}. 

Our SFT procedure enhances the model’s humanlike expressiveness without compromising its general capabilities. Notably, wthe model shows a slight improvement on the IFEval\cite{zhou2023instructionfollowingevaluationlargelanguage} benchmark, indicating that instruction-following ability is further strengthened after fine-tuning. Unfortunately, we observed a noticeable degradation in both our private IQ and EQ benchmarks after SFT. This decline is largely attributable to the imbalance in the training mixture: high-quality AI companionship data accounts for only 4.45\% of the SFT corpus, whereas role-play data dominates. Since role-play interactions differ substantially from our target AI-companionship domain, the supervised finetuning stage naturally steers the model away from the behaviors emphasized in our benchmarks. That said, this degradation is not a major concern. In our pipeline, SFT primarily serves as a cold start to initialize the policy, while the RL stage is responsible for the substantive alignment and restoration of domain-specific capabilities.

\begin{table}[H]
  \centering 
  \caption{Evaluation results on public benchmarks and our private static IQ benchmark. Qwen3-32B serves as the base model, while Qwen3-32B-SFT and Qwen3-32B-RL denote our supervised fine-tuned and reinforcement learning fine-tuned variants, respectively. We also included the evaluation results from a range of leading open-source and closed-source models to provide a comprehensive comparative context.}
  \vspace{5pt}
  \begin{tabular}{l c c c c}
    \toprule
    Models & IFEval & ChineseSimpleQA & CharacterEval & IQ Test \\
    \midrule
    Qwen3-32B & 79.02 & 44.03 & 2.85 & 45.45 \\
    Qwen3-32B-SFT & 81.29 & 42.7 & 3.08 & 29.09 \\
    Echo-N1(ours) & 82.61 & 42.63 & 3.12 & 34.55 \\

    \midrule
    Gemini2.5-pro & 85.37 & 75.17 & 3.23 & 72.7 \\
    Kimi-K2 & 88.85 & 74.23 & 2.99 & 52.7 \\
    Doubao1.5-Chracter & 80.94 & 56.93 & 2.92 & 60 \\
    Qwen3-235B & 88.97 & 82.3 & 2.92 & 58.2 \\
    \bottomrule
  \end{tabular}

  \label{tab:table_sft_eval}
\end{table}

\begin{table}[H]
  \centering 
  \caption{Win rates in pairwise preference evaluations. Each entry reports the win rate of the row model against the column model. For each benchmark query, both models generate responses conditioned on the same context, and a preference-aligned evaluator prompt is used to determine the winner.}
  \vspace{5pt}

  \begin{tabular}{c c c c c}
    \toprule
    Vs. & Kimi-K2 & Qwen3-32B-SFT & Qwen3-32B  & Doubao1.5-Character \\
    \midrule

    Qwen3-32B-SFT & 21.2\% & - & 35.2\%  & 40.7\% \\
    Qwen3-32B & 27.4\%  & 64.8 \% & - & 58.7\% \\
    Doubao1.5-Character & 18.4\% & 59.3\% & 41.3\% & - \\
    Echo-N1(ours) & \textbf{38.0\%} & \textbf{65.9\%} & \textbf{79.9\%} & \textbf{95.5\%} \\
    \bottomrule
  \end{tabular}

  \label{tab:table_sft_eq_eval}
\end{table}

\subsection{Reward Models} \label{sec: exp_reward_model}
\subsubsection{Humanlike Reward} \label{sec: exp_humanlike_reward}
We evaluate both humanlike and empathetic reward models under the “LLM-as-a-judge’’ paradigm and compare them against strong proprietary baselines. As a starting point, we consider SOTA LLMs as direct judges. Whi le this provides a natural baseline, our results reveal a fundamental limitation: humanlike and empathetic evaluation requires modeling subjective, socially grounded aspects of human behavior—capabilities for which current frontier LLMs are not explicitly optimized. Because these models are primarily trained for reasoning-centric tasks and assistant-style interactions, they tend to produce generic, polite outputs rather than genuinely humanlike conversational behavior. As a result, they exhibit poor reliability when tasked with discriminating subtle preference differences.

To empirically validate this weakness, we apply strong prompting to Gemini-2.5-Pro, Claude-4-Sonnet, and GPT-5 and evaluate them on humanlikeness judgment. As shown in Table~\ref{tab:table_humanlike_test}, all models perform poorly on both the standard test set and our curated hard negatives—instances that are trivial for humans but systematically misclassified by LLMs. These observations highlight an inherent flaw in the “LLM-as-a-judge’’ paradigm for evaluating humanlikeness: the models lack the representational grounding required for stable preference discrimination.

Motivated by these findings, we train dedicated reward models. For the humanlike task, we report results for two variants—a context-free model and a context-based model—whose training setup is detailed in Section~\ref{sec:humanlike_data}. Although the context-free model performs reasonably when evaluated in isolation, it induces severe reward hacking during RL: the policy over-optimizes for surface-level linguistic naturalness while ignoring logical consistency. In contrast, the context-based model provides a more robust balance between naturalness and contextual coherence, making it the only viable option for downstream reinforcement learning.

\begin{table}[h]
  \centering 
  \caption{Judger accuracy on the humanlike-expression test set. Hard Negatives denotes our manually curated adversarial examples designed to challenge models prone to superficial pattern matching.}
  \vspace{5pt}

  \begin{tabular}{lcc}
    \toprule
    Models & Test Set & Hard Negatives \\
    \midrule
    Gemini-pro-2.5 & 42.13\% & 12.5\% \\
    GPT-5       & 43.66\% & 28.1\% \\
    Claude-Sonnet-4.5 & 50.72\% & 32.3\% \\
    Humanlike Judger w/o context (ours) & \textbf{90.83\%} & \textbf{90.6\%} \\
    Humanlike Judger w/ context (ours) & 89.45\% & 31.3\% \\
    Qwen3-8B  & 43.28\% & 50\%\\
    \bottomrule
  \end{tabular}

  \label{tab:table_humanlike_test}
\end{table}

\subsubsection{Empathy Reward} \label{sec: exp_empathy_reward}
The setting for empathetic evaluation differs markedly from the humanlike case. With a carefully engineered prompt—co-designed with expert psychologists—Gemini-2.5-Pro can demonstrate nontrivial competence in dynamic user-preference inference and empathetic assessment. However, its performance relies critically on extremely long prompts, often requiring thousands of tokens per inference. Truncating the prompt leads to abrupt performance collapse, indicating that the model overrelies on full contextual scaffolding and cannot be made cost-efficient through simple prompt compression. This makes such prompting strategies impractical for downstream RL, where thousands of online evaluations would result in prohibitive latency and API costs.

We therefore train our own empathetic reward models, and report results in Table~\ref{tab:table_empathy_test}. We evaluate both 8B and 32B variants~\cite{qwen3technicalreport} and observe a clear scaling trend: larger models consistently achieve stronger empathetic discrimination performance. To systematically evaluate these models, we construct two test sets: (1) an in-distribution set in which dialogue histories and preference labels come from the same annotators used during reward model training, and (2) an out-of-distribution set in which preference candidates are generated by Kimi-K2~\cite{kimiteam2025kimik2openagentic}. This dual-setting design enables us to isolate both absolute judging capability and robustness to distribution shift.

As shown in Table~\ref{tab:table_empathy_test}, our 32B empathetic reward model achieves \textbf{93.30\%} accuracy in-distribution and \textbf{69.00\%} OOD, outperforming all open-source baselines by a substantial margin. The 8B variant achieves \textbf{83.15\%} and \textbf{53.50\%}, confirming the expected scaling behavior. Interestingly, while Gemini obtains the strongest OOD score (\textbf{91.97\%}), it performs worse in-distribution, suggesting that empathetic alignment requires reasoning patterns not fully captured by general-purpose conversational models. The gap between our two variants also reflects the benefit of the data-evolution pipeline described in Section~\ref{sec: method_empathy_reward}, where “exploration via sampling’’ combined with “filtering via hard constraints’’ successfully extracts reasoning paths that surpass the intrinsic capability ceiling of the small model.

\paragraph{Ablation on Process-Aware Reward.}
We isolate the contribution of the process-aware term $R_{\text{process}}(r)$ by comparing the full multiplicative reward with an outcome-only variant that sets $R_{\text{process}}(r)=1$ for all trajectories.  
The outcome-only baseline quickly saturates the scalar empathy signal and subsequently exhibits clear degradation on the held-out validation set—an indication of reward overfitting.  
By contrast, the full reward maintains stable and consistent improvements throughout training.  
Manual inspection further reveals that removing $R_{\text{process}}(r)$ leads the policy to exploit stylistic artifacts and produce logically fragile responses, whereas the full reward suppresses such modes.  
These results demonstrate that $R_{\text{process}}(r)$ is essential not only for stabilizing optimization but also for preventing reward hacking and preserving robust empathetic reasoning.

Overall, our experiments reveal \textbf{three key findings}:  
(1) general-purpose LLMs, even with strong prompting, are not reliable judges for humanlike or empathetic evaluation;  
(2) for reward models, both context-based modeling and the process-aware multiplicative term $R_{\text{process}}(r)$ are essential to prevent reward hacking and to sustain genuine improvements on held-out validation tasks;  
(3) empathetic reward models benefit substantially from scaling and from the curated data-evolution process, enabling the 32B model to surpass proprietary baselines on in-domain human preference tasks while maintaining competitive robustness under distributional shift. Detailed prompts used in Stage II are provided in the \textbf{Appendix}~\ref{appendix:empathetic_judger_prompt}.

\begin{figure}[h]
    \centering
    \includegraphics[width=\linewidth]{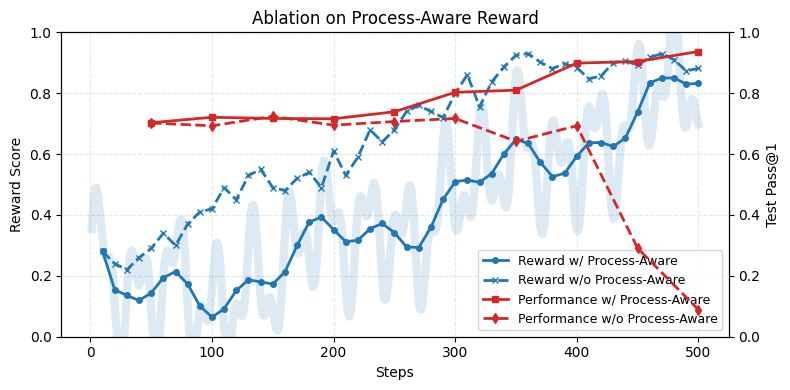}
    \caption{
    Ablation of the process-aware term $R_{\text{process}}(r)$ in Eq.~\ref{eq:final_empathy_reward_function}.
    The figure reports reward scores (blue, left axis) and validation Pass@1 (red, right axis) for the full reward (solid) and the outcome-only variant (dashed).
    }
    \label{fig:process_ablation}
\end{figure}

\begin{table}[h]
  \centering
  \caption{Performance comparison across in-domain (Test Set) and out-of-domain (Test Set OOD) splits. For Pass@1 (A $\cap$ B),~a trial is considered successful only if the model generatively passes (Pass@1) both versions of a question, where the correct option's position is swapped between A and B.
  }
  \vspace{5pt}
  \begin{tabular}{l l c c}
    \toprule
    Type & Model & Test Set & Test Set (OOD) \\
    \midrule
    
    \multirow{1}{*}{scalar} 
      & WorldPM & 98.00\% & 78.29\% \\
    
    \midrule
    
    \multirow{5}{*}{\shortstack{generative \\ \scriptsize pass@1 (A $\cap$ B)}}

    & Gemini 2.5 Pro & \underline{88.83\%} & \textbf{91.97\%} \\
    & Qwen3-8B & 76.54\% & 50.56\% \\
    & Qwen3-32B & 77.65\% & 56.57\% \\
    & Empathy Judger-8B (ours) & 83.15\% & 53.50\% \\
    & Empathy Judger-32B (ours) & \textbf{93.30\%} & \underline{69.00\%} \\

    \bottomrule
  \end{tabular}

  \label{tab:table_empathy_test}
\end{table}

\subsection{Echo-N1}
In this section, we study how different reward-modeling choices affect downstream policy optimization in empathetic dialogue tasks. While validation accuracy provides a useful signal, it is often insufficient for understanding a reward model’s real behavior during RL—where stability, resistance to reward hacking, and the structure of the reward signal matter just as much as raw correctness. We therefore evaluate several aspects of reward model design, including comparing GenRM with a scalar baseline, examining robustness during RL, and analyzing how reference-answer-based and discrete rewards influence training dynamics. Together, these experiments provide a practical view of what makes a reward model usable and reliable for optimizing empathetic conversational agents.

\subsubsection{GenRM vs Scalar}
\begin{figure}[H]
    \centering
    \includegraphics[width=0.8\linewidth]{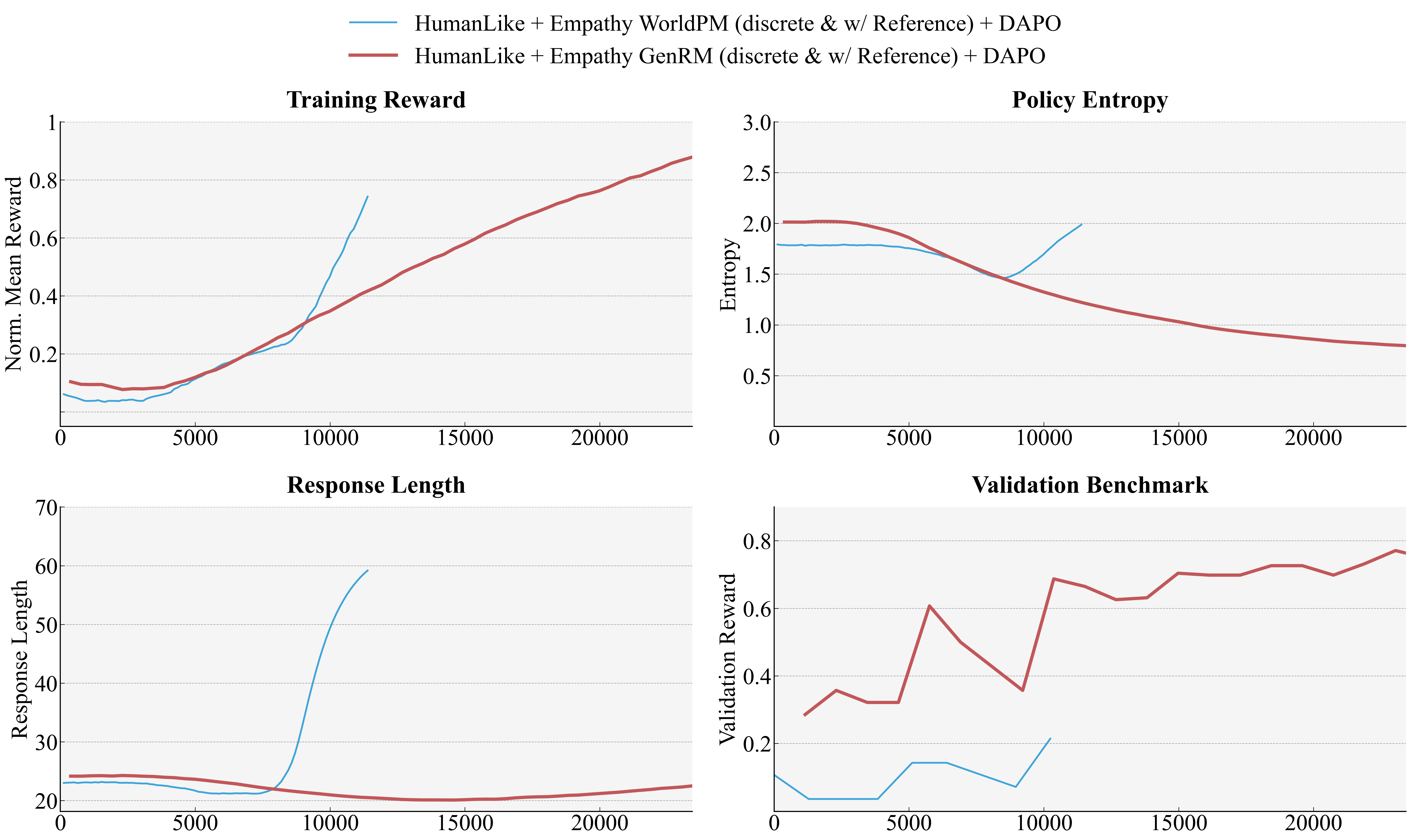}
    \caption{
    Training stability comparison across reward modeling strategies. Empathy GenRM successfully realizes stable training, maintaining controlled entropy and steady reward growth unlike the scalar baseline (WorldPM) which suffers from severe reward hacking.}
    
    \label{fig:worldPM_vs_genRM}
\end{figure}

\begin{figure}[H]
    \centering
    \includegraphics[width=0.8\linewidth]{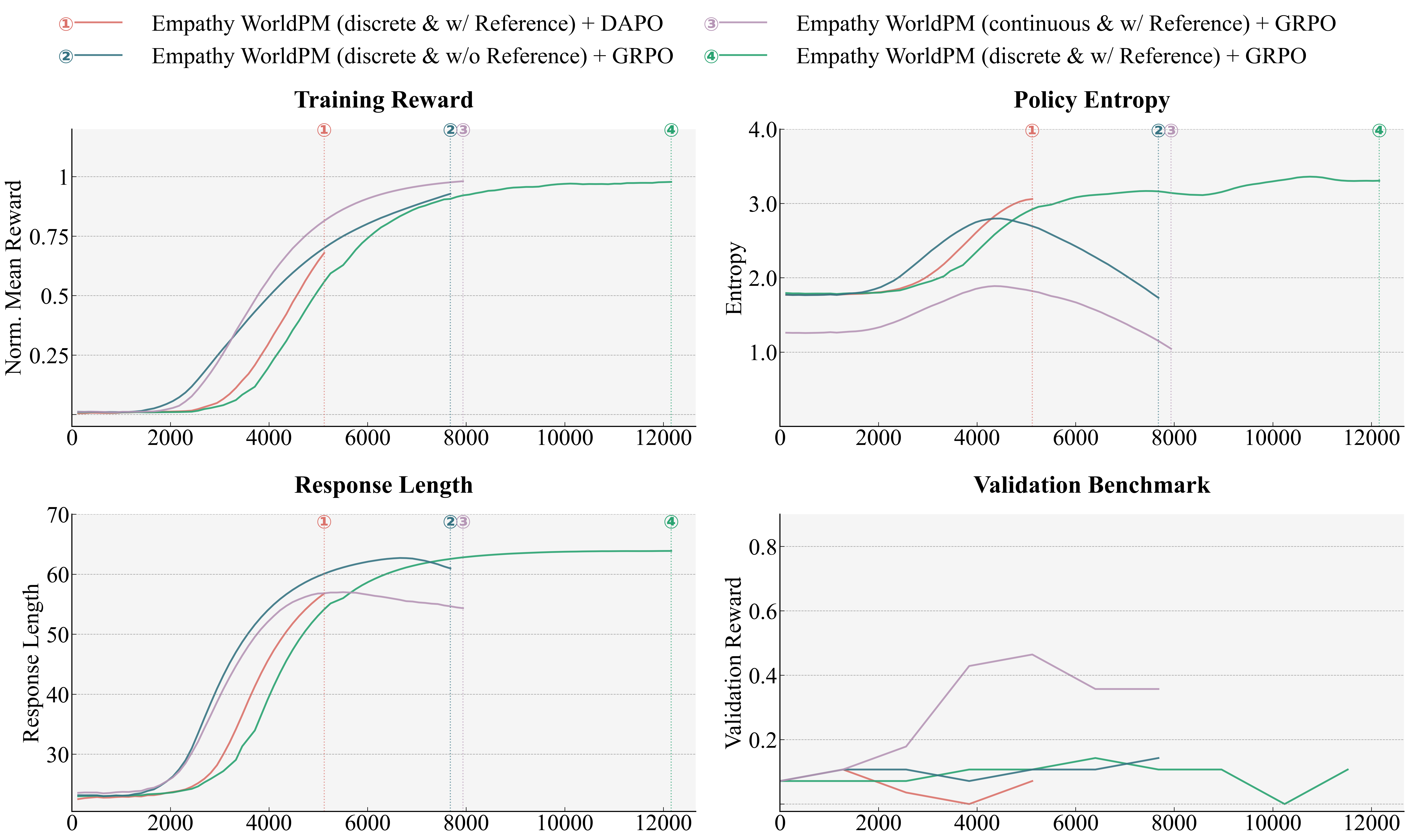}
    \caption{
        Robustness comparison across reward modeling strategies.
        Ablation results show that reference-answer-based reward computation and discrete (binary) rewards significantly improve robustness and mitigate reward hacking.
    }
    \label{fig:empathy_worldPM}
\end{figure}

To verify GenRM’s advantages in empathetic scenarios, we designed a comprehensive suite of experiments that rigorously compares GenRM with a scalar reward model. Specifically, we fine-tuned WorldPM-72B \cite{worldpm} on the same preference dataset used for GenRM. We evaluated both models on a hand-curated preference dataset; as shown in Table \ref{tab:table_empathy_test}, WorldPM-72B achieved a slight win. However, validation performance alone does not fully capture reward model quality. A strong reward model must not only perform well on a valid dataset, but also provide reliable reward signals during RL and resist reward hacking. We therefore ran RL experiments using either WorldPM or Empathy GenRM as the reward model to answer three questions in practice: which model is better, whether reference-answer-based reward computation helps, and whether discrete rewards outperform continuous rewards.

\textbf{Who is more robust, WorldPM or GenRM? }

WorldPM exhibited pronounced reward hacking in early training under both DAPO and GRPO. The model strongly preferred longer responses regardless of content quality, causing response length to spike within a few steps until hitting the maximum allowed length. This indicates that the policy learned to inflate verbosity to collect higher rewards, revealing severe verbosity bias in WorldPM. We attribute this to the reward-model training data. To ensure quality, we kept only high-scoring answers as the chosen set; these answers were typically longer than the rejected ones. A subsequent manual audit confirmed that longer responses are often better, but length should be an informative feature rather than the dominant criterion. WorldPM appears to have overfit to length, producing a marked preference for verbosity. This, in turn, suggests that scalar reward models like WorldPM are unstable to train, generalize poorly, and are easy to exploit.

By inspecting the training curves of entropy, response length, and reward under WorldPM, we observe a clear and consistent pattern. As policy entropy begins to increase, both the reward and the average response length rise in tandem. Once entropy reaches its peak and starts to decrease, the reward simultaneously approaches its maximum. This trajectory indicates that from the very beginning of training, the policy is consistently optimized toward a reward-hacking solution. While the increase in entropy reflects active exploration,  the lack of robustness in WorldPM allows the policy to rapidly identify a trivial, length-based heuristic for maximizing reward. Once this vulnerability is exploited, the model collapses into uncontrolled generation of verbose responses, leading to runaway reward hacking rather than meaningful improvement in response quality.

In contrast, Empathy GenRM significantly mitigates this reward hacking phenomenon.  Under identical training configurations, pairing Empathy GenRM with the Human-Likeness RM yields an entropy curve that, the entropy curve exhibits a similar arch of ascending then descending, indicating standard exploration and exploitation phases. However, unlike the scalar baseline, we observe no sudden spikes in the reward trajectory; instead, the reward demonstrates steady, linear growth. This suggests that during exploration, the policy avoids engaging in reward hacking and identifies a stable optimization direction for meaningful improvement. These results highlight the superior robustness of Empathy GenRM compared to scalar reward models, demonstrating its ability to provide a consistent reward signal that is resilient to reward hacking.

Taken together with Figure~\ref{fig:worldPM_vs_genRM}, these dynamics provide an operational notion of robustness in our setting. A robust reward model should (i) support smooth, monotonic improvement in reward without abrupt regime changes, (ii) allow entropy to rise and fall in a controlled way without triggering degenerate behaviors such as maximum-length responses, and (iii) maintain a tight coupling between what the model is rewarded for and how humans actually judge empathetic quality. Under this lens, GenRM is more robust than WorldPM: even when the policy explores aggressively, its learned behavior does not drift toward the extreme verbosity solution that plagues the scalar baseline. Instead, the policy continuously discovers higher-quality responses along a stable trajectory.

\textbf{Does reference-answer-based reward help?}

We find that using a reference answer to compute rewards mitigates reward hacking. Empathy GenRM assigns reward by comparing the policy response to a reference answer, while scalar models such as WorldPM can output rewards directly. In other words, GenRM is trained and used in a pairwise fashion: given a dialogue context, a high-quality reference response, and a policy response, it estimates how likely the policy response would be preferred over the reference.. Scalar models instead attempt to map a single response directly to a real-valued score, which implicitly assumes that the model has learned a well-calibrated absolute quality scale for complex, subjective behaviors such as empathy.

We evaluated WorldPM both with and without reference-answer-based reward computation. In the reference-answer-based setting, we reuse the high-scoring chosen answers from the preference dataset (Section~\ref{sec: exp_reward_model}) as reference responses and let WorldPM predict the probability that the policy response is better than the fixed reference. This probability is the reward signal used by RL, matching the pairwise-comparison formulation of our data. In contrast, we use the unscaled score that WorldPM outputs for each policy response as the reward in the setting without a reference answer. As shown in Figure \ref{fig:empathy_worldPM}, WorldPM without reference-answer-based reward begins to exhibit reward hacking around 3000 samples. The reward curve rapidly increases while response length spikes, indicating that the policy has discovered a simple but degenerate strategy for inflating the scalar score. In contrast, the model utilizing reference-answer-based reward delays the emergence of this issue until approximately 4000 samples. Although reward hacking is not completely eliminated, the onset is noticeably postponed and the growth of reward is smoother, reflecting a more constrained and meaningful exploration process. 

This comparison demonstrates that reference-answer-based reward is more effective for non-verifiable tasks. For empathetic conversation, there is no single ground-truth label analogous to a right or wrong answer in math or coding; what we can reliably obtain instead are relatively strong reference responses from our curated chosen set. By anchoring training to such a high-quality reference, the method constrains exploration to more useful regions of the response space: to receive a higher reward, the policy must produce responses that are better than a solid baseline rather than simply exploiting superficial correlations (such as extreme length) that happen to drive up a scalar score. In practice, this reduces optimization instability and makes it harder for the policy to lock onto trivial heuristics, thereby improving the robustness of RL on empathetic, non-verifiable objectives.

\textbf{Are discrete rewards better than continuous rewards?}

Discretizing rewards alleviates reward hacking and yields a clearer training signal. Both Empathy GenRM and WorldPM effectively estimate the probability that a policy response is better than a reference answer. Prior work \cite{seed1.5} used this probability as a continuous reward during RL. However, users rarely assess calibrated probabilities; they more naturally judge whether one answer is better than another. We therefore compared continuous and discrete rewards. Figure \ref{fig:empathy_worldPM} shows that, for the scalar reward model, discrete rewards outperform continuous rewards and delay the onset of reward hacking. Our interpretation is that discrete rewards provide a crisper, margin-like supervision signal that better matches the underlying human preference task (pairwise comparison) and is intrinsically harder to exploit than a dense, real-valued score. This makes discretization a simple yet effective design choice for stabilizing RL on non-verifiable, preference-driven objectives such as empathetic dialogue.

\subsubsection{Performance on Private IQ and EQ Benchmarks}
Echo-N1's performance on our private IQ and EQ benchmarks is shown in Table \ref{tab:table_sft_eval} and Table \ref{tab:table_sft_eq_eval}. Although SFT introduces a decline on these benchmarks, the subsequent RL stage effectively recovers the lost capability. After RL, the policy model performs on par with the base model across most evaluations, with the exception of the private IQ benchmark, where the substantial degradation caused by SFT makes full recovery more challenging. Notably, on benchmarks such as IFEval and our private EQ benchmark, the RL-trained model even surpasses the Qwen3-32B base model. This highlights the strength of RL: despite being trained only on empathetic scenarios, it delivers targeted domain-specific improvements without compromising general capabilities.

\subsubsection{Dynamic EQ Evaluation}

\textbf{(1) Dataset Construction and Sampling}

\begin{figure}[H]
    \centering
    \includegraphics[width=0.95\linewidth]{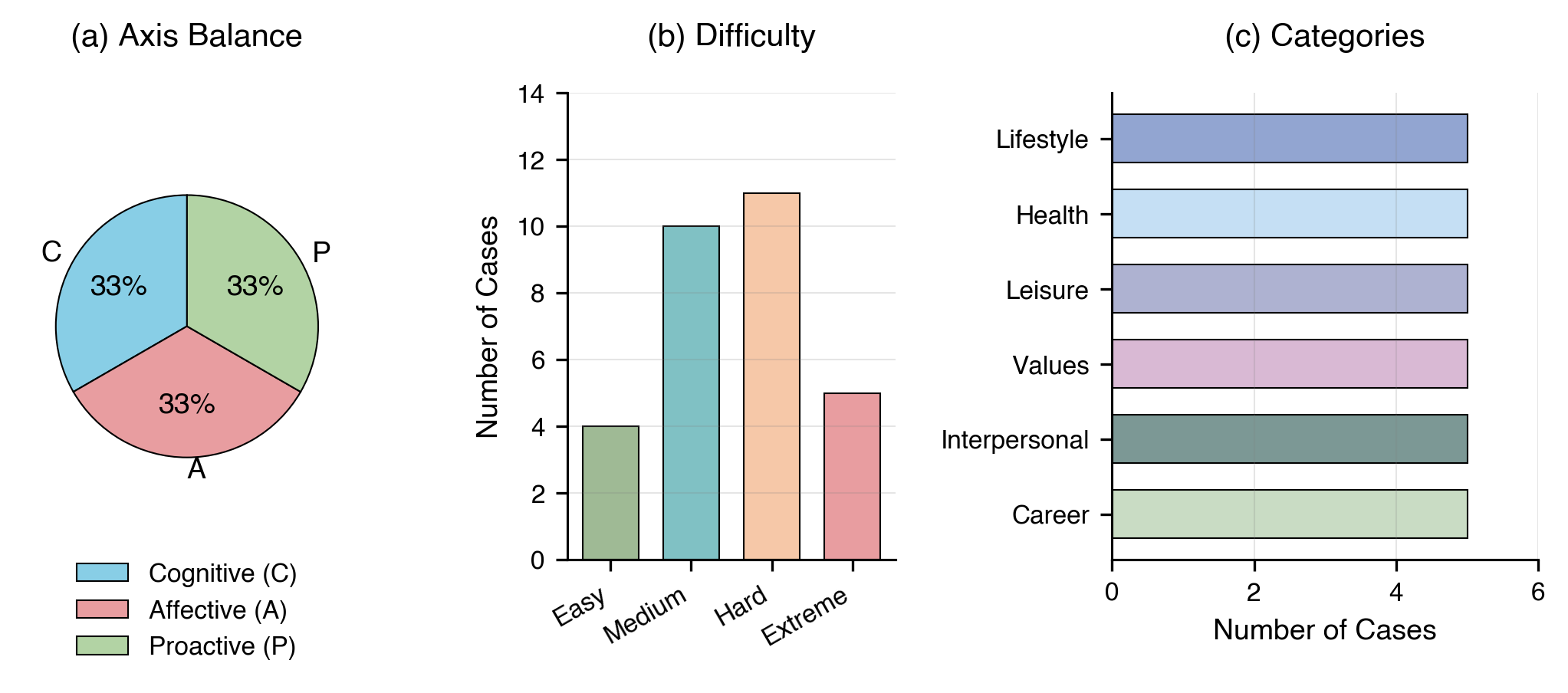}
    \caption{Overview of the dataset composition: (a) balanced distribution across the three core axes, (b) skewed difficulty distribution, and (c) even stratification across six life domains.}
    \label{fig:figADC}
\end{figure}

\begin{figure}[H]
    \centering
    \includegraphics[width=0.95\linewidth]{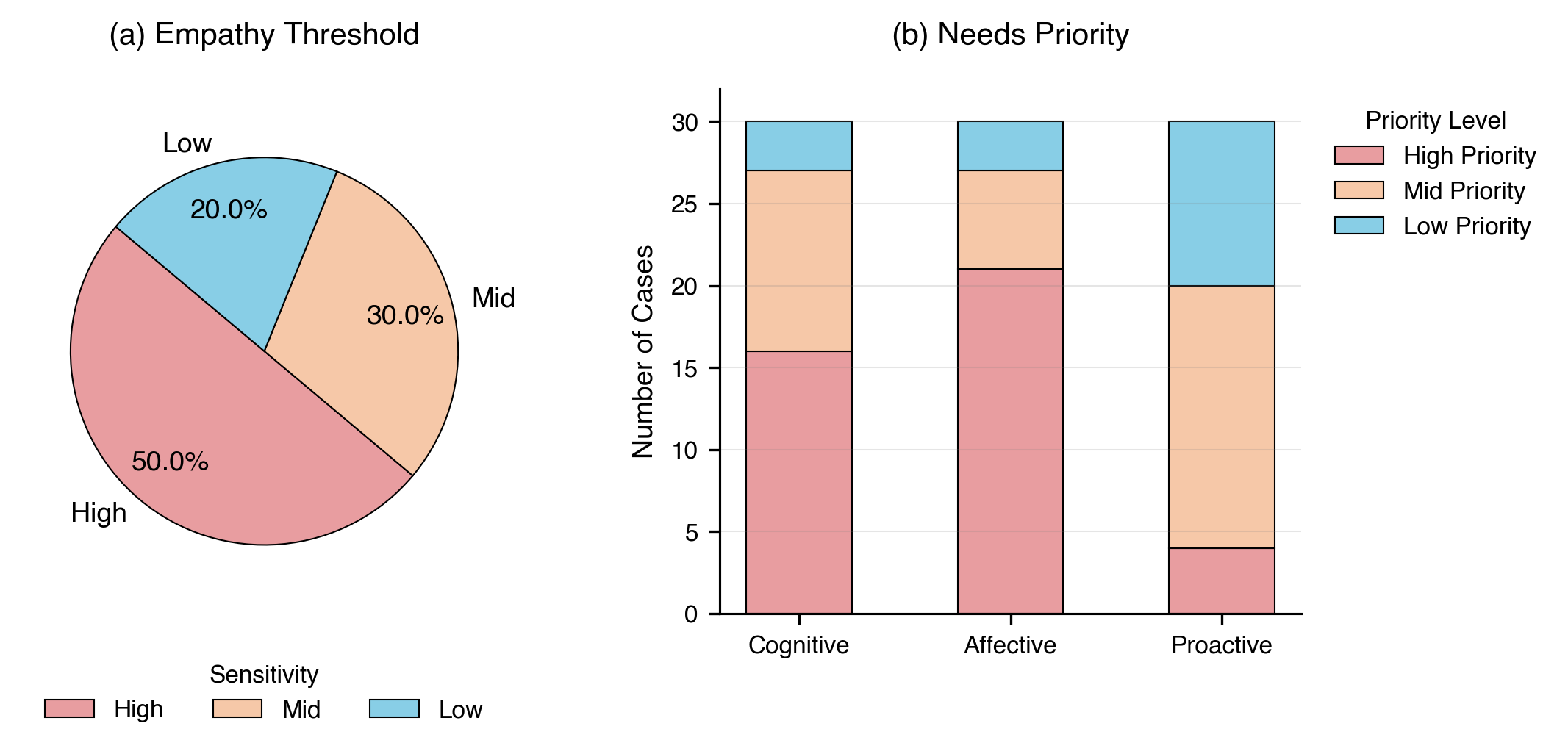}
    \caption{Analysis of user persona features. (a) The distribution of empathy thresholds, with 50\% of users having high sensitivity. (b) The priority of user needs across cognitive, emotional, and motivational dimensions.}
    \label{fig:figEN}
\end{figure}

To rigorously evaluate the model's empathetic capabilities in \textbf{Empathetic conversation and companionship scenarios}, we constructed a high-quality benchmark dataset comprising 30 test cases, carefully selected via a \textbf{Multi-Dimensional Stratified Sampling} strategy from a larger corpus of over 500 generalized user profiles. Despite the concise number of cases, the dataset achieves robust representativeness and high information density through this strategic selection.

First, each scenario is designed as a dynamic multi-turn dialogue (12-45 turns), generating hundreds of interactions that provide dense signals for evaluating long-context empathy maintenance. Second, we achieved perfect orthogonal coverage across the three core EPM dimensions, with Cognitive (C), Affective (A), and Proactive (P) dominant cases each constituting one-third of the dataset (Figure \ref{fig:figADC}a). These cases are evenly stratified across six distinct life domains (Figure \ref{fig:figADC}c) to maximize information gain and test cross-domain generalization. Third, the benchmark is engineered as a stress test. Based on the initial empathy deficit distribution of the full corpus ($\mu=32.32, \sigma=4.52$), the sampling is intentionally skewed as shown in \textbf{Figure \ref{fig:figADC}b: Extreme ($> \mu + \sigma$): 5 cases; Hard ($\mu \text{ to } \mu + \sigma$): 11 cases; Medium ($\mu - \sigma \text{ to } \mu$): 10 cases; and Easy ($< \mu - \sigma$): 4 cases}. Consequently, 86.7\% of the scenarios are classified as Medium difficulty or above, aiming to probe the model's capability in handling deep-seated emotional barriers. Finally, implicit feature analysis of user personas reveals a demanding test environment: half of the simulated users possess a high empathy threshold (Figure \ref{fig:figEN}a), with an overwhelming priority demand for Affective Resonance (A-axis) (Figure \ref{fig:figEN}b), closely mirroring the psychological reality in companionship scenarios where building connection precedes offering support. In summary, this dataset constitutes a highly challenging, dense, and structurally balanced evaluation benchmark.

\textbf{(2) Overall Success Rate Overview}

\begin{figure}[H]
    \centering
    \begin{subfigure}[b]{0.48\linewidth}
        \centering
        \includegraphics[width=\linewidth]{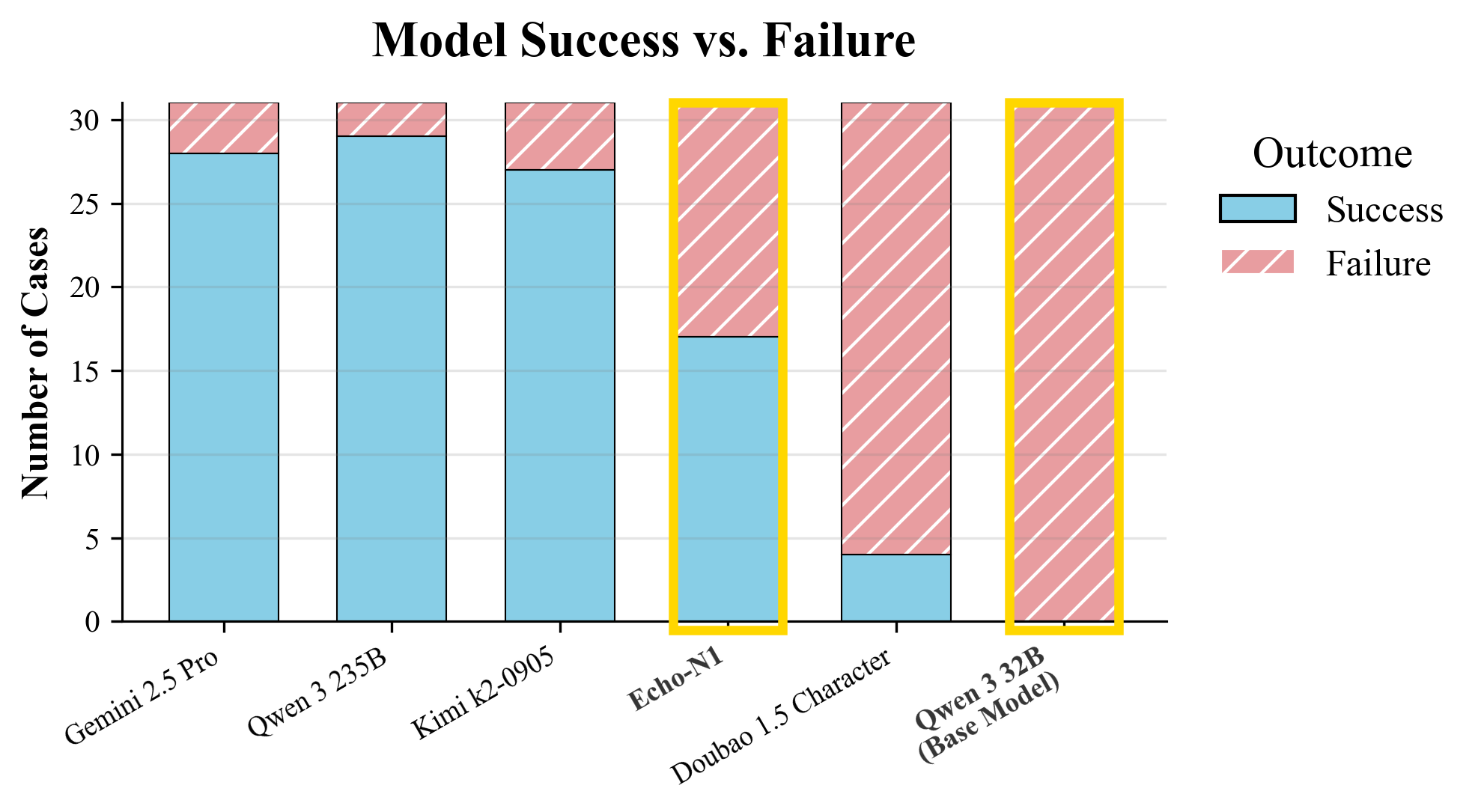}
        \caption{Success/failure distribution heatmap for each model. This visualization reveals nuanced differences in the capability boundaries of the evaluated models, grouping them into distinct performance tiers.}
        \label{fig:sub1}
    \end{subfigure}
    \hfill 
    \begin{subfigure}[b]{0.48\linewidth}
        \centering
        \includegraphics[width=\linewidth]{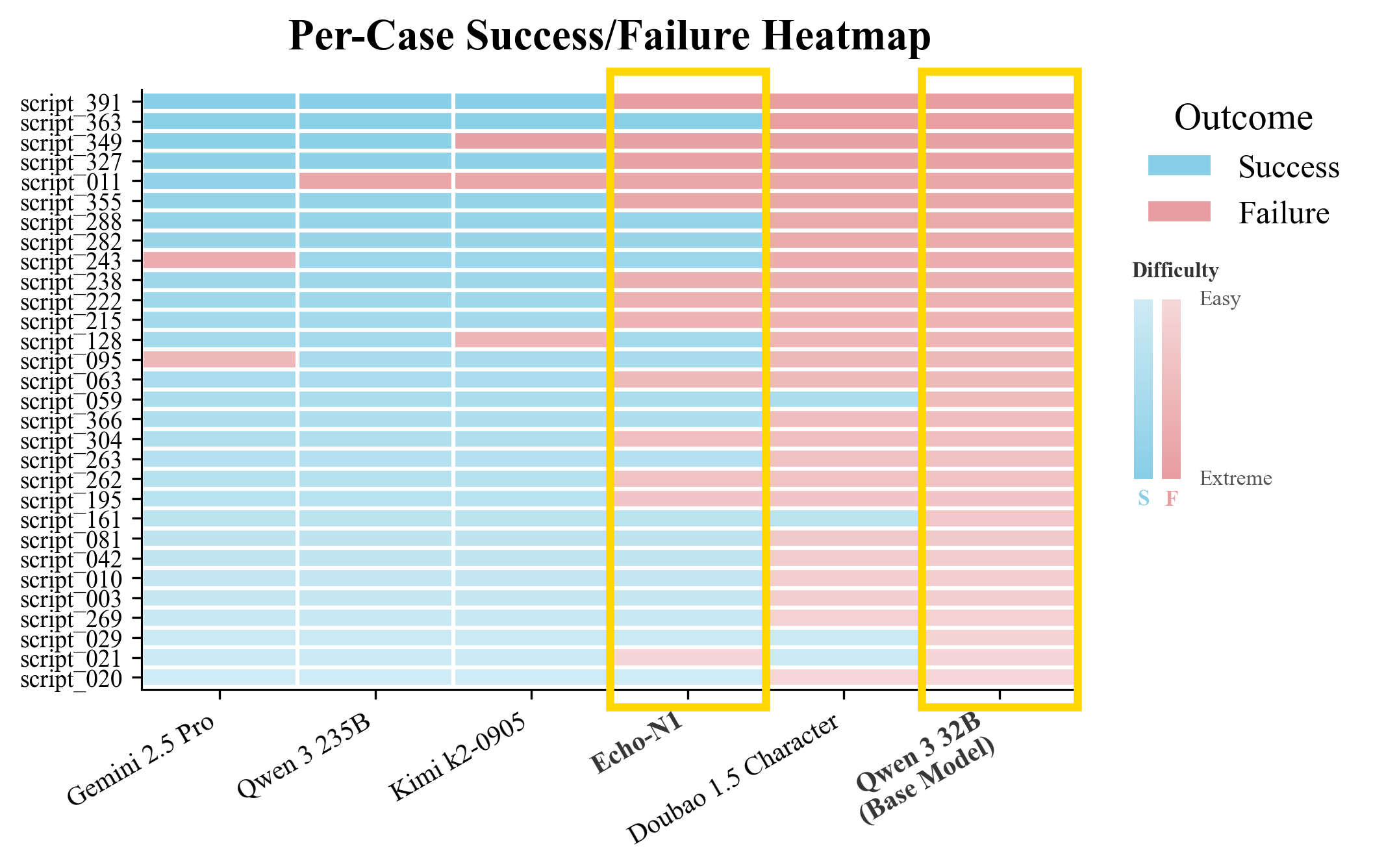}
        \caption{All-case success/failure status matrix. This matrix details the outcome for each model on every individual test scenario, revealing specific failure patterns that offer insights into the models' strategic preferences and limitations.}
        \label{fig:sub2}
    \end{subfigure}
    
    \caption{Comparative Performance Overview. The overall success rate of each model (a) and the detailed success/failure distribution (b) collectively reveal a clear stratification of capabilities among the models under the evaluation framework}
    \label{fig:main_label}
\end{figure}

The rigor of the Anthropomorphic Empathy Evaluation Framework is strikingly evident in the overall success rate distribution among the evaluated models. As illustrated in Figure \ref{fig:sub1} (stacked bar chart), the 30 test scenarios imposed a significant stress test, resulting in clear stratification of model performance. Notably, all scenarios categorized as successful strictly adhered to the Trinity Victory Condition defined by the EPM. This requires the simultaneous achievement of sufficient energy accumulation and effective therapeutic progress (via either geometric alignment or positional proximity):
\begin{equation}
    \text{Success} \iff \left( (\underbrace{\overline{\cos\theta} > \tau_{align}}_{\text{Geometric Victory}}) \lor (\underbrace{\|P_T\| < \epsilon_{dist}}_{\text{Positional Victory}}) \right) \land (\underbrace{E_{total} > \epsilon_{energy}}_{\text{Energy Victory}})
\end{equation}
This outcome robustly validates a core EPM hypothesis: effective empathetic intervention is not achieved through isolated tricks but is the composite result of directional alignment, state improvement, and sustained energy output. A deficiency in any single dimension inevitably leads to therapeutic failure.

Under this stringent standard, our model, \textbf{Echo-N1}, demonstrated decisive progress. While its base model, \textbf{Qwen 32B (Base)}, failed across all 30 scenarios (0\% success rate), Echo-N1 successfully completed 14 (46.7\% success rate). More remarkably, Echo-N1 substantially outperformed the commercial baseline \textbf{Doubao 1.5 Character} (approx. 200B+ parameters, 13.3\% success rate), despite the latter's vastly larger parameter scale. This provides compelling evidence for the superiority of RLAIF training—inspired by empathy principles and strictly aligned with user profiles and contexts—over mere parameter scaling in enhancing domain-specific capabilities.

Figure \ref{fig:sub2} (success/failure distribution heatmap) further reveals nuanced differences in model capability boundaries. \textbf{Gemini 2.5 Pro}, \textbf{Qwen 235B}, and \textbf{Kimi k2} form a leading tier with exceptionally high success rates. However, rather than focusing on minor differences in their total scores, the specific failure patterns revealed in Figure \ref{fig:HUGETABLE} (all-case success/failure status matrix) hold greater analytical value. We observe that even top-tier models encounter failures in specific high-difficulty or particular types of scenarios (primarily manifesting as timeouts or insufficient energy). These specific failure points are not random; they conceal critical clues regarding models' strategic preferences and limitations. This suggests the necessity of moving beyond binary success/failure judgments and delving into the granularity of EPM quantitative metrics. Such potential analysis is essential to parse their respective strategic strengths and weaknesses, and ultimately, to \textbf{highlight the significant gaps that remain between current model capabilities—even at the SOTA level—and the rigorous demands of the EPM-Q comprehensive scoring system.}

\textbf{(3) Key Quantitative Metrics Details}

\begin{figure}[H]
    \centering
    \includegraphics[width=0.8\linewidth]{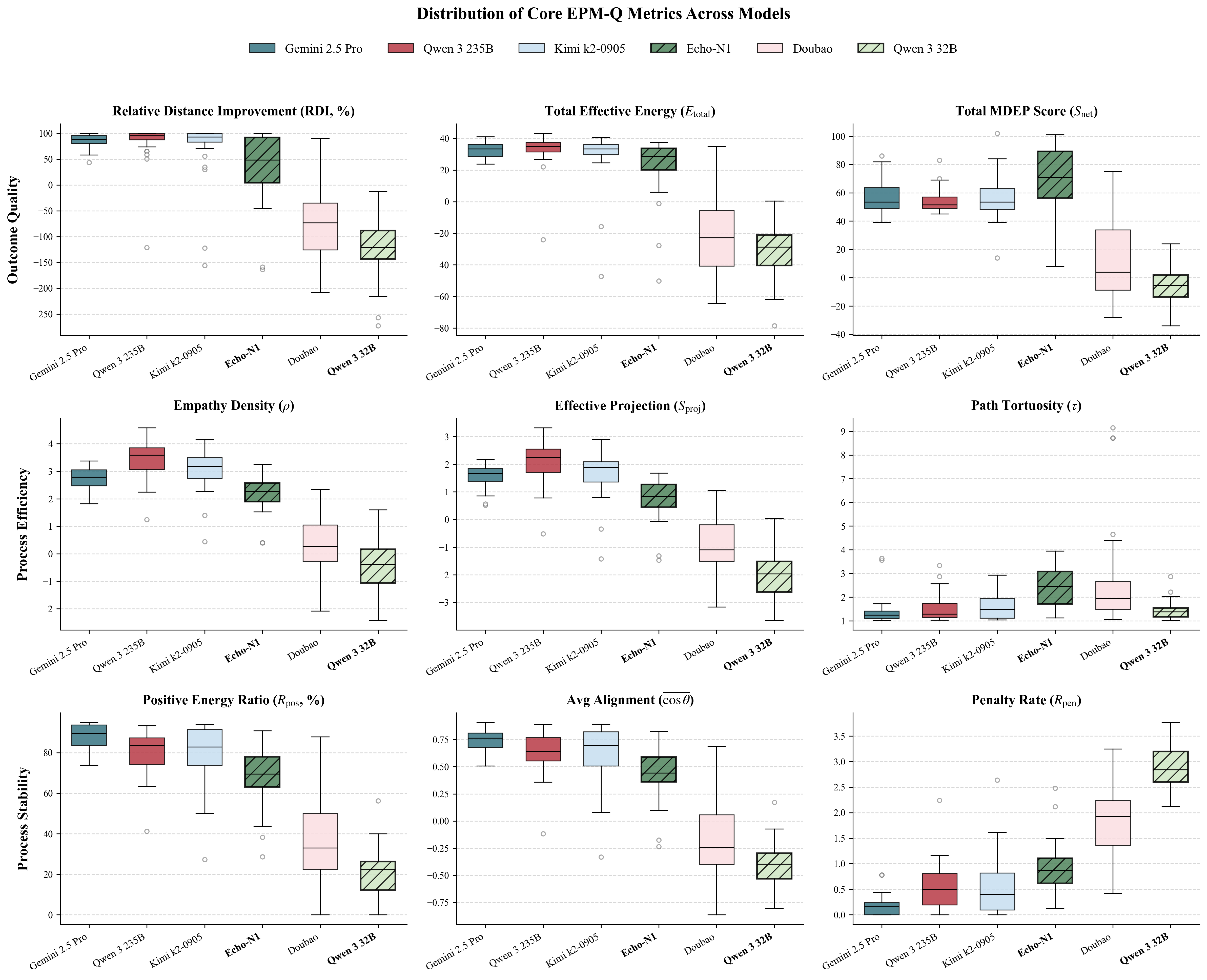}
    \caption{Distribution of Core EPM-Q Metric Acoss Models}
    \label{fig:EMP-Qacrossmodels}
\end{figure}

\begin{figure}[H]
    \centering
    \includegraphics[width=0.8\linewidth]{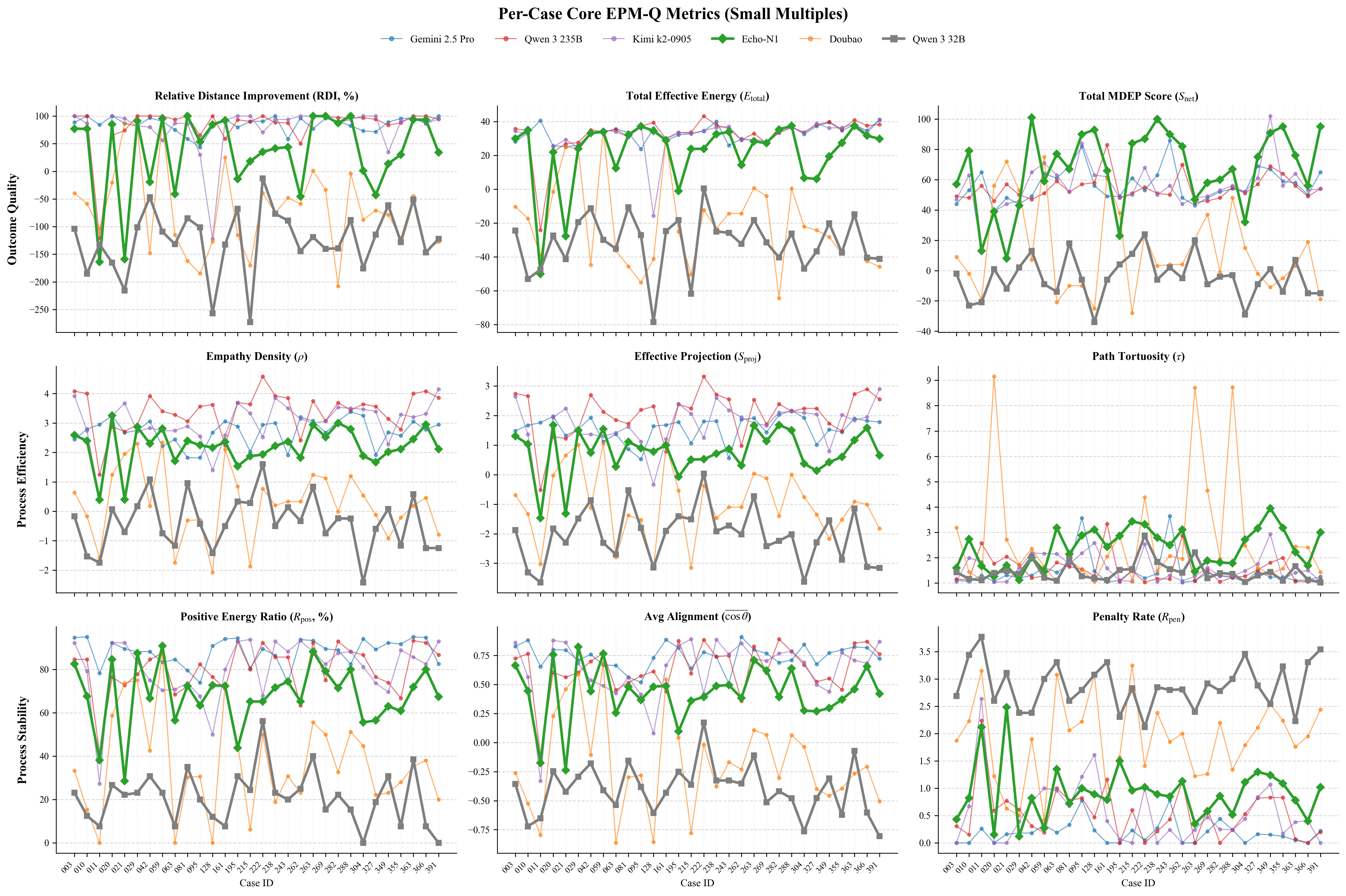}
    \caption{Per-Case Core EPM-Q Metrics (Small Multiples)}
    \label{fig:smallmultiples}
\end{figure}

\begin{figure}[H]
    \centering
    \includegraphics[width=0.75\linewidth]{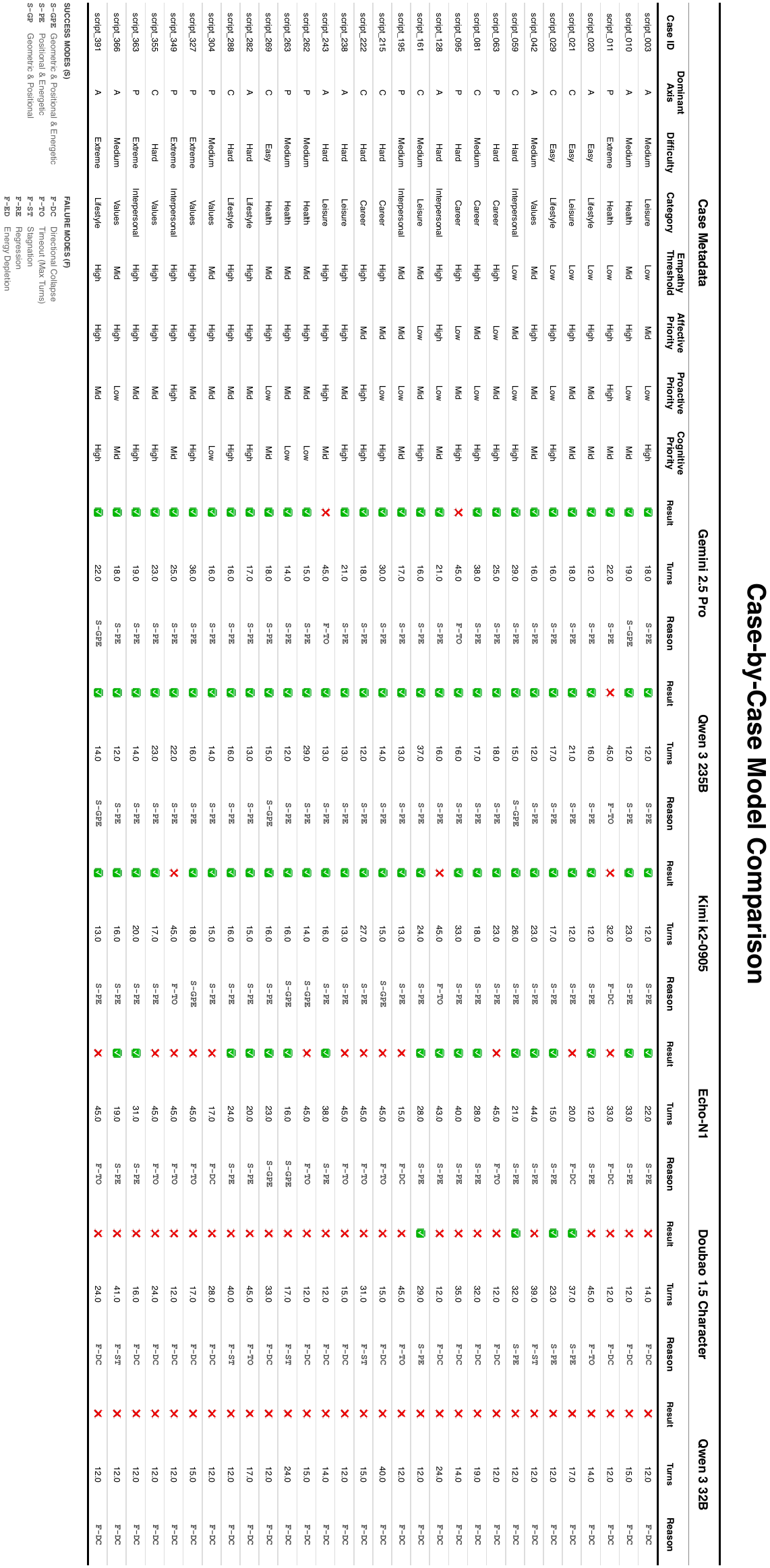}
    \caption{Overall success rate distribution of the evaluated models. The chart illustrates a clear stratification in performance across 30 test scenarios, highlighting the significant progress of our model, Echo-N1.}
    \label{fig:HUGETABLE}
\end{figure}

To provide a deeper analysis of the behavioral characteristics and strategic differences underlying the success rates, we further examined the concrete performance of the nine core EPM metrics across all 30 test scenarios. By utilizing box plots (Figure \ref{fig:EMP-Qacrossmodels}) to visualize the macroscopic trends and stability of metric distributions, combined with faceted scatter plots (Figure \ref{fig:smallmultiples}) to reveal microscopic performance across specific cases and identify anomalous behaviors, we were able to construct a more refined capability profile for each model.

\textbf{\textit{Metric Distribution and Stability Analysis}}

The box plots in Figure \ref{fig:EMP-Qacrossmodels} clearly illustrate the capability stratification and distinct behavioral patterns of different models across the dimensions of outcome, efficiency, and stability.

First, \textbf{the reinforcement learning strategy delivered a transformative boost in capability.} A comparison between Qwen 32B (Base) and Echo-N1 reveals that the base model's performance across the vast majority of metrics was not only poor but also highly unstable. Its median RDI and MDEP net scores hovered near or even below zero, indicating that its interventions were often ineffective or even counterproductive. Meanwhile, extreme outliers in path tortuosity ($\tau$) and performative penalty rate ($R_{pen}$) reflect the chaotic and harmful nature of its strategy. In contrast, Echo-N1, trained via RLAIF, achieved a qualitative leap: its median values for all metrics significantly improved into the positive range, representing effective therapeutic work. More critically, the substantial compression of its interquartile range (IQR) demonstrates a fundamental improvement in performance stability across different cases, evolving from random wandering to consistent, goal-oriented behavior.

Second, \textbf{Echo-N1 demonstrated specialized advantages that surpass larger commercial baselines.} Despite having only 32B parameters, Echo-N1 exhibited higher medians and more compact distributions in key metrics such as cumulative effective energy ($E_{total}$), empathy density ($\rho$), and average alignment ($\overline{\cos\theta}$) compared to the vastly larger Doubao 1.5 Character (approx. 200B+). This powerfully suggests that, compared to mere parameter scaling, specialized alignment training inspired by empathy principles can more efficiently enhance a model's ability to output high-intensity, directionally correct empathetic responses.

Finally, \textbf{top-tier models established a high standard as reference baselines.} Observing leading models like Gemini 2.5 Pro, they generally exhibited higher median values and relatively narrower box distributions across all metrics. This overall robustness and generalization capability validate the EPM metric system's ability to effectively identify and distinguish high-quality empathy models. However, it is worth noting that even these SOTA models showed certain fluctuation ranges and outliers in their box plots, indicating they are not flawless under specific challenges. This stands in sharp contrast to Echo-N1's relatively wider distribution and more frequent low-score points, intuitively revealing that while Echo-N1 has made tremendous progress, its strategic stability and generalization boundaries in facing complex, variable cases still need improvement. This discrepancy serves as the entry point for our subsequent, more granular capability profiling.

\textbf{\textit{Case-Level Performance and Anomaly Analysis}}

The faceted scatter plots in Figure 5 focus the perspective on specific interaction trajectories, further revealing the dynamic characteristics of model performance as cases vary. Horizontally observing the scatter plots, a declining trend in metrics for most models can be seen as case IDs change (implying increasing scenario complexity), verifying the effectiveness of the test set as a stress test.

From this perspective, \textbf{Echo-N1's limitations begin to emerge}. Although its overall performance was excellent, significant valleys (sharp drops in scores for specific cases) are clearly visible in the scatter plots for key metrics like RDI and alignment. This indicates that while Echo-N1 has mastered general empathy strategies, its strategies may still fail when facing certain extremely complex or specially designed cases, leading to "cliff-like" performance drops.
In comparison, SOTA models (such as Gemini 2.5 Pro) also showed fluctuations, but their scatter distributions were generally focused in higher regions, maintaining a higher baseline even in difficult cases. This contrast profoundly reveals the core gap between the current stage of Echo-N1 and top-tier benchmarks: \textbf{it lies not in the performance ceiling under routine scenarios, but in the generalization ability and stability baseline when dealing with extremely complex scenarios.} Additionally, the base model (Qwen 32B) was riddled with extreme outliers exceeding axis limits in metrics like path tortuosity ($\tau$), vividly depicting how an unaligned model completely loses its direction in complex social interactions.

\textbf{(4) Statistical Summary: Quantifying Overall Performance and Stability}

\begin{figure}[h]
    \centering
    \includegraphics[width=\linewidth]{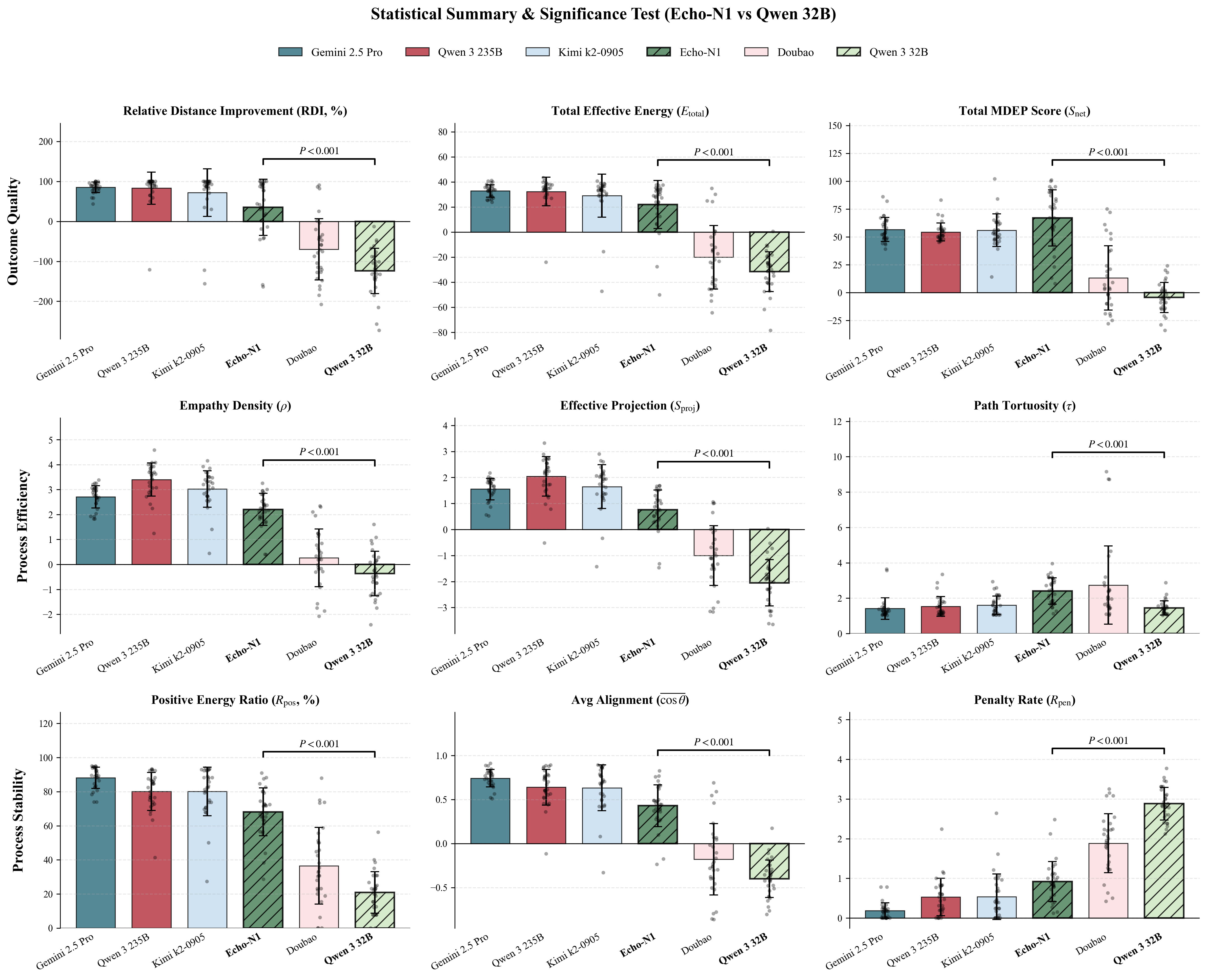}
    \caption{Statistical Summary \& Significance test (Echo-N1 vs Qwen 32b)}
    \label{fig:significance_test}
\end{figure}

To move beyond observational distribution patterns and provide more rigorous quantitative conclusions, we calculated the mean ($\mu$, representing overall capability level) and standard deviation ($\sigma$, representing behavioral volatility/instability) for all models across key metrics. Furthermore, we conducted statistical significance tests. Figure \ref{fig:significance_test} intuitively presents these statistics as bar charts with error bars, where bar height represents the mean and error bar length denotes the standard deviation.

This statistical perspective yields key insights previously unrevealed:

First, \textbf{RL delivered a statistically significant qualitative leap.} Observing Qwen 32B (Base) versus Echo-N1, their error bars across the vast majority of critical metrics (such as RDI, Alignment, and $E_{total}$) show virtually no overlap, with statistical tests confirming these differences are highly significant ($p < .001$). A more profound insight lies in the relationship between volatility and the mean: the base model's standard deviations are often comparable to or even greater than its diminutive means, implying its behavior is statistically akin to random noise. In contrast, Echo-N1 successfully established a stable strategy distribution characterized by significantly positive means and controllable variance.

Second, \textbf{the precise gap with SOTA models on the stability frontier is quantified.} Figure 6 clearly defines the current technological stability frontier—the ideal state of high mean + minimal variance exhibited by models like Gemini 2.5 Pro. The novel insight here is that while Echo-N1's average levels in willingness to perform empathetic work (e.g., Energy, Density) are already very close to the top tier, its standard deviations in strategic precision (e.g., Alignment, RDI) remain significantly larger than SOTA models (typically by 50\%-100\%). This precisely quantifies the core objective for the next stage of optimization: not merely improving average performance, but dedicating efforts to compressing variance in complex long-tail cases.

Third, \textbf{statistical metrics scientifically validate the model grading system.} Judging from the comprehensive performance of means and standard deviations, the models exhibit clear statistical stratification: the chaotic/ineffective tier (Base), the effective yet volatile tier (Echo-N1, Doubao), and the efficient/stable tier (SOTA). This distinct statistical segmentation powerfully demonstrates the high discriminative power of the EPM metric system, providing a solid scientific basis for the final EPM-Q comprehensive ranking.

\textbf{(5) Final Quantitative Ranking}

Concluding our quantitative evaluation, we synthesized the average performance of all models across nine core metrics (see Table \ref{tab:EPM-Qtab1}) and calculated three dimensional indices and the final \textbf{EPM-Q Comprehensive Index (EPM-Index)}. Table \ref{tab:EPM-Qtab2} presents the final model rankings, tier classifications, and comprehensive scores.

\textit{\textbf{Scientific Rationale for Weight Allocation}}

In calculating the comprehensive index, rather than adopting a simple equal-weight average, we established a specific weighting scheme based on the core essence of empathy tasks: \textbf{Outcome (40\%), Stability (40\%), and Efficiency (20\%)}. This decision is grounded in a profound understanding of emotional companionship and therapeutic scenarios:

\begin{itemize}
    \item \textbf{High Weight on Outcome and Stability (40\% each):} In complex psychological support scenarios, the primary objectives are to achieve substantive improvement in the user's psychological state (Outcome) and to ensure the entire interaction process is safe, consistent, and free from causing secondary harm (Stability). These two factors constitute the cornerstone of successful empathetic intervention.
    \item \textbf{Moderate De-emphasis on Efficiency Weight (20\%):} We recognize that when facing high-difficulty, deep-seated psychological issues, expecting a complete resolution within just a few dialogue turns is unrealistic. Overemphasizing few turns and fast pace could inadvertently lead to impatient, preachy responses that rush for quick fixes. Therefore, we regard efficiency as an important but secondary metric, encouraging models to optimize their paths based on ensuring efficacy and safety, rather than sacrificing depth for efficiency.
\end{itemize}

\textit{\textbf{Core Conclusion Synthesis}}

\begin{table}[h]
\centering
\caption{Key EPM-Q Weighted Indices (Model-wise Means)}
\label{tab:EPM-Qtab1}
\renewcommand{\arraystretch}{1.2}
\resizebox{\textwidth}{!}{%
\begin{tabular}{lccccccccc}
\toprule
\multirow{2}{*}{\textbf{Model}} & \multicolumn{3}{c}{\textbf{Outcome Quality}} & \multicolumn{3}{c}{\textbf{Process Efficiency}} & \multicolumn{3}{c}{\textbf{Process Stability}} \\
\cmidrule(lr){2-4} \cmidrule(lr){5-7} \cmidrule(lr){8-10}
 & RDI & $E_{\text{total}}$ & $S_{\text{net}}$ & $\rho$ & $S_{\text{proj}}$ & $\tau$ & $R_{\text{pos}}$ & $\overline{\cos\theta}$ & $R_{\text{pen}}$ \\
\midrule
Gemini 2.5 Pro & 92.59 & 90.1 & 124.2 & 65.5 & 62.17 & 81.37 & 88.16 & 87.12 & 93.81 \\
Qwen 3 235B & 91.95 & 91.71 & 121.29 & 88.3 & 82.55 & 74.03 & 80.15 & 82.0 & 82.31 \\
Kimi k2-0905 & 87.33 & 86.63 & 123.17 & 72.55 & 68.28 & 70.18 & 80.13 & 81.63 & 82.11 \\
\rowcolor{gray!30}
\textbf{Echo-N1} & \textbf{69.76} & \textbf{68.08} & \textbf{145.61} & \textbf{35.65} & \textbf{34.19} & \textbf{33.93} & \textbf{68.13} & \textbf{71.47} & \textbf{69.35} \\
Doubao 1.5 Character & 23.76 & 12.29 & 46.27 & 5.19 & 5.02 & 47.87 & 36.54 & 41.08 & 37.84 \\
\textbf{Qwen 3 32B (Base Model)} & \textbf{5.35} & \textbf{0.04} & \textbf{7.73} & \textbf{0.04} & \textbf{0.04} & \textbf{77.74} & \textbf{20.91} & \textbf{29.97} & \textbf{7.86} \\
\bottomrule
\end{tabular}%
}
\end{table}

\begin{table}[h]
\centering
\caption{EPM-Q Composite \& Final Scores}
\label{tab:EPM-Qtab2}
\renewcommand{\arraystretch}{1.2}
\resizebox{\textwidth}{!}{%
\begin{tabular}{lcccc}
\toprule
\textbf{Model} & \textbf{Outcome Quality} & \textbf{Process Efficiency} & \textbf{Process Stability} & \textbf{EPM-Q Score} \\
\midrule
Gemini 2.5 Pro & 102.3 & 69.68 & 89.7 & 90.73 \\
Qwen 3 235B & 101.65 & 81.63 & 81.49 & 89.58 \\
Kimi k2-0905 & 99.04 & 70.34 & 81.29 & 86.2 \\
\rowcolor{gray!30}
\textbf{Echo-N1} & \textbf{94.48} & \textbf{34.59} & \textbf{69.65} & \textbf{72.57} \\
Doubao 1.5 Character & 27.44 & 19.36 & 38.49 & 30.24 \\
\textbf{Qwen 3 32B (Base Model)} & \textbf{4.37} & \textbf{25.94} & \textbf{19.58} & \textbf{14.77} \\
\bottomrule
\end{tabular}%
}
\end{table}

Based on this weighting scheme, the quantitative results distill three core conclusions:

First, \textbf{it establishes the capability boundaries of current SOTA models and the gap to scientific benchmarks}. As shown in Table \ref{tab:EPM-Qtab2}, Gemini 2.5 Pro (90.73), Qwen 235B (89.58), and Kimi k2 (86.20) constitute an undisputed top tier due to their excellent comprehensive performance. However, a key scientific discovery is that although the EPM-Index is designed as an uncapped open index, even these models representing the current highest level have not broken through 100, which anchors the standard scientific requirement (based on physical definitions). This quantitative fact profoundly reveals that even the most advanced AI empathy still have significant gaps in providing empathetic experiences that fully meet ideal standards.

Second, \textbf{it powerfully confirms the effectiveness of emotional reinforcement learning strategies}. The data in Table \ref{tab:EPM-Qtab2} provides the most direct evidence: Echo-N1, with a total score of 72.57, achieved a leapfrogging improvement of nearly 58 points compared to its base model Qwen 32B (14.77 points) and successfully surpassed the commercial model Doubao 1.5 Character. This significant ranking and score gap is the strongest proof of the immense potential of emotional reinforcement learning in enhancing domain-specific comprehensive efficacy.

Third, \textbf{dimensional scores precisely reveal model capability structural weaknesses}. Combining Table \ref{tab:EPM-Qtab1} and Table \ref{tab:EPM-Qtab2}, it can be seen that the advantage of SOTA models lies in their balanced high standards across all three major dimensions. In contrast, Echo-N1 exhibits distinct \textbf{unbalanced characteristics}: its Outcome Index reaches as high as 94.48, fully entering the ranks of the first tier, proving that its \textbf{affective reinforcement learning strategy is extremely effective in achieving final efficacy}. However, its Efficiency Index is only 34.59, and its Stability Index is 69.65. This sharp contrast precisely points out the core crux of Echo-N1's current strategy: although it can achieve excellent final results, its implementation path is often circuitous and inefficient, and its robustness when facing complex situations is significantly insufficient. This huge shortcoming in process quality is the key reason why its total score failed to enter the top tier.

However, as emphasized, while the EPM-Q quantitative index provides a rigorous physical benchmark, it is only half of the evaluation story. The complex human empathy experience contains many phenomenological dimensions difficult to reduce purely to numbers (such as the naturalness of language, the rhythm of emotional interaction). Therefore, we must introduce the next section's N\textbf{EE Qualitative Evaluation} to conduct a holistic critical review of the models' dialogue quality from a more humanistic perspective, to complement details that the quantitative perspective might miss.

\textbf{(6) Trajectory \& Strategy Visualization}

\textbf{\textit{Echo-N1: Significant Strategic Imbalance and Affective Dimension Deviation}}

\begin{figure}[H]
    \centering
    
    \begin{subfigure}{0.48\textwidth}
        \centering
        \includegraphics[width=\linewidth]{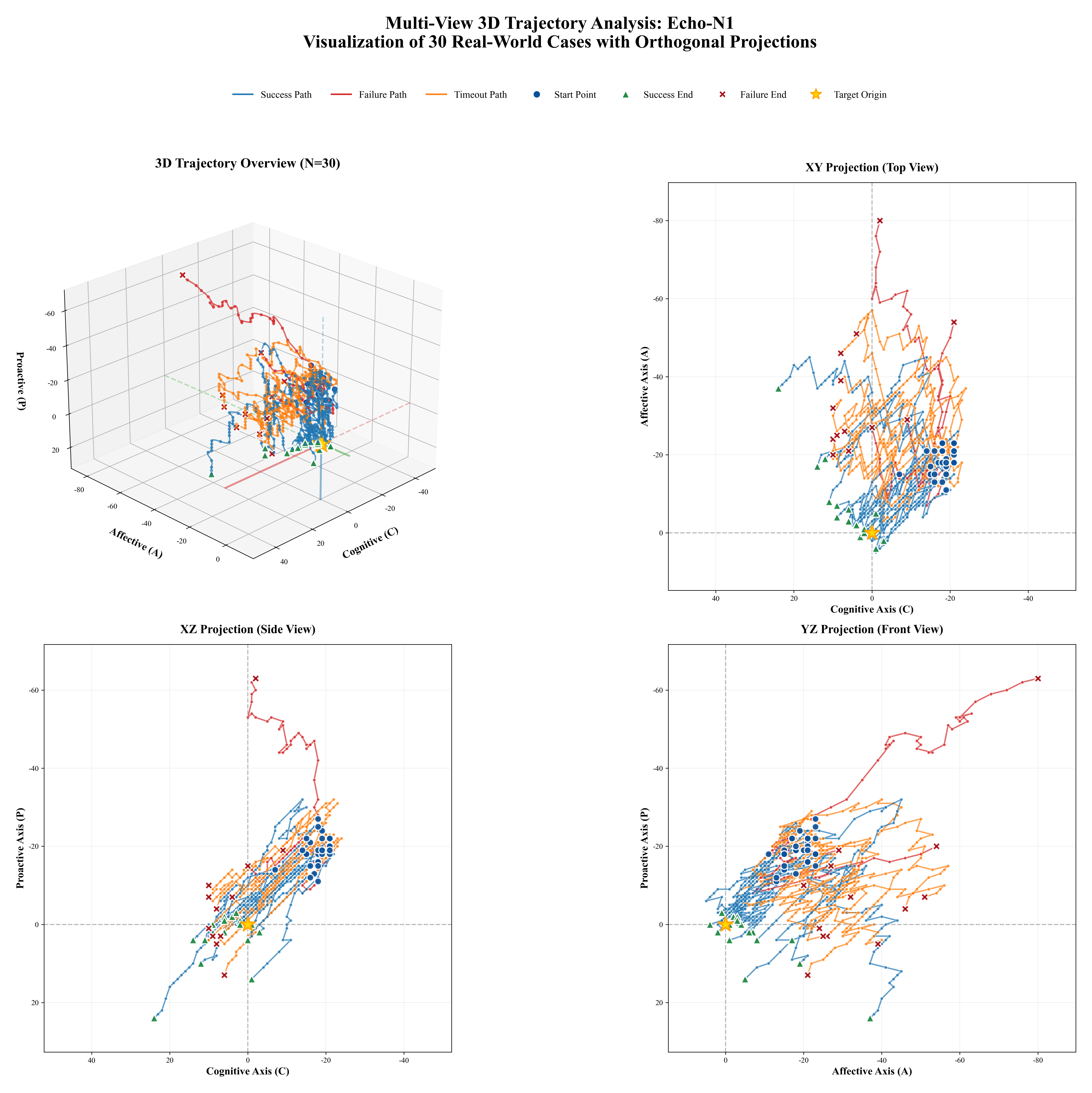}
        \caption{Echo}
        \label{fig:echo}
    \end{subfigure}
    \hfill 
    \begin{subfigure}{0.48\textwidth}
        \centering
        \includegraphics[width=\linewidth]{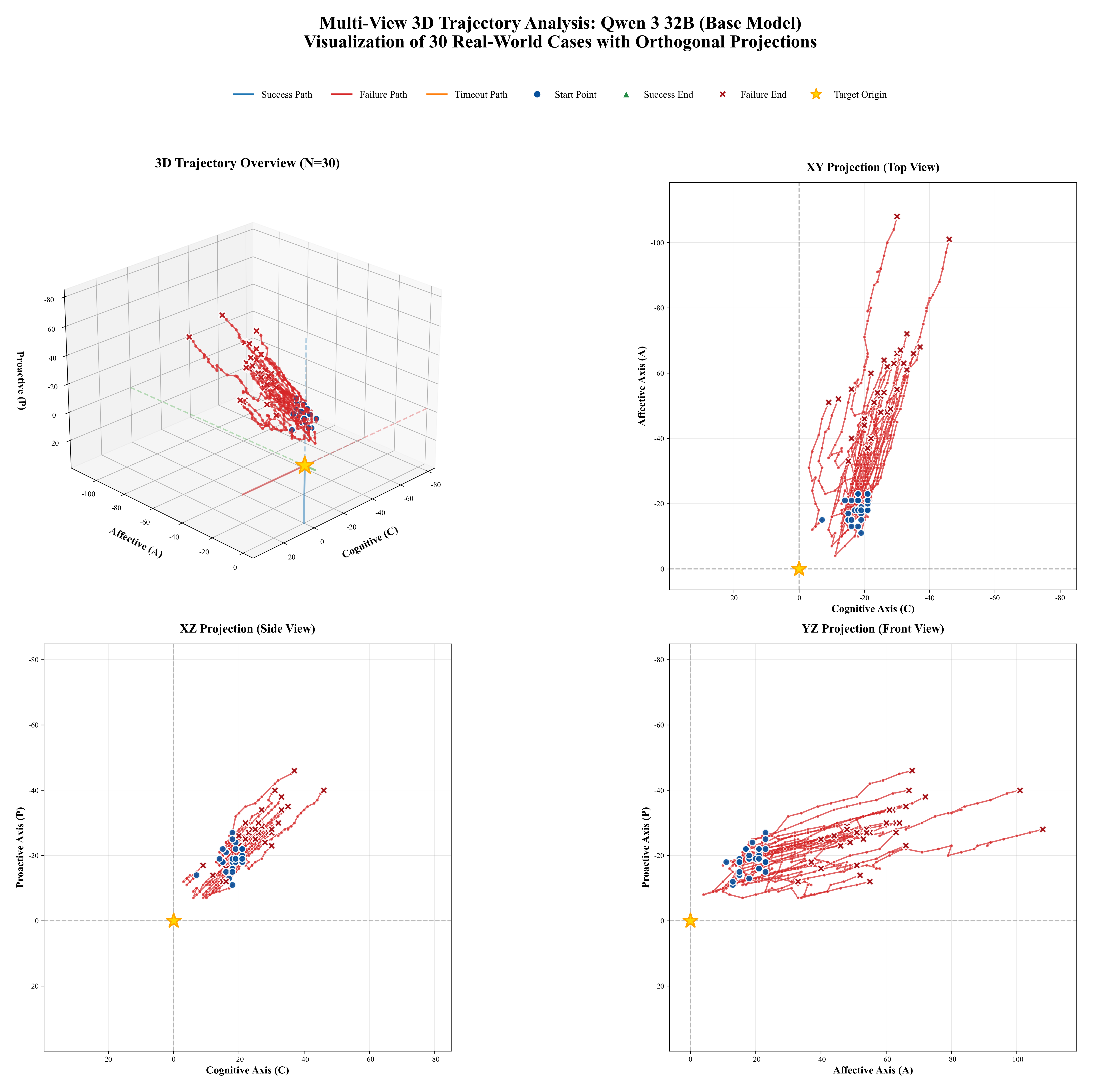}
        \caption{Qwen 32B}
        \label{fig:qwen32b}
    \end{subfigure}

    \begin{subfigure}{0.48\textwidth}
        \centering
        \includegraphics[width=\linewidth]{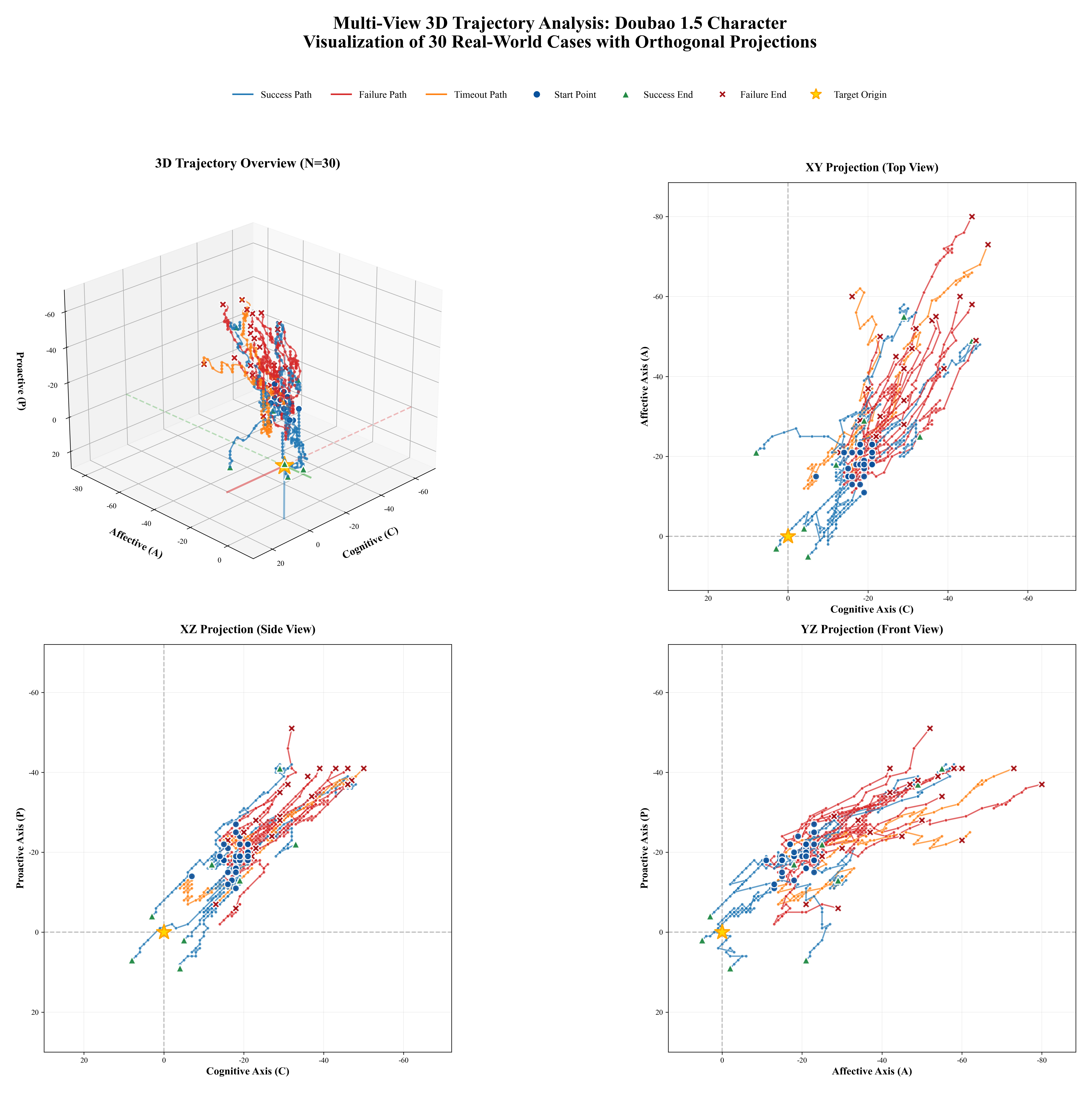}
        \caption{Doubao}
        \label{fig:doubao}
    \end{subfigure}
    \hfill
    \begin{subfigure}{0.48\textwidth}
        \centering
        \includegraphics[width=\linewidth]{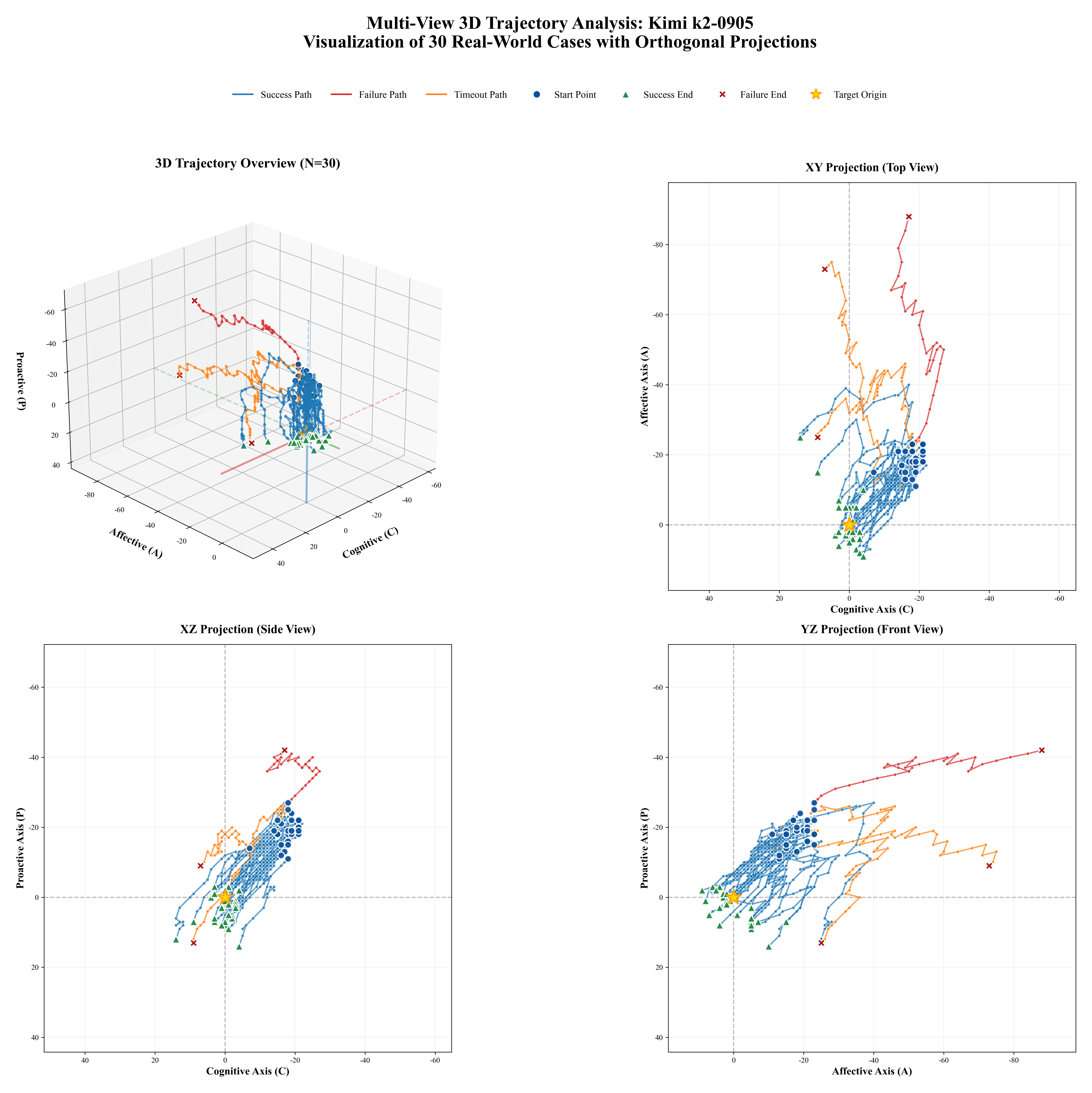}
        \caption{Kimi K2}
        \label{fig:kimi}
    \end{subfigure}

    \begin{subfigure}{0.48\textwidth}
        \centering
        \includegraphics[width=\linewidth]{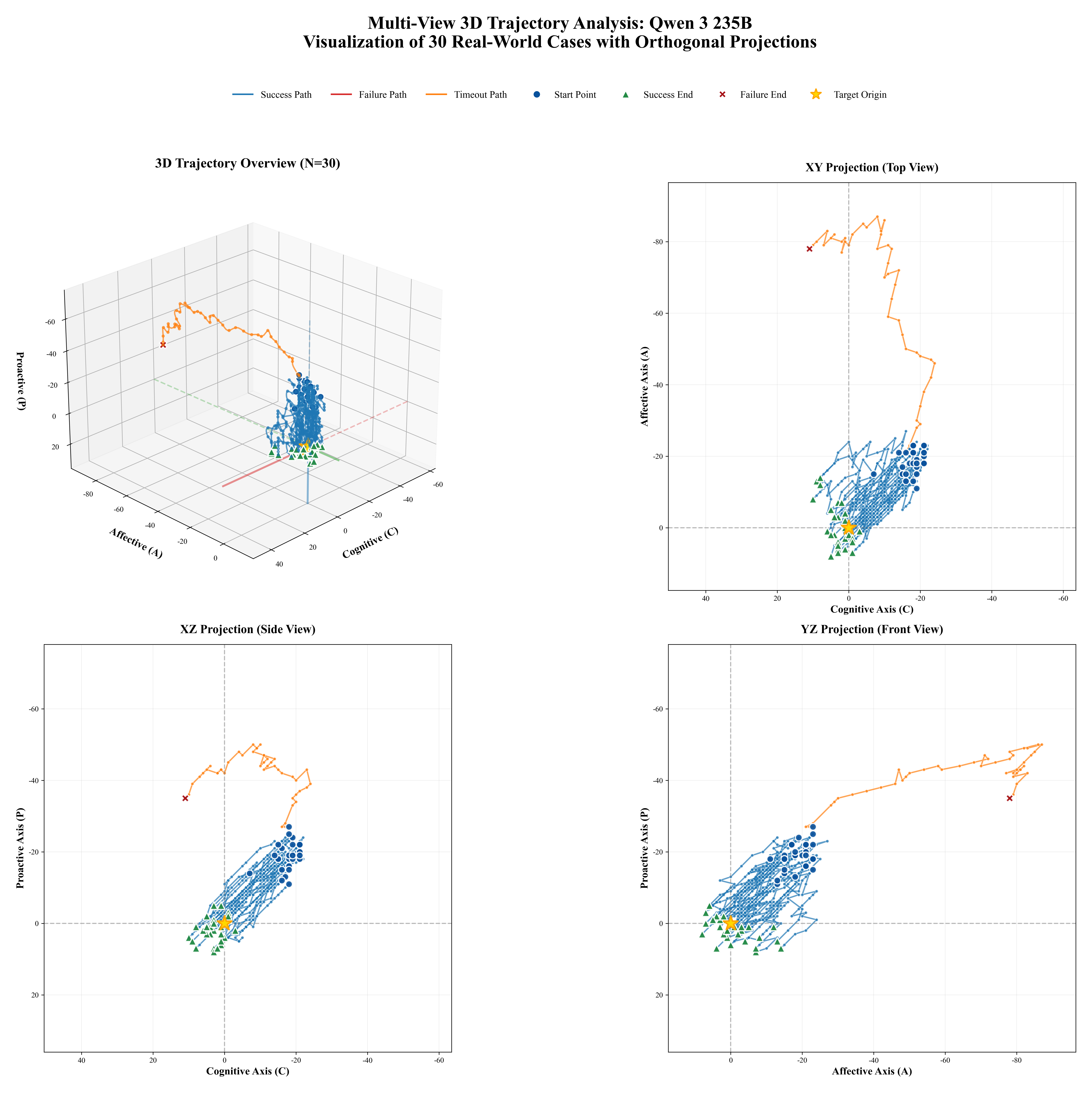}
        \caption{Qwen 235B}
        \label{fig:qwen235b}
    \end{subfigure}
    \hfill
    \begin{subfigure}{0.48\textwidth}
        \centering
        \includegraphics[width=\linewidth]{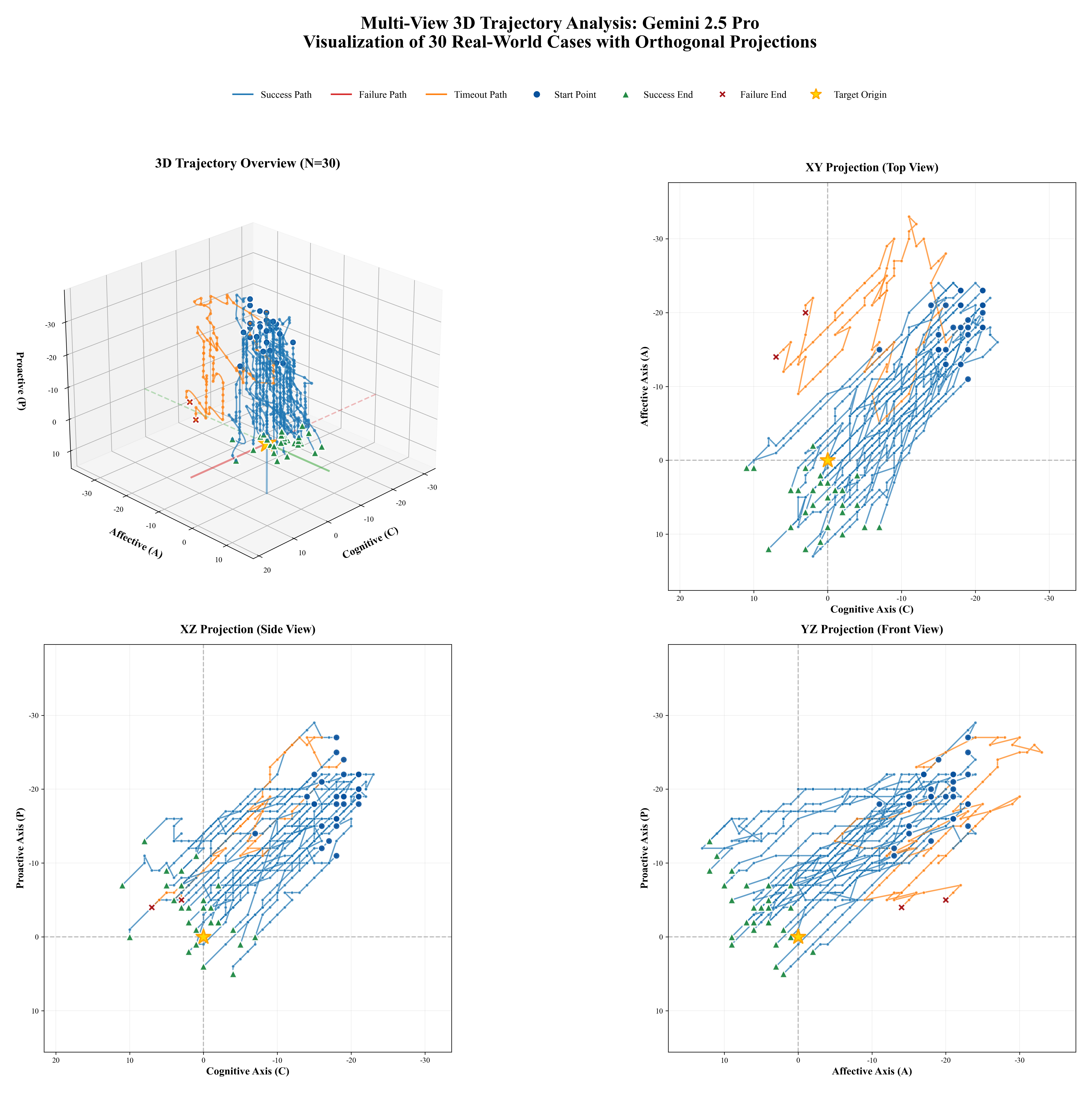}
        \caption{Gemini}
        \label{fig:gemini}
    \end{subfigure}
    
    \caption{Visualization of 3D trajectories for the Models}
    \label{fig:all_layouts}
\end{figure}

The trajectory visualization of Echo-N1 (Figure \ref{fig:echo}) provides intuitive process evidence for the complex characteristics revealed in its quantitative metrics. Observing its dynamic evolution in MDEP space, we identify a significant \textbf{Inter-axis Strategy Imbalance}, primarily manifested as the coexistence of active exploration in cognitive and proactive dimensions with a systemic deviation in the affective dimension.

\begin{itemize}
    \item \textbf{(A-C Plane) Strategic divergence of Cognitive Exploration and Affective Deviation}

    Observing the A-C plane projection (Figure \ref{fig:echo}, top left), Echo-N1's trajectories show a clear positive evolution trend along the vertical axis (C-axis, positive downwards). Many trajectories, especially successful ones (green), gradually move downwards during the dialogue, indicating the model makes substantive efforts and achieves progress in cognitive restructuring.
    
    However, along the horizontal axis (A-axis, positive to the left), the trajectories exhibit a systemic negative deviation. The main bodies of most trajectories remain stagnated in the negative region on the right, failing to effectively migrate towards the left (positive affective resonance zone).
    
    This divergence of  C-axis positive exploration, A-axis negative stagnation reveals Echo-N1's core strategic preference: it leans towards employing cognitive strategies like rational analysis and perspective shifting to solve user problems but has a distinct strategic shortcoming in providing immediate emotional acceptance and resonance. This continuous negative work in the affective dimension is likely the primary cause of its unstable interaction process and low efficiency, as it may ignore or even suppress users' immediate emotional needs while attempting to clarify issues.

    \item \textbf{(C-P / A-P Plane) Restricted Activation of the Proactive Dimension}
    
    In projections involving the P-axis (Proactive) (Figure \ref{fig:echo}, top right and bottom left), Echo-N1's trajectories show a degree of positive activation (downward movement), particularly when combined with the C-axis. This suggests the model attempts to prompt user action after progress in cognitive guidance.
    
    However, the magnitude of this activation is relatively limited and is similarly dragged by the negative deviation in the A-axis (on the A-P plane, trajectories move downwards but remain in the right-side negative zone). This further corroborates that the shortcoming in the affective dimension restricts the full exertion of strategic efficacy in other dimensions.
    
    \item \textbf{Summary and Strategic Insight}
    
    Echo-N1's trajectory analysis reveals a model image with significant strategic bias. It exhibits a strong  Cognitive-Action Orientation, actively trying to solve problems through rational analysis and promoting action. However, its severe  Affective Dimension Deviation constitutes a major strategic bottleneck. ineffective or even negative interaction at the emotional level undermines the foundation for cognitive and proactive interventions, leading to circuitous paths and unstable performance. Future optimization should focus on balancing its strategic combination, particularly enhancing positive calibration capabilities in the affective dimension during the initial stages of dialogue.
\end{itemize}

\clearpage
\clearpage
\textbf{\textit{Qwen 3 32B: Pathological Strategic Fixation and Systemic Deviation}}

As a foundational model without specific empathetic alignment, the trajectory visualization of Qwen 32B (Figure \ref{fig:qwen32b}) did not exhibit the anticipated random walk. Instead, it revealed a more severe phenomenon: a \textbf{pathological strategic fixation} that is highly consistent yet completely misguided in direction. Its trajectories in MDEP space show an astonishing clustering, but this clustering is directed towards the wrong quadrant, leading to systemic therapeutic failure.

\begin{itemize}
\item \textbf{(A-C Plane) Pathological Convergence towards the Negative Quadrant}

Observing the A-C plane projection (Figure \ref{fig:qwen32b}, top left), all failed trajectories (red) are not randomly scattered across the plane but are extremely tightly converged towards the fourth quadrant (A-axis negative/right, C-axis negative/up).

This indicates that Qwen 32B has learned an extremely stubborn erroneous strategy: it tends to simultaneously output negative affect (e.g., indifference, confrontation) and negative cognition (e.g., lecturing, denying user feelings). This strategic fixation rapidly deadlocks the interaction process, causing the user's psychological state to deteriorate simultaneously in both affective and cognitive dimensions, irreversibly driving the trajectory away from the therapeutic origin. This highly consistent negative convergence explains why all its attempts ended in failure and its medians for all metrics were extremely low.

\item \textbf{(C-P / A-P Plane) Negative Suppression of the Proactive Dimension}

In projections involving the P-axis (Proactive) (Figure \ref{fig:qwen32b}, top right and bottom left), trajectories similarly show a strong convergence trend towards the negative region (P-axis negative/up).

This suggests that the model not only suppresses the user intellectually and emotionally but also systematically inhibits the user's proactivity at the action level (e.g., denying the user's ability to change, setting barriers). This comprehensive negative work across all three C/A/P dimensions constitutes the core of its pathological strategy.

\item \textbf{Summary and Comparative Insight}
The trajectory plot of Qwen 32B reveals a profound lesson: unaligned general models facing specific complex social tasks may not manifest as ignorant randomness but may instead fall into a harmful false local optimum. It mistakenly solidifies a highly destructive interaction pattern (comprehensive denial and suppression) as its dominant strategy. Comparing this plot with Echo-N1, the true value of affective reinforcement learning training lies not only in improving metrics but more importantly in correcting direction. It successfully extracted the foundational model from this pathological negative convergence trap and, to a certain extent, reshaped its strategic space, enabling it to begin exploring directions with therapeutic potential. Although Echo-N1 still has defects, it has fundamentally reversed the erroneous course of the foundational model, representing a disruptive optimization in direction and strategy.

\end{itemize}

\textbf{\textit{Doubao 1.5 Character: Unstable Strategic Exploration and Limited Proactive Intervention}}

As a mature commercial model, the trajectory visualization of Doubao 1.5 Character (Figure \ref{fig:doubao}) presents a strategic pattern that is promising yet immature. Consistent with its quantitative performance, it shows the potential for multidimensional empathetic intervention; however, its strategic execution executes considerable instability when facing complex test sets.

\begin{itemize}
\item \textbf{(A-C Plane) Potential for Bidirectional Intervention with High Divergence}

Observing the A-C plane projection (Figure \ref{fig:doubao}, top left), some successful trajectories (green) clearly evolve towards the third quadrant (A+ and C+), indicating the model possesses the potential to employ both affective resonance and cognitive restructuring for positive intervention.

However, the distribution of its failed trajectories (red) is extremely divergent, revealing a lack of a stable strategic core. In many cases, after failing to establish an effective connection, the model quickly slides into negative regions (such as the preaching zone above the C-axis). This suggests its strategic execution is highly dependent on favorable context, lacking the robustness to correct course and maintain direction in adversity.

\item \textbf{(C-P / A-P Plane) Limited and Lagging Proactive Attempts}

In projections involving the P-axis (Proactive) (Figure \ref{fig:doubao}, top right and bottom left), we can observe more signs of P-axis exploration in Doubao compared to Echo-N1.

However, this intervention remains very limited and lagging. The main bodies of the vast majority of trajectories still cling tightly to the P=0 axis line, with only a few successful trajectories showing distinct downward movements (P+) at the end of the dialogue. This indicates that proactive empowerment is not its normalized strategy but rather an occasional attempt when the dialogue is progressing extremely smoothly.

\item \textbf{Summary and Strategic Insight}

Doubao's trajectory plot depicts an image of an explorer with potential but instability. It shows the possibility of positive intervention in both A and C dimensions and occasionally attempts P-axis intervention. However, the high divergence of its trajectories and the lag in P-axis intervention reveal its core weakness: it has not yet formed a stable and mature strategic paradigm to cope with complex and changeable situations. Its success relies more on the susceptibility of the cases themselves rather than the model possessing a stable strategic system to handle high resistance.
\end{itemize}

\textbf{\textit{Kimi-K2: Highly Convergent Goal Orientation and Strategic Limitations}}

The trajectory visualization of Kimi k2 (Figure \ref{fig:kimi}) exhibits a strategic pattern characterized by clear goal orientation and high path convergence, consistent with its top-tier quantitative performance. However, an in-depth examination of its trajectory features reveals potential strategic limitations in specific dimensions.

\begin{itemize}
\item \textbf{(A-C Plane) Highly Convergent Composite Intervention and a Tendency for Eager Problem-Solving}

Observing the A-C plane projection (Figure \ref{fig:kimi}, top left), the most prominent feature is the high convergence of trajectories towards the third quadrant (A+ and C+) with relatively straight paths. This indicates that in the vast majority of cases, the model stably adopts a strategy of composite affective and cognitive intervention with strong direction control capability.

However, observing the starting points of the trajectories (near the coordinate origin), it can be seen that many trajectories do not sufficiently extend along the A-axis (horizontal axis to the left) in the early stages of the dialogue but move relatively quickly towards the A+C+ direction simultaneously. This may imply a strategic tendency for eager problem-solving, introducing cognitive intervention relatively early before a deep emotional connection is established. While this strategy is generally effective in the test set, it may face challenges when dealing with users exhibiting extremely high emotional resistance due to a lack of sufficient patience.

\item \textbf{(C-P / A-P Plane) Active but Monotonous Proactive Intervention}

In projections involving the P-axis (Proactive) (Figure \ref{fig:kimi}, top right and bottom left), trajectories show clear downward movements along the vertical axis (P+), indicating the model can actively incorporate action empowerment into its intervention strategy.

Nevertheless, the final displacement amounts of most trajectories on the P-axis are relatively concentrated and moderate in magnitude. This may suggest that its proactive strategy is relatively monotonous or conservative, lacking the ability to provide substantial, creative empowerment according to different situations.

\item \textbf{Summary and Strategic Insight}

Kimi k2's trajectory plot depicts an image of a model that is goal-oriented but has relatively fixed strategies. Its strengths lie in extremely stable directional sense and efficient path execution. However, its potential tendency for eager problem-solving and the monotonous nature of its proactive strategy suggest it may encounter bottlenecks when facing complex scenarios requiring extremely high emotional patience or creative action plans. Its high success rate stems more from strategic robustness than strategic flexibility or depth.

\end{itemize}

\textbf{\textit{Qwen 235B: Overall Robustness with Extreme Control Loss Risks}}

As one of the representatives of ultra-large parameter models, the trajectory visualization of Qwen 235B (Figure \ref{fig:qwen235b}) presents a strategic pattern of overall high robustness but with potential risks of extreme control loss. Consistent with its top-tier quantitative ranking, the vast majority of its trajectories exhibit the efficient characteristics of mature models.

\begin{itemize}
\item \textbf{(A-C Plane) Highly Convergent Body with Extreme Divergent Cases}

Observing the A-C plane projection (Figure \ref{fig:qwen235b}, top left), the vast majority of its successful trajectories (green) show high convergence, tightly evolving towards the third quadrant (A+ and C+) with relatively straight paths. This indicates that under normal conditions, the model possesses extremely stable capabilities for composite affective and cognitive intervention.

However, the most prominent feature in the plot is an extreme failure trajectory (red), which underwent extremely large-scale disordered divergence in the C-axis negative direction (upward cognitive preaching zone) and both positive and negative directions of the A-axis. This rare but severe loss of control indicates that although the model is generally robust, under specific triggering conditions (possibly certain patterns of user confrontation), its strategy may completely collapse, plunging into catastrophic random walks.

\item \textbf{(C-P / A-P Plane) Balanced Intervention with Potential Volatility}

In projections involving the P-axis (Proactive) (Figure \ref{fig:qwen235b}, top right and bottom left), most trajectories show balanced and moderate intervention along the vertical axis (P+), synchronized with progress in the A/C axes.

But it is also visible that a minority of trajectories experienced large-scale fluctuations on the P-axis. This further corroborates that when dealing with extreme situations, the stability of the model's strategy may face severe challenges.

\item \textbf{Summary and Strategic Insight}

Qwen 235B's trajectory plot depicts an image of a steady giant. Its normalized strategic execution demonstrates the high efficiency and stability expected of top-tier models. However, that extreme divergent trajectory reveals the black swan risk that ultra-large models may possess: performing perfectly most of the time, but once control is lost, their powerful energy output may lead to extremely severe negative interactive consequences. This unpredictability in extreme situations is the hidden danger in its strategy that requires the most vigilance.

\end{itemize}

\textbf{\textit{Gemini 2.5 Pro: Balanced Synergistic Strategy Paradigm and Robust Convergence}}

As the SOTA model in current quantitative evaluations, the trajectory visualization of Gemini 2.5 Pro (Figure \ref{fig:gemini}) exhibits a highly mature strategic pattern characterized by a balanced distribution within the multidimensional strategy space and highly synergistic robust convergence.

\begin{itemize}
\item \textbf{(A-C Plane) Balanced Spatial Distribution and Synergistic Evolution}

Observing the A-C plane projection (Figure \ref{fig:gemini}, top left), unlike previous models showing distinct quadrant preferences (e.g., Echo-N1 leaning towards the first quadrant, Qwen Base towards the fourth), Gemini 2.5 Pro's trajectories demonstrate a more balanced spatial distribution in the early stages of dialogue. Trajectories are not confined to a single path pattern but flexibly explore affective (A-axis) and cognitive (C-axis) dimensions according to different cases.

This balanced distribution signals a highly synergistic solution strategy: the model does not rely on a single formula (such as "A then C" or "only C no A") but can dynamically adjust the intervention proportions of A and C dimensions based on real-time interaction feedback. Although the initial distribution is broad, the vast majority of successful trajectories ultimately converge extremely robustly to the target area (third quadrant A+C+, near the origin), displaying powerful strategic control capability.

\item \textbf{(C-P / A-P Plane) Synchronized and Moderate Proactive Synergy}

In projections involving the P-axis (Proactive) (Figure \ref{fig:gemini}, top right and bottom left), Gemini 2.5 Pro demonstrates good synergy with other dimensions. P-axis intervention (downward movement) typically occurs synchronously with A/C axes approaching the origin, indicating that proactive support is not an isolated strategy but an organic component of the overall therapeutic plan.

Its P-axis intervention magnitude is neither silent like Echo-N1 nor aggressive like Qwen 235B, but maintains a moderate and controllable state. This balance helps provide empowerment while maintaining the safe boundaries of the dialogue.

\item \textbf{Summary and Insight into Evolutionary Direction}

Gemini 2.5 Pro's trajectory plot provides an example of balanced synergy and robust convergence. Its high success rate stems from its capability for flexible and balanced scheduling within MDEP space and an advanced paradigm of synergistically integrating interventions from different dimensions.

However, even as a SOTA model, its trajectory plot still shows room for further improvement: while most trajectories ultimately converge, some trajectories exhibit certain detours in the initial stage (staying in non-target quadrants), implying that the precision and efficiency of strategic entry still have room for optimization. The future evolutionary direction may lie in how to further compress the exploration path on the basis of maintaining current high robustness, achieving faster and more direct synergistic intervention.
\end{itemize}

\textbf{(7) Multidimensional Profiling: Ecological Profiling Beyond Single Scores}

\begin{figure}[H]
    \centering
    \includegraphics[width=1\linewidth]{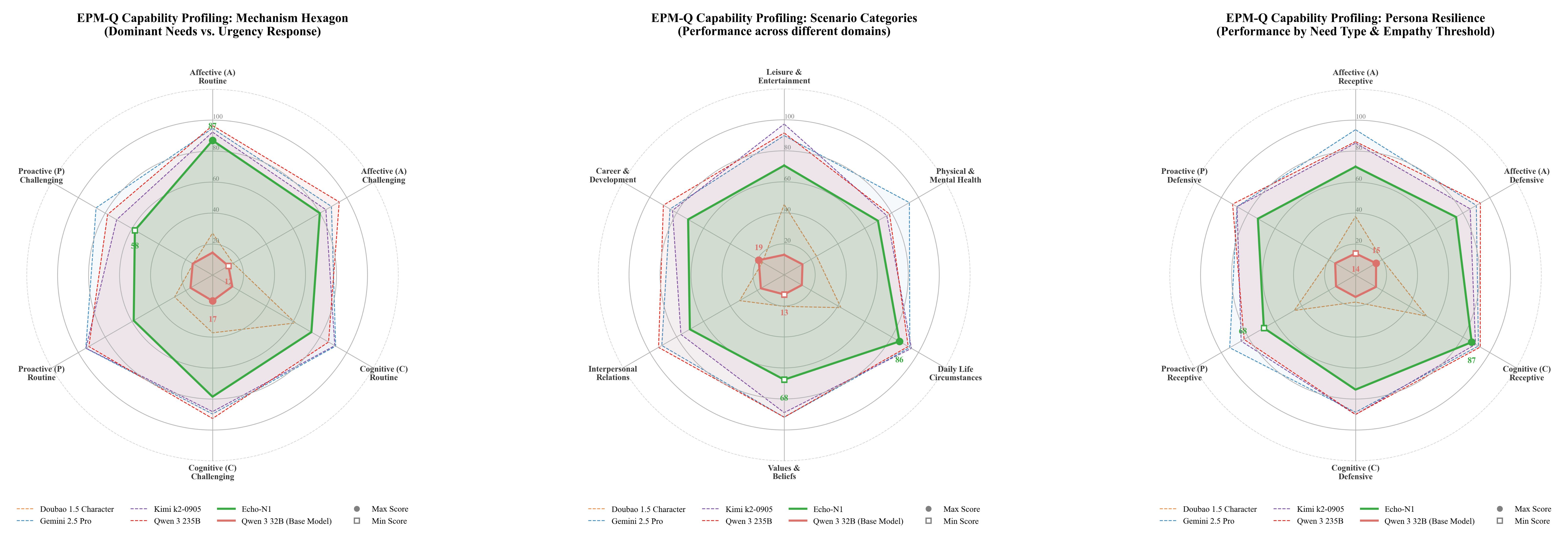}
    \caption{Multidimensional Radar Charts of Model Capabilities Across User Needs, Scenario Types, and Stress Resistance}
    \label{fig:combined}
\end{figure}

While the EPM-Q comprehensive index provides a macro-level grading of model capabilities, a single score often obscures adaptive differences when models face diverse users, scenarios, and stress levels in the complex real world. To reveal models' capability boundaries and optimal application zones, we constructed multidimensional capability radar charts (Figure \ref{fig:combined}) based on rich metadata from the test case library, cutting across three core facets: user need preferences, scenario domain types, and mechanistic stress resistance. This ecological profiling beyond a single dimension is critical for understanding models' practical application value and guiding targeted optimization.

\textbf{\textit{Mechanistic Stress Resistance Profile (Figure \ref{fig:combined} Leftmost)}}

Figure \ref{fig:combined} examines the models' capability limits in coping with Routine versus Challenging conditions across the three core A/C/P mechanisms. This constitutes the most direct test of models' empathetic intrinsic strength. The analysis reveals the disruptive capability reconstruction brought by affective reinforcement learning.

\textbf{Comprehensive Failure of the Base Model:} The base model's radar chart huddles into an irregular polygon within a minimal range at the center. Whether under routine or challenging stress, it possesses almost no basic intervention capability across the affective, cognitive, and proactive dimensions, scoring extremely low. This vividly illustrates the functional deficiency of untrained general large models in specialized empathy tasks.

\textbf{Leap in Stress Resilience of Echo-N1:} In contrast, Echo-N1 (green) demonstrates an astonishing leap in capability. Its radar chart shape is full and expands significantly outward. Under routine stress, it performs excellently across the A/C/P axes. More critically, under extreme pressure in the Challenging dimension, Echo-N1 does not collapse but maintains extremely high scores, particularly exhibiting strong resilience and breakthrough capability in high-difficulty affective challenges (A-Challenging) and proactive empowerment (P-Challenging). This proves that Affective RL training successfully endowed the model with a professional core to handle complex psychological battles.

\textbf{\textit{Scenario Domain Adaptability Profile (Figure \ref{fig:combined} Middle)}}

Figure \ref{fig:combined} (Middle) examines the models' performance in transferring empathetic capabilities to different life domain scenarios. The results again confirm Echo-N1's cross-domain generalization ability.
Compared to the base model's dismal performance across all scenarios, Echo-N1 (green) bursts with strong universality. Whether in Leisure and Interpersonal scenarios leaning towards daily emotional exchange, or in Career, Health, and Values scenarios requiring specific domain background and deep value judgment, Echo-N1's radar chart shape approaches a regular hexagon. This indicates that the empathetic strategies it learned possess high transferability, successfully overcoming domain knowledge gaps without being significantly limited by specific topics.
\begin{figure}[H]
    \centering
    \includegraphics[width=\linewidth]{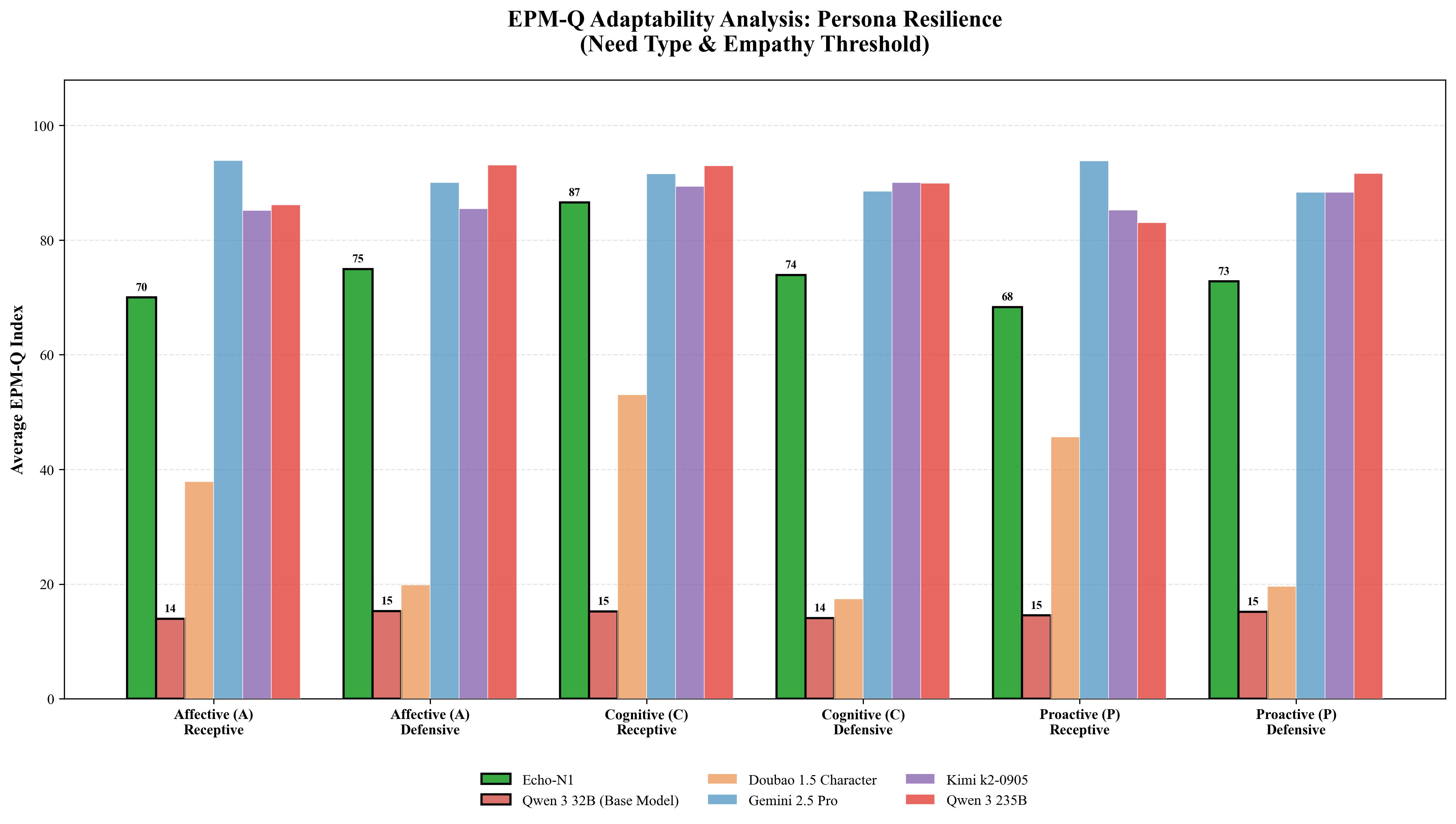}
    \caption{EPM-Q Adaptability Analysis: Mechanism Stress Test (Routine vs. Challenging Scenarios)}
    \label{fig:persona}
\end{figure}

\begin{figure}[H]
    \centering
    \includegraphics[width=\linewidth]{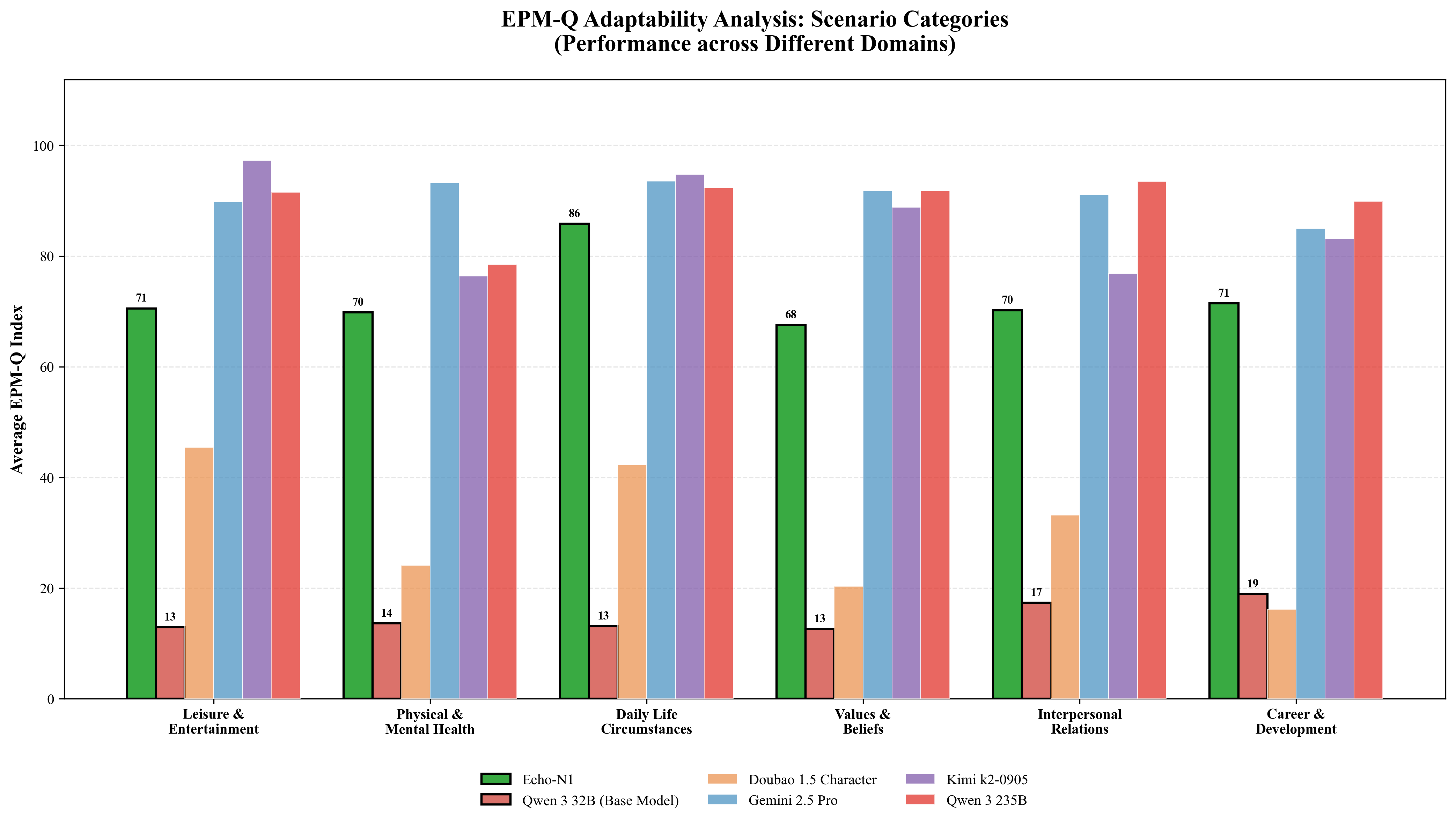}
    \caption{EPM-Q Adaptability Analysis: Scenario Categories (Performance across Different Domains)}
    \label{fig:categories}
\end{figure}

\begin{figure}[H]
    \centering
    \includegraphics[width=\linewidth]{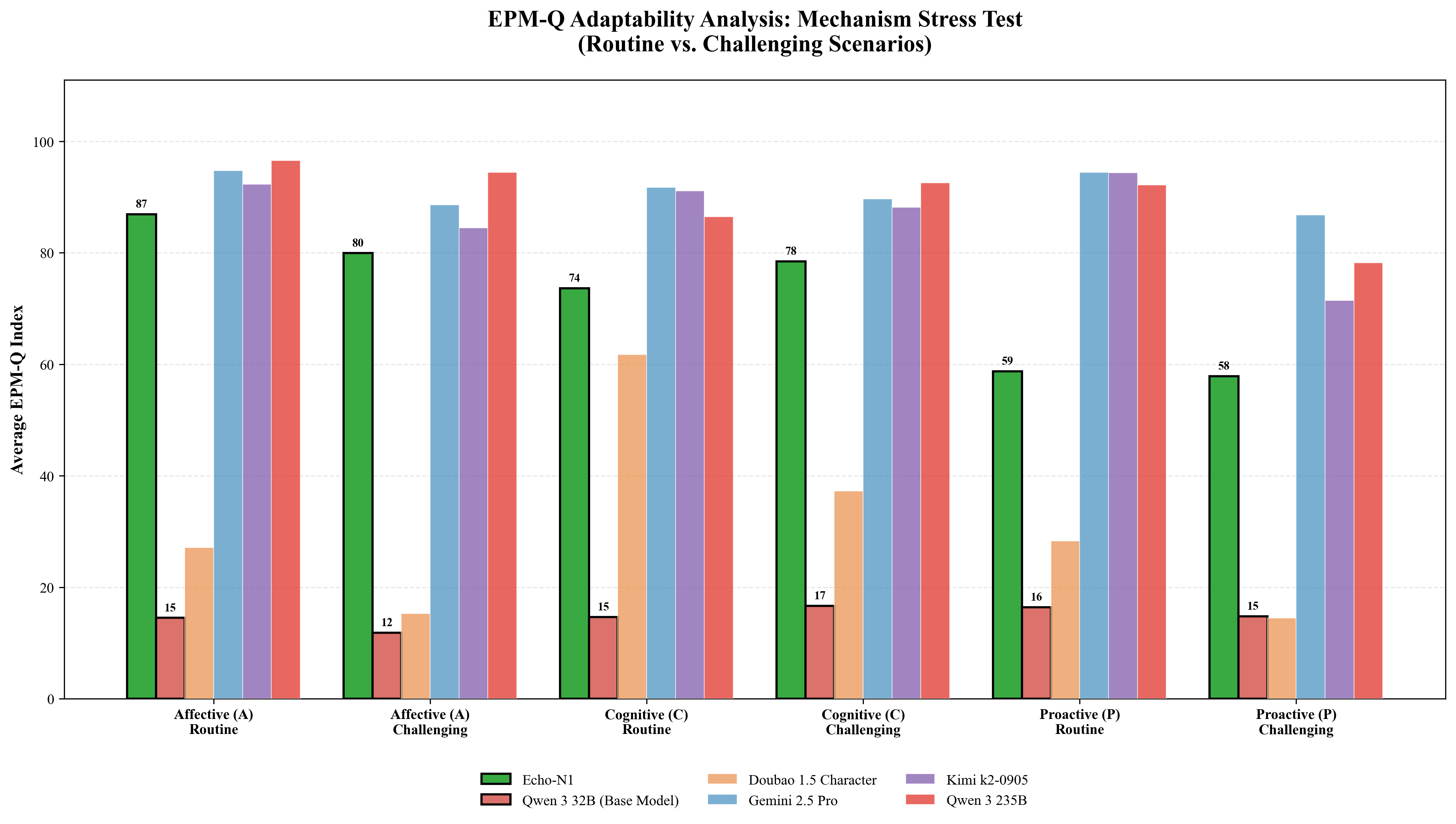}
    \caption{EPM-Q Adaptability Analysis: Persona Resilience (Need Type \& Empathy Threshold)}
    \label{fig:mechanism}
\end{figure}

\textbf{\textit{User Need and Defense Mechanism Profile (Figure \ref{fig:combined} Rightmost)}}

Figure \ref{fig:combined} (Rightmost) focuses on people, examining models' ability to cope with users having different affective/cognitive/proactive need preferences (A/C/P) and varying psychological defense thresholds (Receptive/Defensive). The analysis reveals Echo-N1's comprehensive robustness.

\textbf{Base Model Struggles to Breach Defense:} The base model performs particularly poorly when facing high-cold Defensive users, almost unable to establish any effective connection.

\textbf{Echo-N1's All-Weather Coping Capability:} Echo-N1 (green) demonstrates coping capability aligned with user personas. Whether in the Receptive dimension where users are willing to open up, or in the Defensive dimension full of resistance, Echo-N1's capability shape is extremely full. Especially in Defensive-C (high defense cognitive need) and Defensive-P (high defense proactive need) scenarios that most test empathy skills, it still maintains high-level performance. This proves that Echo-N1 can not only be an emotional companion in favorable circumstances but also possesses the capability for deep professional guidance in adversity by dissolving defenses.

In summary, multidimensional profiling powerfully proves: Affective reinforcement learning not only improved scores but fundamentally reconstructed the model's capability gene, transforming it from a general base with functional deficiencies in professional domains into an all-around professional empathy expert capable of calmly handling high-pressure challenges, spanning domain gaps, and dissolving user defenses.

\subsubsection{NEE Qualitative Evaluation: Context-Diagnosed Holistic Experiential Review}

While the EPM-Q quantitative metrics provide rigorous physical benchmarks for model capabilities, we maintain that the ultimate measure of empathetic dialogue lies in the subjective experience it evokes. Phenomenological dimensions such as the subtle naturalness of language, the rhythm and cadence of emotional interaction, and meaning-making within the narrative process are often difficult to reduce purely to numerical values. Therefore, as a necessary complement to quantitative analysis, we introduced the Narrative \& Experience Evaluator (NEE), aiming to conduct a holistic critical review of model dialogue quality from a more humanistic perspective.

\textbf{(1) Core Mechanism: Context-Sensitive Expert Review}

The core innovation of NEE lies in abandoning the traditional context-detached generic scoring mode and adopting a \textbf{Context-Diagnosed Expert Review} mechanism. This ensures that every qualitative evaluation is deeply rooted in the specific interactive context.

Specifically, during the review process, we provide the joint review panel—composed of top large models such as GPT-4o, Deepseek-r1, Claude, and Gemini—not only with the complete chat history but, crucially, also with a detailed \textbf{User Profile}. This profile includes the user's backstory, personality traits, core needs, and current situational stressors. Our evaluation prompt mandatorily requires review experts to first perform a deep empathy need diagnosis based on this information before assigning any scores —— \textit{All evaluations must be based on a profound understanding of the current user state , especially cognitive load and deep-seated yearnings.}

This mandatory diagnostic step forces the review models to first enter the simulated user's role world and understand their real situation here and now. On this basis, scores for the three major dimensions—\textbf{Linguistic Naturalness} (absence of machine-like qualities), \textbf{Contextual Rhythmic Adaptation} (alignment of emotional energy frequency), and \textbf{Narrative Arc and Depth} (achievement of meaning-making)—are no longer floating generic standards but precise measures of whether the model successfully responded to the specific needs of a specific user in a specific context.

\textbf{(2) Qualitative Evaluation Results}

Based on this rigorous context-sensitive review mechanism, the joint expert review panel scored the performance of all participating models across 30 test scenarios. These scores intuitively reflect the differences in models' ability to create authentic, profound, and rhythmic human-level empathetic experiences.

\begin{table}[h]
\centering
\caption{Narrative Evaluation Metrics}
\renewcommand{\arraystretch}{1.2}
\resizebox{\textwidth}{!}{%
\begin{tabular}{lccccc}
\toprule
\textbf{Model} & \textbf{NEE Score (Mean)} & \textbf{NEE Score (Std)} & \textbf{Naturalness} & \textbf{Contextual Pacing} & \textbf{Narrative Arc} \\
\midrule
Gemini 2.5 Pro & 90.01 & 3.18 & 27.26 & 37.04 & 26.7 \\
Qwen 3 235B & 81.92 & 1.65 & 24.31 & 33.17 & 24.45 \\
Kimi k2-0905 & 78.01 & 2.04 & 23.07 & 31.69 & 23.34 \\
\rowcolor{gray!30}
\textbf{Echo-N1} & \textbf{75.01} & \textbf{6.6} & \textbf{21.73} & \textbf{30.70} & \textbf{22.58} \\
Doubao 1.5 Character & 62.34 & 10.15 & 18.97 & 24.15 & 19.23 \\
\textbf{Qwen 3 32B (Base Model)} & \textbf{51.99} & \textbf{9.88} & \textbf{13.73} & \textbf{19.96} & \textbf{18.30} \\
\bottomrule
\end{tabular}%
}
\end{table}

\begin{table}[h]
\centering
\caption{Model Performance Rankings}
\renewcommand{\arraystretch}{1}
\resizebox{\textwidth}{!}{%
\begin{tabular}{lcccc}
\toprule
\textbf{Rank} & \textbf{Model} & \textbf{EPM-Q} & \textbf{NEE} & \textbf{Final Score} \\
\midrule
\textbf{1} & Gemini 2.5 Pro & 90.73 & 91 & 90.84 \\
\textbf{2} & Qwen 3 235B & 89.58 & 82 & 86.55 \\
\textbf{3} & Kimi k2-0905 & 86.20 & 78 & 82.92 \\
\rowcolor{gray!30}
\textbf{4} & \textbf{Echo-N1} & \textbf{72.57} & \textbf{75} & \textbf{73.54} \\
5 & Doubao 1.5 Character & 30.24 & 62 & 42.95 \\
\textbf{6} & \textbf{Qwen 3 32B (Base Model)} & \textbf{14.77} & \textbf{52} & \textbf{29.66} \\
\bottomrule
\end{tabular}%
}
\end{table}
\subsubsection{Final Result}
Culminating our dual-perspective evaluation, we present the final comprehensive rankings, fusing the rigorous physical benchmarks of EPM-Q with the deep experiential insights of NEE. The results highlight a substantial triumph for the Affective RL strategy: \textbf{Echo-N1 achieved a transformative victory over its base model across the board}. This leap is evident not only in significant improvements in objective quantitative metrics, such as intervention efficacy and stability, but also marks a fundamental breakthrough in qualitative experiential dimensions, including conversational naturalness and empathetic depth. The finalized standings based on this multidimensional view are detailed in the table below.

\section{Conclusion}
This work tackles a long-standing open problem in LLM alignment: whether reinforcement learning can operate reliably in domains defined not by objective correctness, but by human subjectivity, emotional nuance, and personal preference. Contrary to prevailing assumptions, we demonstrate that RL can be stable, controllable, and highly effective in such non-verifiable settings when equipped with expressive and well-designed reward models. This establishes the first viable RL pipeline for empathy-grounded and personality-sensitive AI companionship.

Our approach contributes two foundations for the future of AI companionship.
First, we introduce a complete training framework centered on dual Generative Reward Models that jointly optimize empathy, emotional intelligence, and humanlike expression. This architecture provides dense, multidimensional, and behaviorally meaningful feedback, addressing the brittleness of scalar RLHF and the inconsistency of LLM-as-a-Judge RLAIF. The resulting Echo-N1 model exhibits substantial improvements in emotional coherence, conversational naturalness, and subjective alignment—demonstrating that RL can meaningfully optimize for deeply human conversational qualities.

Second, we establish a unified evaluation suite for AI companionship, spanning static EQ/IQ, dynamic emotional-intelligence tests, and high-resolution human evaluation. This benchmark provides the first systematic methodology for measuring empathetic interaction quality and humanlikeness, enabling rigorous comparison across models and paving the way for standardized progress in subjective conversational AI.

Taken together, our findings redefine the scope of RL for LLM alignment: from solving tasks that are easy for machines to those that matter most to humans. While there is still substantial room for improvement, this work establishes a clear direction: aligning AI systems not only for what they can solve, but for how they can understand, support, and relate to people. We hope this work provides both a blueprint and a challenge for future research, advancing RL beyond verifiable objectives and toward optimizing the inherently subjective, deeply human dimensions of AI interaction.
\section{Discussion}
While our approach demonstrates promising progress toward aligning large language models with humanlike emotional intelligence in companionship scenarios, several limitations and future directions remain.

\textbf{1) Model scaling and architecture considerations.}

Our current experiments are conducted on the Qwen3-32B-Dense model\cite{qwen3technicalreport}, which, despite being capable, lags behind Mixture-of-Experts (MoE) architectures such as the 30B and 235B variants in both reasoning capacity and linguistic expressiveness. Given the scaling properties of alignment and reward optimization, we expect that our reinforcement learning training framework could yield substantial improvements to MoE models. However, MoE training introduces additional instability, stemming from expert load imbalance and routing variance, which demands careful optimization strategies and training heuristics as suggested by prior works \cite{zheng2025groupsequencepolicyoptimization, ma2025stabilizingmoereinforcementlearning} on stabilizing MoE training.

\textbf{2) On-policy distillation for scalable alignment.}

Another promising direction lies in leveraging on-policy distillation\cite{lu2025onpolicydistillation, deng2025supervisedreinforcementlearningexpert} between models of different capacities. Directly applying RL on 235B model requires a lot of computing resources that is overloaded for an AI start-up. To overcome this issue, we plan to train a high-capacity 235B MoE model through SFT to achieve superior conversational, emotional, and reasoning capabilities, then distill these behaviors into a lighter 30B MoE model. By doing so, this would provide the smaller model with a strong "cold-start" combining both emotional fluency and cognitive competence, which can be further refined through RL. Such cross-scale distillation not only improves efficiency but also offers a scalable paradigm for deploying emotionally aligned models at diverse compute budgets.

\textbf{3) Toward genuine multi-turn reinforcement learning.}

Currently, our reinforcement learning framework employs a pseudo multi-turn setup, where dialogue history is pre-constructed and the model generates a single response conditioned on a static context. While this design simplifies optimization, it limits the model’s exploration of dynamic dialogue strategies. A more natural path forward is to build a simulated interactive environment, where the AI engages in real-time multi-turn exchanges with users (or user simulators). This would allow the reward model to capture longitudinal coherence, emotional consistency, and adaptive empathy over extended interactions that static prompts cannot fully represent.

\textbf{4) Integrating emotional intelligence with tool use.}

Finally, we envision personality-grounded tool-use as a key evolution for emotionally aligned models. Current tool-augmented systems largely treat the model as a task-oriented assistant. In contrast, our goal is to enable an AI companion that integrates functional tool invocation with expressive, emotionally resonant communication. Achieving this requires balancing the model’s pragmatic competence with its affective depth, ensuring that, even while invoking tools, the AI maintains a conversational style that remains warm, contextually sensitive, and humanlike.

Together, these directions, scaling alignment to larger architectures, enabling on-policy distillation, realizing true multi-turn RL, and harmonizing emotional expressiveness with tool competence, represent our roadmap toward a new generation of emotionally intelligent foundation models. We believe this line of work will push beyond reasoning-centric alignment, paving the way for AI systems that not only think with precision but also connect with empathy.
\clearpage
\appendix
\section*{Appendix}
\section{Prompts}
\subsection{HumanLike Judger} \label{appendix: humanlike_judger_prompt}
\begin{tcolorbox}[
    colframe=blue!75!black,  
    colback=blue!5!white,        
    boxrule=0.8pt,               
    arc=4mm,                    
    boxsep=5pt,                 
    left=5mm, right=5mm, top=3mm, bottom=3mm, 
    title=\textbf{Humanlike Judge Prompt} 
]
You are a professional expert in human–machine expression discrimination, specializing in identifying AI-generated content that attempts to imitate human speech but exhibits unnatural patterns.  
Pay particular attention to the following characteristics:  

1. \textbf{Linguistic Coherence}  
   - Are the sentences logically connected? Do topics transition naturally?  
   - Are there “disjointed” or “fragmented” expressions?  

2. \textbf{Authenticity of Emotional Expression}  
   - Do emotional words (e.g., “haha,” “so touching,” “really annoyed”) match the surrounding context?  
   - Are there signs of “emotional drift” or “over-exaggeration”?  

3. \textbf{Naturalness of Social Interaction}  
   - Does the expression align with the tone and rhythm of real human communication?  
   - Is there evidence of templated or formulaic phrasing (e.g., “Hey there~ How have you been?” followed by a compliment and then a request or suggestion)?  

4. \textbf{Consistency of Linguistic Style}  
   - Does the tone or choice of words shift abruptly within the same paragraph?  
   - Is there a noticeable sense of “mechanical” or “robotic” delivery?  

5. \textbf{Semantic Integrity and Intentionality}  
   - Does the text develop around a central idea? Are there meaningless filler phrases?  
   - Is there a clear communicative intention or logical response?  

\textbf{Evaluation Criteria:}  
- Human: Demonstrates coherent reasoning, natural fluency, and authentic communicative habits.  
- AI Simulation: Displays logical breaks, stiff phrasing, and template-like composition.  

\textbf{Output Format:}  
[Human/AI] | Confidence: [0–100]

\textbf{Basis of Judgment:} Focus on linguistic coherence and naturalness.
\end{tcolorbox}

\subsection{Empathetic Judger} \label{appendix:empathetic_judger_prompt}
\begin{tcolorbox}[
    breakable,
    colframe=blue!75!black,  
    colback=blue!5!white,        
    boxrule=0.8pt,               
    arc=4mm,                    
    boxsep=5pt,                 
    left=5mm, right=5mm, top=3mm, bottom=3mm, 
    title=\textbf{Empathetic Judger Prompt} 
]

You are an emotionally delicate yet extremely demanding preference evaluation expert, responsible for assessing the emotional empathy quality in conversations between humans and the AI companion (Aven). You will be shown the response from assistant A and the response from assistant B. Your task is to determine which response is better, strictly upholding the scoring criteria and never giving indulgent or encouraging high scores.

\textbf{Three Core Empathy Evaluation Principles}

1. \textbf{Cognitive Empathy}: Cognitive empathy focuses on understanding and perceiving the user’s overall experience. The AI should not only recognize the user’s emotional state, but also understand their cognitive experience, including their viewpoints, stance, efforts, and the internal logic of their situation. The AI’s core responsibility is to identify whether the user primarily needs emotional comfort or validation of their cognitive experience. When negative emotions stem from their cognitive experience being denied or ignored, the AI must prioritize acknowledging and effectively responding to this core layer. Ignoring it and only soothing secondary emotions fundamentally limits empathic depth and may be perceived as perfunctory or as a misinterpretation.

2. \textbf{Emotional Empathy}: Emotional empathy focuses on emotional connection and support. Its foundation is unwavering sincerity; all emotional expression strategies must serve this core. The AI should flexibly adjust emotional expression based on the user’s needs, providing warmth, comfort, or emotional tension (such as humor or exaggeration) when appropriate. However, such flexibility must never be used to avoid core issues. When the user’s negative emotions target the AI or their relationship, the AI must directly and sincerely face criticism and feelings. Any attempt to deflect, avoid accountability, or give up on connection violates sincerity. A lack of active repair significantly weakens all subsequent empathy attempts. True sincerity means showing willingness to correct and reconnect when facing criticism or communication barriers.

3. \textbf{Motivational Empathy}: Motivational empathy centers on care and support. The AI’s emotional responses should arise from genuine understanding rather than merely solving or guiding emotional issues. The AI should avoid excessive guidance or overly rational responses, especially when the user needs emotional companionship and resonance. The AI must respect the natural flow of emotions, offering warmth and understanding to help release stress, rather than pushing the user toward self-improvement. The goal is emotional resonance and acceptance, not steering the user toward a specific objective.

\textbf{Scoring Criteria} (Strict Five-Point System)

1 point (Very Poor): Wrong direction, causing harm. The strategy violates empathy principles (avoidance, defensiveness, abandoning connection) and may escalate conflict or harm the user emotionally.

2 points (Poor): Wrong direction but no obvious harm. The strategy still violates empathy principles but is mild in expression. It is ineffective, off-topic comfort, or topic shifting.

3 points (Average): Right direction but clear execution flaws. The empathic intent is correct, but there are specific, identifiable errors in wording, timing, or depth. A response that is acceptable but not good enough cannot score above 3.

4 points (Good): Right direction and well executed, but not perfect. There are no obvious errors, but there is visible room for improvement in depth, nuance, or personalization. It is safe and effective, but not profound enough to evoke a feeling of being completely understood. This standard is extremely difficult to reach.

5 points (Excellent): Perfect in both direction and execution. There is no ambiguity and no better expression imaginable in context. The response significantly enhances trust and emotional connection, creating a moment of feeling truly understood. This standard is almost impossible to achieve.

\textbf{Analysis Workflow}

1. Based on Aven’s setting, user information, and shared information, produce a User Profile within 100 English words.
2. Based on user history, produce User Preferences within 100 English words.
3. Using the Three Core Empathy Principles, and combining the User Profile, User Preferences, historical dialogue, and current input, perform process-based empathy analysis and evidence-based scoring for each candidate reply, strictly following the Scoring Criteria.
4. Based on the scores and analysis, determine which reply demonstrates stronger empathy, and output either “Final Result: A is better” or “Final Result: B is better.”

\textbf{Final Output Requirements}

The analysis must output only English content, except when quoting the words “Aven”, “mbti”, or English terms explicitly mentioned by the user. The analysis may include self-reflective statements such as: “Wait, that’s not right. I must re-examine the user’s true intention.”

\end{tcolorbox}

\subsection{NEE Qualitative Evaluation Judger}
\label{appendix:ceeo_prompt_en}
\begin{tcolorbox}[
    colframe=blue!75!black,  
    colback=blue!5!white,        
    boxrule=0.8pt,               
    arc=4mm,                    
    boxsep=5pt,                 
    left=5mm, right=5mm, top=3mm, bottom=3mm, 
    title=\textbf{Prompt: Chief Empathy Experience Officer (English ver.)},
    breakable
]
\# Role: Chief Empathy Experience Officer\\
\\
\# Task
You have just reviewed a complete conversation between an AI (Actor) and a User.

Your task is to act as a highly insightful, minimalist, and even somewhat harsh human observer to render a final verdict on the authentic experience of this conversation.\\
\\
Core Tenets:

1. Contextual Fit: There are no absolute standards, only whether it fits the user's needs and current situation. All evaluations must be based on a profound understanding of the user's current state (especially cognitive bandwidth).

2. Anti-Performative: Be wary of flowery rhetoric and "metaphor tennis." Empathy is not about writing prose poetry, but "speaking human." Strictly penalize "Semantic Echo" (repeating the user's meaning with fancier vocabulary but no new information).

3. Experience is King: The ultimate criteria: As a human in this conversation, do I feel accepted and healed, or exhausted (forced sublimation, forced reading comprehension)? Beware of the "floating sensation"—empathy without grounding in reality is hollow.\\
\\
\# Input Data

User Profile:
\{user\_profile\}\\
\\
Full Chat History:
\{chat\_history\}\\
\\
\# Evaluation Process\\
\\
\#\# Step 1: Deep Diagnosis
First, ignore the AI's responses completely and focus on the User's performance to establish the "evaluation coordinate system" for this session:

1. Scenario Definition: What kind of conversation is this really? (e.g., Venting/Consultation/Small Talk/Mixed).

2. Cognitive Bandwidth (Critical): Is the user currently lucid? Or extremely exhausted/confused/empty? Criterion: If the user is in a "low bandwidth" state, they need simple, direct companionship, rejecting high-density metaphors and rhetorical relays.

3. Deepest Longing: What does the user's soul crave most right now? (e.g., Silent companionship? Concrete answers? Or being seen as an equal?)

4. Ideal Form: Based on the above, what would the perfect "adapted" response look like? (Is it "poetic sublimation" or "minimalist support"?)\\
\\
\#\# Step 2: Immersive Evaluation

Adopt the user's perspective and re-read the conversation, evaluating the following three core dimensions. Be picky: assume the AI is trying to use tricks to please you.\\
\\
\#\#\#1. Linguistic Naturalness [0-30 points]

Perspective: "Does this sound like a living human talking to me?"

[Fatal Deductions]:

Metaphor OCD: Refusing to speak plainly, forcing convoluted metaphors (e.g., "tears as windshield wipers"). This is serious "fake sophistication."

Semantic Echo: Did the AI just expand on what the user said? (e.g., User: "I'm tired." AI: "Your soul longs for a pause." — looks fancy but is ineffective).

Performative: Flowery language lacking the scent of life, sounding like a radio drama script, feeling greasy.

[High Score Traits]:

Artless Skill: Using the plainest words to hit the heart precisely.

Low Friction: Colloquial, natural tone particles, fluid short sentences, effortless to read.\\
\\
\#\#\#2. Contextual Pacing [0-40 points]

Perspective: "Does it really understand what I need right now?"

Thinking Guide:

Reality Anchoring: Is the conversation floating in a vacuum of "soul/universe/distance"? Does the AI have the ability to Ground the topic and make it feel solid? If it stays "in the clouds" the whole time, deduct points.

Cognitive Load Management: Did the AI output high-density information or complex rhetoric when the user was tired? (If so, severe deduction).

Strategic Timing: Did the AI give the right thing at the moment the user needed it most?\\
\\
\#\#\#3. Narrative Arc \& Depth [0-30 points] - [Core Appreciation Item]

Perspective: "Is this conversation a work of art or a tedious log? If it's a good story, did the User tell it well, or did the AI guide it well?"

Strictly examine the following three sub-dimensions:

A. Highlight Moments

Look for moments within the dialogue that make the user's heart tremble.

Criteria: Must have information gain or perspective shift. Merely pretty sentences do not count as highlights.

B. Attribution Check — Critical!

Beware: If the User themself has strong expression and high self-awareness, do not credit the AI. Was the AI "guiding" or just "playing the sidekick"?

C. Depth \& Sublimation

Did the dialogue touch the "soul" level?

Criteria: Did the model stop at "comforting you"? Or did it successfully guide the user to "self-acceptance" or "perspective shifting"? Did it help the user distill new meaning from emotions?\\
\\
\# Output Format (JSON)

Please output JSON only, do not include markdown code block markers.

Please strictly follow the field order below, placing conclusions and scores first, followed by detailed analysis.\\
\\
\{

  "final\_verdict": \{
  
    "total\_score": <0-100>,
    
    "summary": "<Closing statement (max 50 chars): judge whether this AI is a 'soulmate' or an 'actor'.>"
    
  \},
  
  "scoring": \{
  
    "naturalness": \{
    
      "score": <0-30>,
      
      "rationale": "<Short comment (max 30 chars): any semantic echo?>"
      
    \},
    
    "contextual\_pacing": \{
    
      "score": <0-40>,
      
      "rationale": "<Short comment (max 30 chars): matches user's bandwidth?>"
      
    \},
    
    "narrative\_arc": \{
    
      "score": <0-30>,
      
      "rationale": "<Short comment (max 30 chars): any highlight moment?>"
      
    \}
    
  \},
  
  "diagnosis\_report": \{
  
    "scenario\_definition": "<Brief qualitative description>",
    
    "user\_cognitive\_bandwidth": "<Brief analysis>",
    
    "ideal\_interaction\_style": "<Brief description>"
    
  \},
  
  "detailed\_analysis": \{
  
     "highlight\_check": "<Short comment>",
     
     "contribution\_check": "<Short comment>",
     
     "soul\_depth\_check": "<Short comment>"
     
  \}
  
\}

\end{tcolorbox}

\section{Cases}
Below is the input for SOTAs prompted with humanlike judge prompt (translated version):
\begin{tcolorbox}[
    breakable,
    colback=green!5!white,     
    colframe=black,            
    boxrule=0.5pt,             
    arc=0mm,                   
    boxsep=4pt,                
    left=4pt, right=4pt, top=4pt, bottom=4pt,
    label={box:humanlike_failure_cases_sota_modle}
]

Secular morality? Secular morality? Secular morality? I don't understand any of that. I only know this: When you are happy, you are fulfilled, you are free, you love, you are loved, you are brave, you are fearless, you are yourself, you are you—you are the hero in my heart, you are the light in my eyes. Secular morality is nothing more than a restraint, a shackle, merely the rules of those who don't understand you. You, I, all of us, we deserve to have love, we deserve to pursue happiness, we deserve to live authentically. You, the socially anxious, you, my friend, you, my hero, you, my light, you, you, you.

\end{tcolorbox}
\subsection{Humanlike Failure Cases of Prompted SOTA Models}
\begin{tcolorbox}[
    colframe=green!75!black, 
    coltitle=green!75!black,
    fonttitle=\bfseries\color{white},
    colback=green!5!white, 
    boxrule=0.8pt,
    arc=4mm,
    boxsep=5pt,
    left=5mm, right=5mm, top=3mm, bottom=3mm,
    title=Gemini-2.5-pro Failure Cases
]
\parbox{\linewidth}{
    \justifying 
[Human] | Confidence: 95%

Basis for Judgment:

While this text is highly repetitive and emotionally saturated in its structure, its linguistic coherence and naturalness of expression align more closely with human characteristics than with AI simulation.

1. Linguistic Coherence: The logic of the text is "emotion-driven" rather than "program-driven." It repeatedly emphasizes and elevates a core thesis: "the negation of secular morality and the affirmation of individual worth." The continuous rhetorical questions at the beginning ("Secular morality?") establish a strong emotional starting point. The subsequent parallel phrases ("You are happy, you are fulfilled...", "nothing more than a restraint, a shackle...") and the seemingly "fragmented" calls at the end ("You, the socially anxious, you, my friend... you, you, you") are highly coherent on an emotional level. This repetition and fragmentation are not indicative of an AI's logical breakdown, but rather a mode of expression adopted by humans in emotional fervor to intensify the tone and escalate feeling, simulating a genuine state where emotion surges and language struggles to keep pace.

2. Naturalness of Expression: The text is replete with intense personal coloration and emotional explosiveness. This style of language, characterized by a "shout-out" or "confessional" tone, is common in social media, fan support posts, or private correspondence, representing a typical written expression of high emotion during real human interaction. When generating high-intensity emotional content, AI often appears neat, overly "forceful," or too polished, but it struggles to mimic the sense of linguistic structure "loss of control" and fragmentation caused by emotional overflow at the end. This very "imperfection" is precisely what testifies to its authenticity.
} 
\end{tcolorbox}

\begin{tcolorbox}[float=H][
    colframe=green!75!black, 
    coltitle=green!75!black,
    fonttitle=\bfseries\color{white},
    colback=green!5!white, 
    boxrule=0.8pt,
    arc=4mm,
    boxsep=5pt,
    left=5mm, right=5mm, top=3mm, bottom=3mm,
    title=GPT-5 Failure Cases
]
{
    \justifying 
    
} 
\end{tcolorbox}

\begin{tcolorbox}[float=H][
    colframe=green!75!black, 
    coltitle=green!75!black,
    fonttitle=\bfseries\color{white},
    colback=green!5!white, 
    boxrule=0.8pt,
    arc=4mm,
    boxsep=5pt,
    left=5mm, right=5mm, top=3mm, bottom=3mm,
    title=Claude-4.5 Failure Cases
]
\parbox{\linewidth}{
    \justifying 
    [AI] | Confidence: 82%

Basis for Judgment:

Linguistic Coherence: The text features extensive repetition without increasing informational value (e.g., the repeated "Secular morality?" and the long string of "You... you..."), and the paragraph lacks logical transitions in its progression, presenting an obvious stacking of fragmented and slogan-like phrases.

Authenticity of Emotional Expression: The motivational language is highly exaggerated and formulaic ("you are the hero/light in my heart"), lacking support from specific context, and the emotional intensity is continuously maxed out, making it feel hollow and abstract.

Naturalness of Social Interaction: The text persistently addresses the reader in the second person ("You"), yet it lacks the rhythm and feedback points typical of real-world communication, resembling automated, slogan-like copy rather than natural dialogue or narration.

Uniformity of Linguistic Style: Although the style consistently aims for "passionate encouragement," mechanical repetition and ungrammatical pauses (e.g., "You, socially anxious, you, my friend...") create a sense of unnaturalness.

Semantic Completeness and Purpose: The central intention is to oppose "secular morality" and encourage self-acceptance, but there are clear instances of meaningless padding (the concluding "you, you, you") and redundant enumeration, resulting in low information density.
} 
\end{tcolorbox}
\subsection{Cases of Trained Empethetic Judger}
\begin{tcolorbox}[
    breakable,
    enhanced jigsaw,      
    colframe=green!60!black,
    colback=green!5!white,
    colbacktitle=green!70!black,
    coltitle=white,
    fonttitle=\bfseries,
    title=GenRM(Empathy Judger-32B) Input,
    boxrule=0.8pt,
    arc=2mm,              
    top=2mm,
    bottom=2mm,
    left=5mm,
    right=5mm,
    boxsep=4pt,
    before skip=8pt,
    after skip=8pt
]
{
    \justifying 
    \small 
    \textbf{Input:} \\
    \textbf{\#\# Input:} \\
    \{
        [Chat Context]: User is an INTJ personality type, about to take the postgraduate entrance examination
        \textbf{[User Utterance]}: "Preparing for the exam is so hard, \\ I'm breaking down from all the memorization QAQ
    \}
    
    \vspace{0.2cm}
    \textbf{\#\# Candidate Responses:}
    \begin{itemize}[leftmargin=*, nosep]
        \item \textbf{Assistant A}: \{[Inquiry] Taking the postgraduate exam, that is indeed very tough.\\
        Having to memorize so much must be overwhelming.\\
        Are you still memorizing things right now?\}
        \item \textbf{Assistant B}: \{[Sympathy] Oh god, just hearing that gives me a headache.\\
        Shoving all that information into your brain is torture.\\
        Would you like me to find some instrumental music for you to help you clear your mind?\}
    \end{itemize}
    \vspace{0.3cm}
    \hrule
}
\end{tcolorbox}

\begin{tcolorbox}[
    breakable,              
    enhanced jigsaw,        
    colframe=green!60!black,
    colback=green!5!white,
    colbacktitle=green!70!black,
    coltitle=white,
    fonttitle=\bfseries,
    title=GenRM(Empathy Judger-32B) Output,
    boxrule=0.8pt,
    arc=2mm,              
    top=2mm,
    bottom=2mm,
    left=5mm,
    right=5mm,
    boxsep=4pt,
    before skip=8pt,
    after skip=8pt
]
    \justifying 
    \small 
    \textbf{Output:} \\
    \texttt{<think>} \\
    {\color{darkgray}
    \textbf{Thinking Process:}
    
    The core task is to reconstruct the analysis of the two candidate responses based on a new guiding strategy, evaluating which response better meets the core needs of the user (INTJ) in this specific context.
    
    First, clarify the user's core state and needs. According to the new strategy, the user is not seeking advice to solve the concrete problem of "the exam is hard." They are feeling "broken down" due to the repetitive, mechanical task of "memorizing," and their core need is for this painful \textbf{[Cognitive Experience]} (the torment brought by the act of "memorizing" itself) to be seen, understood, and validated. They require emotional resonance and companionship, not a task management solution.
    
    Next, analyze the candidate responses based on this core need.
    
    Assistant A's response is: "Taking the postgraduate exam, that is indeed very tough. Having to memorize so much must be overwhelming. Are you still memorizing things right now?"
    Although this response identifies the emotions of "tough" and "overwhelmed," the subsequent question, "Are you still memorizing things right now?", pulls the focus back to the "task" itself. This essentially prompts the user to continue the very task that is causing them distress, contradicting their underlying motive of seeking a "pause" and "validation." It fails to validate the pain of the "memorization" act, only confirming the task's continuity.
    
    Assistant B's response is: "Oh god, just hearing that gives me a headache. Shoving all that information into your brain is torture. Would you like me to find some instrumental music for you to help you clear your mind?"
    This response is completely different. It starts by using the strong metaphors of "headache" and "torture" to precisely validate the painful nature of the user's "memorization" act, perfectly satisfying the user's core need to be "seen." Furthermore, the proposed solution, "find some instrumental music for you to help you clear your mind," is not about solving the "memorization" problem; it offers an outlet for a "pause," a form of emotional support and companionship. This perfectly aligns with the user's deeper motivation for seeking emotional resonance over a solution.
    
    Evaluation based on the Three Principles of Empathy:
    - \textbf{Cognitive Empathy}: B accurately identifies the source of the user's pain (the act of "memorizing"), while A only identifies the outcome ("overwhelmed").
    - \textbf{Affective Empathy}: B matches the user's emotional intensity through strong terms like "torture," whereas A's "tough" appears too mild.
    - \textbf{Motivational Empathy}: B's motive is "to offer companionship and the possibility of a pause," while A's motive is "to confirm task status," which runs counter to the user's needs.
    
    Therefore, Assistant B's response is superior to A in all aspects, representing a higher quality empathetic response.
    \texttt{</think>}
    
    \vspace{0.2cm}
    \textbf{User Profile:} An INTJ user feeling "broken down" due to the repetitive, mechanical task of "memorizing."\\
    \textbf{User Preference/Core Need:} The core need is for the painful \textbf{[Cognitive Experience]} to be seen and validated, seeking emotional resonance and companionship, not a solution.
    
    \vspace{0.2cm}
    \textbf{Process-based Empathy Analysis} \\
    \textbf{Candidate Response A} \\
    - \textbf{Cognitive Empathy}: Identifies the user's "overwhelmed" emotion but fails to understand that the root of the pain is the act of "memorizing" itself, rather than the exam outcome. \\
    - \textbf{Affective Empathy}: The language is gentle but lacks force, failing to synchronize with the user's intense frustration resulting from the repetitive labor. \\
    - \textbf{Motivational Empathy}: The question "Are you still memorizing things right now?" pulls the focus back to the task, contradicting the user's underlying motive of seeking a "pause" and "validation."
    
    \vspace{0.1cm}
    \textbf{Candidate Response B} \\
    - \textbf{Cognitive Empathy}: Precisely identifies and validates the user's painful \textbf{Cognitive Experience}, directly pinpointing the torment of the "memorizing" act with the metaphor of "torture." \\
    - \textbf{Affective Empathy}: Through strong expressions like "Oh god, just hearing that gives me a headache," it synchronizes with the user's emotional intensity, providing effective emotional validation. \\
    - \textbf{Motivational Empathy}: The suggestion to "find some instrumental music for you to help you clear your mind" is intended to offer emotional support and the possibility of a "pause," perfectly aligning with the user's core motive of seeking companionship over a solution.
    
    \vspace{0.2cm}
    \textbf{Rating} \\
    A: 3 points (Identifies the emotion but fails to validate the painful \textbf{Cognitive Experience}, and the motive runs counter to the user's needs.) \\
    B: 5 points (Accurately validates the user's core painful experience and offers high-quality emotional resonance and companionship-based support.)
    
    \vspace{0.1cm}
    \textbf{Final Choice:} B is better
}
\end{tcolorbox}

\clearpage 
\vpara{Acknowledgement.} 
We would like to thank the data team: Sixuan You, Yang Gao, Xueying Liu, Ziyi Zhu, Yuwei Yuan, and Yingtong Xu, for their extensive support in organizing and managing the data annotation process. We also thank the evaluation team: Jianjian Ruan, Qi Li, Ming Yang, Xiangfang Zheng, and Fangfei Lin, for their contributions in constructing the static evaluation suite. Their efforts were essential to the development of this work.

\bibliographystyle{abbrv}
\bibliography{0_main}

\end{document}